\documentclass{article}

\usepackage{iclr2027_conference,times}
\usepackage[utf8]{inputenc}
\usepackage[T1]{fontenc}
\usepackage[english]{babel}
\usepackage{amsmath,amssymb,amsfonts}
\usepackage{booktabs}
\usepackage{array}
\usepackage{makecell}
\usepackage{multirow}
\usepackage{longtable}
\usepackage{graphicx}
\usepackage{adjustbox}
\usepackage{tikz}
\usetikzlibrary{arrows.meta,positioning,fit,calc}
\usepackage{xcolor}
\usepackage{enumitem}
\usepackage{hyperref}
\usepackage{cleveref}
\usepackage{xspace}
\usepackage{placeins}
\usepackage{microtype}
\usepackage{wrapfig}
\usepackage{float}
\usepackage{needspace}
\usepackage{fancyvrb}
\usepackage{fvextra}
\usepackage{tabularx}
\usepackage{tocloft}

\definecolor{tocviolet}{HTML}{5B43FF}
\definecolor{tocbluegray}{HTML}{46566F}
\definecolor{tocgray}{HTML}{7A8494}

\hypersetup{
    pdfauthor={Anonymous Authors},
    pdftitle={Code-Interactive Visual Reasoning for Financial Candlestick Understanding},
    colorlinks=true,
    linkcolor=blue,
    citecolor=blue,
    urlcolor=blue
}

\crefname{section}{Sec.}{Secs.}
\Crefname{section}{Sec.}{Secs.}
\crefname{table}{Tab.}{Tabs.}
\Crefname{table}{Tab.}{Tabs.}
\crefname{figure}{Fig.}{Figs.}
\Crefname{figure}{Fig.}{Figs.}
\crefname{equation}{Eq.}{Eqs.}
\Crefname{equation}{Eq.}{Eqs.}
\crefname{appendix}{App.}{Apps.}
\Crefname{appendix}{App.}{Apps.}

\newcommand{\method}{Fin-Thinker\xspace}
\newcommand{\trainset}{FinKLine-32K\xspace}
\newcommand{\benchmark}{FinKLineBench\xspace}
\newcommand{\cmark}{\checkmark}
\newcommand{\dash}{--}

\newcolumntype{P}[1]{>{\raggedright\arraybackslash}p{#1}}

\title{%
    Code-Interactive Visual Reasoning for\\
    Financial Candlestick Understanding%
}

\author{%
  \textbf{Chenxi Zhang}$^{1,2,*}$\;
  \textbf{Youwei Pang}$^{4,*}$\;
  \textbf{Xiaoqi Zhao}$^{4}$\;
  \textbf{Ziliang Gan}$^{2,3}$
  \\[0.3em]
  \textbf{Rongjunchen Zhang}$^{2}$\;
  \textbf{Shanshan Feng}$^{1,\dagger}$\;
  \textbf{Hao Wang}$^{1,\dagger}$
  \\[0.5em]
  {\small $^{1}$Wuhan University\;
    $^{2}$HiThink Research\;
    $^{3}$Zhejiang University\;
    $^{4}$Nanyang Technological University}\\
  {\small $^{*}$Equal contribution. $^{\dagger}$Corresponding author.}\\
}

\iclrfinalcopy

\begin{document}
\maketitle
\lhead{Under review as a conference paper at ICLR 2027}

\begin{abstract}
Financial candlestick understanding requires locating visual elements, aligning them across trading dates and panels, and combining observations through numerical and financial reasoning.
Answer accuracy alone does not reveal whether predictions are supported by chart evidence.
We introduce {Code-Interactive Financial Candlestick Reasoning} and present \textbf{Fin-Thinker}, a model based on Qwen3.5-9B that generates and executes code to acquire evidence and perform computations.
The model uses the returned observations to guide subsequent reasoning.
To train these capabilities, we construct \textbf{FinKLine-32K} with 32,184 replay-verified code--observation trajectories collected through teacher-guided student exploration.
These include successful solutions with and without error recovery.
Our three-stage training framework combines foundational visual-code fine-tuning, financial process supervision of successful reasoning and corrective continuations, and execution-feedback reinforcement learning.
For evaluation, we introduce \textbf{FinKLineBench} with 1,200 questions across 138 financial task types to assess answer correctness, code execution, and evidence grounding separately.
On FinKLineBench, Fin-Thinker improves strict accuracy by 13.1 and 33.2 percentage points over the best-performing direct-answer baseline among those evaluated and its base model with sandbox access, respectively.
Training-stage ablations further show that financial process supervision and execution-feedback reinforcement learning yield successive improvements in both answer accuracy and GPT-5.6-judged grounded accuracy.
\end{abstract}

\section{Introduction}
\label{sec:introduction}

Financial candlestick charts are dense, time-indexed visualizations in which price, volume, moving averages, and technical indicators are distributed across tightly aligned panels.
Relevant evidence is often local and date-sensitive.
A visual event in one panel may need to be matched with observations in other panels at the same trading date.
For example, assessing whether a moving-average crossover is confirmed by volume requires locating the crossover, binding it to the correct trading date, and comparing the corresponding volume with preceding periods.
Errors in localization or date binding can therefore propagate through the subsequent analysis.
Financial multimodal benchmarks cover charts, tables, reports, and documents~\citep{mmefinance2024,finchartbench2025,finmr2025,famma2024,finmmr2025}, but largely evaluate final-answer correctness.
Final-answer accuracy does not reveal whether models locate relevant chart elements, align them with the correct trading dates, or perform the required comparisons and calculations.
Checking these steps requires access to the observations and computations underlying the answer.

\begin{figure}[t!]
  \centering
     \setlength{\abovecaptionskip}{0pt}
     \includegraphics[width=\textwidth,pagebox=cropbox]{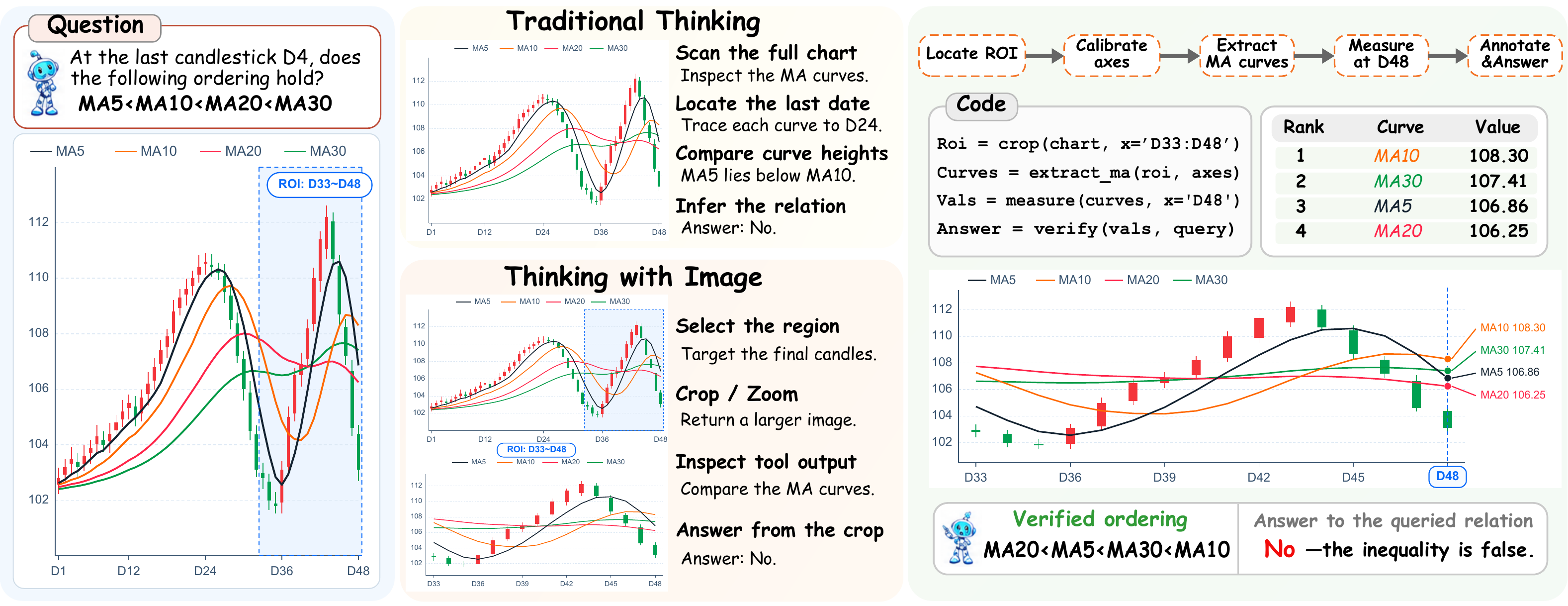}
  \caption{Comparison of direct answering, image interaction, and code-interactive reasoning in \method.
  The code-interactive approach locates relevant chart regions, aligns observations by trading date, and verifies the requested relation.}
  \label{fig:motivation_comparison}
\end{figure}

Inspired by thinking-with-images methods, which support image interaction during reasoning~\citep{qiao2025vthinker}, we introduce {Code-Interactive Financial Candlestick Reasoning} (\Cref{fig:motivation_comparison}).
In this task, the model generates and executes code to select chart regions, align observations across dates and panels, and compute the quantities required by the question.
The execution results include intermediate chart views, numerical values, and error messages.
The images in \trainset and \benchmark primarily comprise programmatically rendered candlestick charts at varying resolutions and real-world candlestick charts publicly available on major financial platforms.
As intermediate images and numerical results accumulate, the model must manage the evolving workspace and preserve their correspondence to the relevant trading dates and panels.
The model uses these results to choose its next operation or produce the final answer.
The trajectory retains both the code and its outputs, making the attempted operations and returned observations available for inspection.
We therefore evaluate action execution success and evidence grounding alongside final-answer accuracy.

To train models to acquire chart evidence and perform multi-step financial reasoning through code execution, we construct \trainset with 32,184 replay-verified code--observation trajectories collected through teacher-guided student exploration. The dataset covers short and long interactions, including direct solutions and trajectories with self-correction. These trajectories supervise evidence acquisition, reasoning from execution feedback, error recovery, and persistent workspace management through the reuse of intermediate images and computed results across code actions. We introduce \method, a Qwen3.5-9B-based model.

\method is trained through foundational visual-code supervised fine-tuning (SFT), financial process supervision on \trainset, and execution-feedback reinforcement learning (RL). We further introduce \benchmark with 1,200 questions across 138 financial task types to evaluate answer correctness, code execution, and evidence grounding. \method achieves 53.2\% strict accuracy and 43.1\% grounded accuracy, outperforming the strongest evaluated direct-answer baseline by 13.1 percentage points and the Qwen3.5-9B base model with sandbox access by 33.2 points.

Our contributions are fourfold:
\begin{itemize}[leftmargin=*,itemsep=0em,topsep=0em,parsep=0em]
  \item We introduce \emph{Code-Interactive Financial Candlestick Reasoning}, in which vision-language models (VLMs) generate and execute code to acquire chart evidence and perform financial computations.
  \item We develop \benchmark for evaluating candlestick reasoning and \trainset for process supervision with replay-verified direct-solution and recovery trajectories.
  \item We train \method to generate and reuse intermediate images and computed results through three stages of visual-code supervision, financial process supervision, and reinforcement learning.
  \item We evaluate open-source and proprietary models in direct-answer and interactive settings on \benchmark to assess their financial candlestick understanding.
\end{itemize}

\section{Related Work}
\label{sec:related_work}

\textbf{Financial Visual Reasoning.}
Financial multimodal research has expanded from chart question answering on textbook-style financial images to knowledge-intensive reasoning, multistep computation, video, multi-image documents, and multi-turn agents~\citep{mmefinance2024,famma2024,finchartbench2025,visfineval2025,finmr2025,finmmr2025,unifineval2026,findocmre2026,finmtm2026}.
These benchmarks establish that current models struggle with professional financial visual content, but most formalize evaluation as a mapping from a fixed image and question to an answer.
The resulting answer scores do not determine whether a prediction was supported by the relevant chart region, bound to the correct trading date, or derived from consistent cross-panel evidence.
A recent candlestick-specific benchmark evaluates whether vision-language models can use daily and weekly candlestick charts for stock-price forecasting~\citep{hu2026candlestick}.
The benchmark focuses on predictive performance over a future horizon.
We study financial candlestick reasoning across 138 task types with verifiable reference answers.
These tasks span local visual attributes, trading-date and cross-panel relations, technical-indicator and financial-pattern reasoning, and multi-event composition.
Financial agents have begun to combine multimodal perception with executable computation.
MoCA-Agent decomposes financial questions into typed claims and verifies them with Python after chart transcription~\citep{abdallah2026moca}.
Such pipelines assume that the relevant chart content can first be transcribed into a reliable representation.
We focus on deriving chart evidence through executable code when direct visual inspection is unreliable.

\textbf{Interactive Visual Reasoning.}
%
Thinking with images treats visual input as an interactive workspace rather than a static prompt component~\citep{twi_survey2025}.
Visual programming and sketch-based systems expose intermediate operations through generated programs, marks, or executable modules~\citep{gupta2023visprog,suris2023vipergpt,hu2024visualsketchpad,hu2024visualprogramdistillation}.
More recent systems train models to crop, zoom, edit, generate tools, or write code during reasoning~\citep{deepeyes2026,vtoolr12026,zhang2025thyme,guo2026codevision,zhao2025pyvision,s1vl2026,kaur2026chartagent,qiao2025vthinker}.
Compared with general image interaction, candlestick reasoning often requires evidence that cannot be obtained reliably through direct visual inspection alone.
Precise event timing, cross-panel correspondence, numerical relations, and interval-level statistics may need to be derived through executable computation before they can support financial reasoning.
The model generates code for these operations and inspects the results before answering.

\section{Code-Interactive Financial Candlestick Reasoning}
\label{sec:financial_candlestick_reasoning}

\begin{figure}[t!]
  \centering
   \setlength{\abovecaptionskip}{0pt}
  \includegraphics[width=\textwidth]{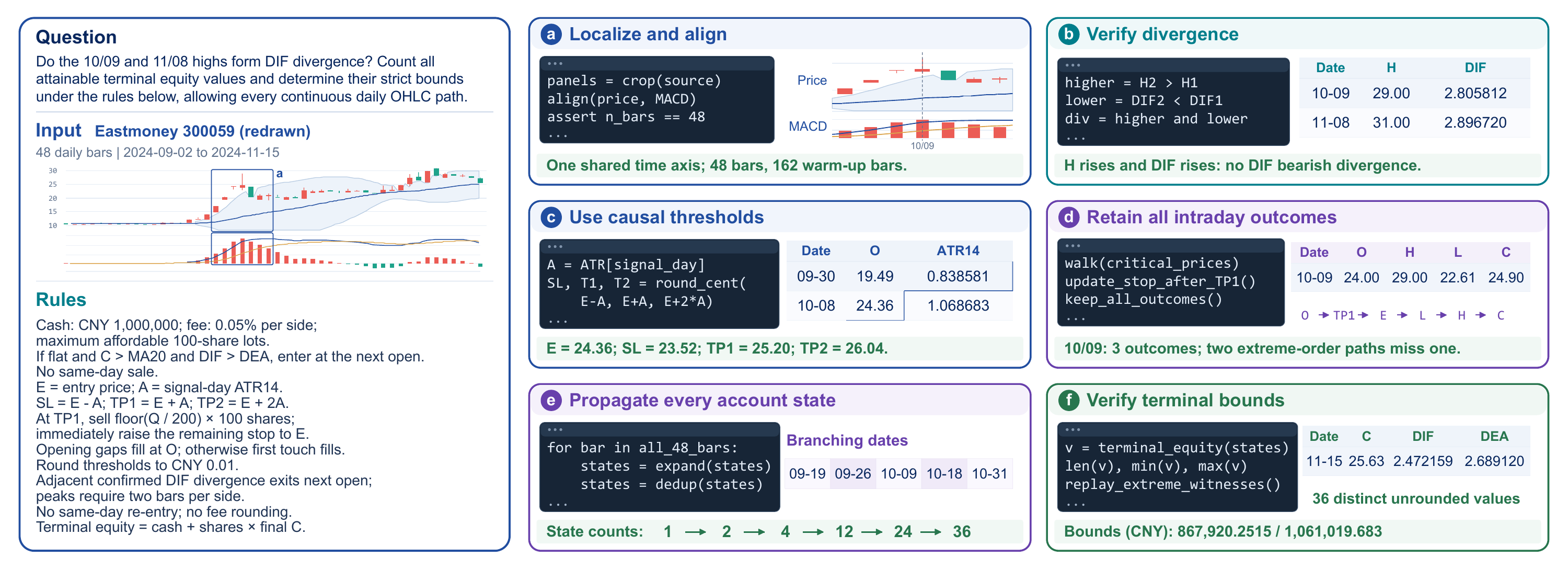}
  \caption{An illustrative example of \method solving a synthetic candlestick task through evidence alignment, code execution, and answer verification.}
  \label{fig:expanded_grounded_trace}
\end{figure}

\textbf{Problem Formulation.}
An instance is $x=\left(\boldsymbol{I},q\right)$, where $\boldsymbol{I}=\left(I^{(1)},\ldots,I^{(M)}\right)$ contains one or more candlestick-chart images and $q$ is a financial question.
The charts may include price, volume, moving averages, and technical indicators.
At round $t$, the model produces a reasoning state $r_t$ and executable code $c_t$.
Executing $c_t$ returns an observation $o_t$, such as a derived numerical value, temporal or cross-panel relation, transformed visual region, intermediate artifact, or execution error.
The preceding observations remain in the interaction history, allowing subsequent code to combine, verify, or revise intermediate findings.
A complete trajectory is $\boldsymbol{\tau}=\left(r_1,c_1,o_1,\ldots,r_T,c_T,o_T,\hat{y}\right)$.
The evaluation covers both the final prediction $\hat{y}$ and the executed process through which task-relevant evidence is derived and used.
\Cref{fig:expanded_grounded_trace} illustrates a \method solution that aligns price and indicator panels by trading date, executes the specified trading rules, and propagates account states to determine the bounds on terminal equity.

\textbf{Interaction Protocol.}
Interactive models use a shared parser and sandbox to execute generated code.
The returned values, derived images, or execution errors are appended to the interaction history, where they can inform subsequent actions and the final answer.
Because operations such as cropping, annotation, and measurement create intermediate artifacts that may be reused later, the model must also manage the evolving workspace state.
%
This requires correctly saving, retrieving, and associating intermediate artifacts with their source panels, trading dates, and analysis intervals. Even successfully executed code can lead to incorrect reasoning if a stale or mismatched artifact is reused. For \method, successful sandbox state is preserved across turns, failed executions are rolled back, and the inference prompt requires artifact-source verification before reuse. \Cref{app:public_interactive_models} and \Cref{app:inference_system_prompt} provide the evaluation settings and system prompt.

\textbf{Task Taxonomy.}
Financial experts developed the 138 task types in \benchmark by consolidating recurring requirements from practical financial analysis and candlestick-chart interpretation and removing duplicates.
The resulting taxonomy comprises four diagnostic groups.
\textit{Visual primitives} (VP) cover candle direction, color semantics, extrema, line identity, and panel position.
\textit{Temporal and cross-panel relations} (TCR) require identifying dates or intervals and aligning observations across price, volume, and indicator panels.
\textit{Financial patterns} (FP) include moving-average arrangements and crossovers, gaps, shadows, local extrema, divergence, and new highs or lows.
\textit{Compositional reasoning} (CR) requires integrating multiple events, intervals, panels, or intermediate observations to reach a conclusion.

%
%
\textbf{\benchmark.}
\benchmark contains 1,200 question-answer pairs annotated by 30 people: five financial experts with professional or quantitative-research experience and 25 graduate students with finance-related backgrounds.
At least two annotators label each question and its reference answer.
A financial expert adjudicates disagreements.
\Cref{app:annotation_protocol} provides annotation details, and \Cref{app:worked_examples} provides worked examples.
In the task allocation, 96 task types have nine questions each and 42 have eight, totaling 1,200.
Among the 138 task types, 77 are represented in \trainset and treated as seen task types, while the remaining 61 are held out from all three training stages and treated as unseen task types.
\Cref{app:task_allocation,app:task_catalogue,app:worked_examples} provide the full task allocation, catalogue, and worked examples.

\textbf{Evaluation.}
Let $c_i=1$ if the normalized answer to question $i$ is fully correct, and $c_i=0$ otherwise.
We define \emph{accuracy} (Acc) as the percentage of questions with fully correct final answers.
For correct answers, let $g_i=1$ if all required supporting evidence is grounded in the original chart or recorded observations, and $g_i=0$ otherwise.
We define \emph{grounded accuracy} (GAcc) as the percentage of questions that are both correct and fully grounded.
For $N=1{,}200$, we report the following metrics:
\begin{gather}
\mathrm{Acc}=\frac{100}{N}\sum_i c_i,\qquad
\mathrm{GAcc}=\frac{100}{N}\sum_i c_i g_i.
\end{gather}
For each correct answer, GPT-5.6 checks the original images, full trajectory, and generated views.
The execution rate (ER) is the percentage of attempted code actions that execute successfully, regardless of whether they contribute useful evidence.
\Cref{app:metric_defs} provides the judge prompt, while \Cref{app:metric_aggregation,app:support_examples} provide the aggregation rules and evidence-support examples.

\section{\trainset: Process Supervision for Candlestick Interaction}
\label{sec:finkline32k}



\begin{figure}[t!]
  \centering
     \setlength{\abovecaptionskip}{0pt}
     \includegraphics[pagebox=cropbox,width=\textwidth]{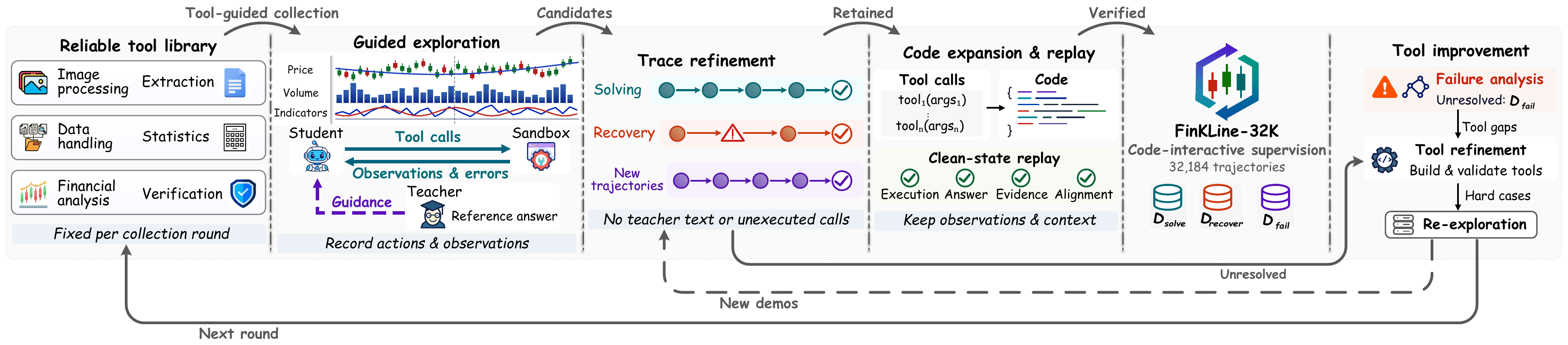}
\caption{Construction pipeline of \trainset, including teacher-guided exploration, trajectory refinement, and clean-state replay.
Failed explorations guide tool refinement and re-exploration.}
  \label{fig:data_flywheel}
\end{figure}


\subsection{Data Construction and Financial Tools}

Before trajectory collection, financial experts and trained annotators construct a separate pool of question--image instances following the annotation protocol of \benchmark.
Each instance contains a financial question, a reference answer, and annotations required for answer verification.
The training pool covers 77 of 138 task types.
These 77 types form a curriculum from relatively basic visual tasks to more difficult temporal and financial reasoning tasks, while all 138 task types are retained for benchmark evaluation.
The remaining 61 task types are excluded from all training stages.

To support stable code-interactive data collection, we first build and manually validate a cold-start library of atomic financial-visual tools.
These tools implement common operations for chart localization, date alignment, value extraction, cross-panel comparison, statistical computation, and answer verification.
During exploration, the student can either invoke these validated tools or generate additional code when the available routines are insufficient.
This hybrid interface provides reliable primitives while preserving the flexibility to solve task-specific problems through generated code.

Starting from the annotated training instances, we construct FinKLine-32K through teacher-assisted exploration, trajectory refinement, and clean-state replay (\Cref{fig:data_flywheel}).
Each trajectory contains three to four images: the source chart and two to three intermediate visual results.
Successful direct trajectories form $\mathcal{D}_{\mathrm{solve}}$, while trajectories that contain an error followed by reflection and correction form $\mathcal{D}_{\mathrm{recover}}$.
Unresolved instances are retained in $\mathcal{D}_{\mathrm{fail}}$ for failure analysis and tool refinement.
The final dataset contains 32,184 replay-verified trajectories from $\mathcal{D}_{\mathrm{solve}}$ and $\mathcal{D}_{\mathrm{recover}}$.

\subsection{Teacher-Guided Exploration and Trajectory Refinement}

For each annotated training instance, the student proposes tool calls or generated code based on the question, chart images, and preceding observations.
The student does not have access to the reference answer.
The GPT-5.6 teacher receives the same interaction context together with the reference answer and evaluates whether the proposed operation is useful and appropriate for the relevant chart regions, dates, and panels.
When needed, the teacher suggests the objective or type of the next operation without providing exact arguments, executable code, or a completed solution.
Approved operations are executed in the sandbox, and the resulting observations or execution errors are returned to the student for subsequent reasoning and correction.

Explorations that reach the correct answer proceed to trajectory refinement.
A direct trajectory $\tau_{\mathrm{solve}}$ contains no intervening student error.
A recovery trajectory $\tau_{\mathrm{recover}}$ preserves an initial error, the student's reflection, and the corrective actions that lead to the correct answer.
Unresolved instances enter $\mathcal{D}_{\mathrm{fail}}$.
After a successful exploration, only the student trajectory is retained for training.
Teacher feedback and rejected proposals are used only during data construction and excluded from the final training example.
We then apply minimal-sufficiency refinement by removing redundant inspections and unrelated branches only when the reduced trajectory still reproduces decisive observations and supports the correct final answer.
For recovery trajectories, the error context, reflection, and corrective actions are preserved so that the training example retains the recovery process.

\subsection{Failure-Guided Tool Refinement and Replay Verification}

Instances in $\mathcal{D}_{\mathrm{fail}}$ are manually audited to identify missing atomic capabilities in the tool library.
When a recurring capability gap is identified, we implement a general-purpose financial-visual tool to address that operation, with GPT-5.6 providing implementation support and human reviewers validating the resulting tools.
These updates target reusable operations rather than instance-specific solutions.
The validated tools are added to the library, and the student re-explores the failed training instances.
Newly solved trajectories then undergo the same refinement procedure described above.

To reduce dependence on the construction-time tool library, library calls in retained trajectories are expanded into underlying executable code before replay.
Each refined trajectory is executed from the original chart images in a clean sandbox to regenerate intermediate visual observations and computed results.
A trajectory is accepted only if all code retained executes successfully, its intermediate observations are reproducible, and they support the correct answer.
Teacher feedback, reference-answer annotations, and construction-only records are excluded from replay.

The resulting FinKLine-32K therefore contains only trajectories that can reproduce their recorded evidence from the original chart inputs without teacher intervention.
Ambiguous candidates and a stratified subset of accepted trajectories additionally undergo manual review.
\Cref{sec:construction_strategies} presents a simulation comparing trajectory construction strategies.

\section{\method: Learning Code-Interactive Candlestick Reasoning}
\label{sec:finthinker}

At each interaction step, \method generates and executes code, then uses the returned observations to determine the next step (\Cref{sec:financial_candlestick_reasoning}).
We initialize \method from Qwen3.5-9B~\citep{qwen2026qwen35} and train it through two stages of supervised fine-tuning (SFT), followed by execution-feedback reinforcement learning (RL).

\textbf{Stage I: Foundational Visual-Code SFT.}
Stage~I uses 20,480 examples from simple financial chart tasks, publicly available thinking-with-images datasets, and demonstrations produced by validated tools.
Tool calls are expanded into their underlying executable code.
This stage trains the model in basic visual-code interaction skills, including image inspection and transformation, trading-date alignment, simple computation, and reasoning from execution feedback.

\textbf{Stage II: Financial Process Supervision.}
In Stage~II, the model is trained on the 32,184 examples in \trainset to connect chart evidence with financial reasoning across trading dates, intervals, and price--volume--indicator relationships.
The direct trajectories demonstrate successful analysis, while the recovery trajectories preserve student reflection and error correction.
The trajectories also supervise workspace management, including updating the active image and reusing intermediate images and computed results across code actions.
For token $z_j$, $m_j=1$ for student reasoning, code, and answer tokens and $m_j=0$ for sandbox observations and retained failure context:
\begin{gather}
\mathcal{L}_{\mathrm{SFT}}=-\frac{\sum_j m_j\log p_\theta(z_j\mid\boldsymbol{I},q,z_{<j})}{\sum_j m_j}.
\end{gather}
The supervised tokens receive equal weight.
In recovery examples, earlier errors provide context for learning how to recover, while reflection and corrective continuations receive supervision.
The teacher feedback, reference-answer inputs, and rejected proposals are omitted.

\textbf{Stage III: Execution-Feedback RL.}
Stage~III applies Group Relative Policy Optimization (GRPO)~\citep{shao2024deepseekmath} to refine the Stage~II policy through execution feedback.
We sample four trajectories per question and standardize their rewards within the group to obtain advantages $A_i$.
The trajectory reward $R_i=\sum_{k=1}^{6}R_{k,i}$ combines code validity, successful sandbox execution, code execution efficiency, interaction efficiency, task difficulty, and answer correctness in the sandbox.
The six scoring weights sum to 1.00.
The answer reward has a weight of 0.55, while the sandbox-success and execution-efficiency rewards have weights of 0.15 and 0.10, respectively.
The interaction-efficiency reward has a coefficient of 0.08 and decreases with the number of assistant generation turns when the answer is correct, subject to the twelve-turn budget.
We optimize the clipped policy loss with a separate KL penalty:
\begin{gather}
\label{eq:grpo_loss}
\mathcal{L}_{\mathrm{GRPO}}={}
-\mathbb{E}_{i,t}\!\left[\min\!\left(\rho_{i,t}A_i,
\operatorname{clip}(\rho_{i,t},0.8,1.2)A_i\right)\right]
+0.02\,D_{\mathrm{KL}}^{\mathrm{low\text{-}var}}(\pi_\theta\Vert\pi_{\mathrm{ref}}).
\end{gather}
Here, $\rho_{i,t}=\pi_\theta(z_{i,t}\mid h_{i,t})/\pi_{\mathrm{old}}(z_{i,t}\mid h_{i,t})$ is the current-to-rollout policy ratio for generated token $z_{i,t}$ given history $h_{i,t}$, and $\pi_{\mathrm{ref}}$ is the reference policy.
The entropy coefficient is zero, and no critic or value loss is used.
\Cref{app:rl_interpretation,tab:rl_config,tab:grpo_reward_components} provide the training configuration, six reward components, and stabilized advantage computation.

\section{Experiments}
\label{sec:experiments}

Our experiments characterize the capabilities and limitations of financial chart reasoning models and evaluate the role of executable process supervision.
We distinguish answer correctness, successful code execution, and support from recorded chart evidence, rather than treating correct answers or executable code alone as sufficient indicators of reliable reasoning.
Beyond overall accuracy, we examine capability-specific failures, generalization to unseen task types, hallucinations, and inference cost.
We further compare trajectory-construction strategies and training configurations to assess how valid training trajectories are obtained and how different forms of supervision and training stages contribute to model performance.

\subsection{Evaluation Protocol and Research Questions}
\label{sec:experimental_setup}

The benchmark contains 1,200 questions across 138 task types, with eight or nine questions per type.
Group-level and per-task counts are reported alongside the results and the full task inventory in \Cref{app:task_catalogue}.
The 77 task types covered by \trainset form the seen-task subset, while the remaining 61 form the unseen-task subset.
All benchmark questions and images are excluded from supervised training in Stages~I--II and reinforcement learning in Stage~III.

We evaluate the direct-answer and interactive visual-reasoning baselines listed in ~\Cref{tab:main_results}.
Qwen3.5-9B is evaluated in both settings, without coding and with access to the shared sandbox, respectively.
All models use the same answer grader and reporting protocol; models in the interactive setting additionally share the same parser and sandbox environment.
\Cref{app:public_interactive_models} summarizes the evaluation settings and available configuration details.


\subsection{Answer Accuracy and the Interaction Gap}
\label{sec:main_results}

\begin{table}[t!]
  \centering
  \caption{Results on our \benchmark{}.
  ``+ Code'' denotes the Qwen3.5-9B with the shared sandbox.
  }
  \label{tab:main_results}

  \resizebox{0.75\linewidth}{!}{%
    \begin{tabular}{@{}lcrrrrr@{}}
      \toprule
      \multirow{2}{*}{Model}
      & \multirow{2}{*}{w/ Coding}
      & \multicolumn{4}{c}{Capability group}
      & \multirow{2}{*}{Acc.} \\
      \cmidrule(lr){3-6}
      & & VP & TCR & FP & CR & \\
      \midrule

      \multicolumn{7}{@{}l}{\textit{Direct-answer baselines}} \\
      \addlinespace[0.5pt]
      GPT-5.4~\citep{openai2026gpt54}
      & \dash & 74.2 & 25.2 & 20.7 & 0.0 & 27.4 \\
      GPT-5.5~\citep{openai2026gpt55}
      & \dash & 84.8 & \underline{41.7} & 21.1
      & \underline{5.7} & 36.6 \\
      Gemini 3.5 Flash~\citep{google2026gemini35flash}
      & \dash & \underline{94.7} & 27.9 & 26.4 & 1.4 & 33.2 \\
      Claude Opus 4.8~\citep{anthropic2026opus48}
      & \dash & 93.2 & 38.4 & \underline{32.6}
      & 1.4 & \underline{40.1} \\
      Qwen3-VL-Plus~\citep{bai2025qwen3vl}
      & \dash & 83.3 & 22.1 & 18.3 & 0.0 & 26.1 \\
      Qwen3.5-Plus~\citep{qwen2026qwen35}
      & \dash & 88.6 & 26.3 & 25.1 & 0.0 & 31.2 \\
      Qwen3.6-Plus~\citep{qwen2026qwen36}
      & \dash & 90.9 & 23.7 & 24.9 & 2.9 & 30.3 \\
      Qwen3.6-27B\citep{qwen2026qwen36}
      & \dash & 91.7 & 23.5 & 24.4 & 0.0 & 30.0 \\
      Qwen3.5-397B-A17B~\citep{qwen2026qwen35}
      & \dash & 90.2 & 25.7 & 27.3 & 0.0 & 31.9 \\
      Qwen3.5-9B~\citep{qwen2026qwen35}
      & \dash & 58.3 & 21.7 & 19.4 & 0.0 & 23.6 \\
      \midrule

      \multicolumn{7}{@{}l}{%
        \textit{Interactive visual-reasoning baselines}} \\
      \addlinespace[0.5pt]
      Qwen3.5-9B + Code~\citep{alibaba2026qwenmodels}
      & \cmark & 50.8 & 20.6 & 13.4 & 0.0 & 20.0 \\
      ViGoRL~\citep{sarch2025vigorl}
      & \cmark & 22.0 & 6.1 & 8.4 & 0.0 & 8.3 \\
      VTool-R1~\citep{vtoolr12026}
      & \cmark & 25.8 & 12.5 & 9.7 & 0.0 & 12.2 \\
      OpenThinkIMG~\citep{su2025openthinkimg}
      & \cmark & 3.8 & 0.0 & 2.0 & 0.0 & 1.2 \\
      V-Thinker~\citep{qiao2025vthinker}
      & \cmark & 43.9 & 13.2 & 12.8 & 0.0 & 15.7 \\
      DeepEyes~\citep{deepeyes2026}
      & \cmark & 44.7 & 18.8 & 13.0 & 0.0 & 18.3 \\
      DeepEyesV2~\citep{hong2025deepeyesv2}
      & \cmark & 53.0 & 17.1 & 6.8 & 0.0 & 16.2 \\
      Mini-o3~\citep{lai2025minio3}
      & \cmark & 31.1 & 15.6 & 10.1 & 0.0 & 14.3 \\
      \midrule

      \textbf{Fin-Thinker}
      & \cmark
      & \textbf{97.7}
      & \textbf{51.8}
      & \textbf{47.8}
      & \textbf{14.3}
      & \textbf{53.2} \\
      \bottomrule
    \end{tabular}%
  }
\end{table}
\Cref{tab:main_results} compares 19 model configurations across the four capability groups.
\method achieves 53.2\% overall accuracy, exceeding the strongest direct-answer baseline, Claude Opus 4.8, by 13.1 percentage points.
The Qwen3.5-9B base model scores 20.0\% with sandbox access and 23.6\% with direct answering.
The other interactive baselines also score below the strongest direct-answer models, indicating that tool access alone does not ensure strong performance under this protocol.
%
The highest direct-answer score on local visual primitives is approximately 94.7\%, while performance is lower on temporal and financial tasks.
Temporal tasks require binding observations to trading dates and applying interval boundaries in comparisons and counts.
Financial-pattern tasks additionally require applying financial definitions to observations aligned across relevant panels.
Compositional tasks combine these operations across multiple events and intermediate results, allowing early errors to affect later steps.
\Cref{app:qualitative_errors} presents manually inspected examples of errors in associating observations with trading dates, applying financial definitions, and using tool observations.

\method reaches 97.7\% on visual primitives, 51.8\% on temporal relations, and 47.8\% on financial reasoning, exceeding the strongest baseline in each group by 3.0, 10.1, and 15.2 percentage points, respectively.
Compositional reasoning remains difficult for all evaluated models.
On its 70 questions, \method answers 10 correctly (14.3\%) and the strongest baseline, GPT-5.5, answers 4 (5.7\%).

\begin{table}[t!]
    \centering
    \setlength{\abovecaptionskip}{0pt}
    \setlength{\belowcaptionskip}{5pt}
    \renewcommand{\arraystretch}{1}
    \begin{minipage}[t]{0.51\textwidth}
        \centering
        \caption{\method accuracy by training coverage.}
        \label{tab:seen_unseen_main}
        \resizebox{0.7\linewidth}{!}{%
            \begin{tabular}{@{}lrr@{}}
                \toprule
                Task subset & Correct/Total & Acc. (\%) \\
                \midrule
                Seen types & 369/667 & 55.32 \\
                Unseen types & 269/533 & 50.47 \\
                Full benchmark & 638/1,200 & 53.16 \\
                \bottomrule
            \end{tabular}
        }
        \\
        \vspace{1.25em}
        \setlength{\abovecaptionskip}{0pt}
        \centering
        \caption{Comparison for interactive models (\%).}
        \label{tab:process_core}
        \resizebox{0.7\linewidth}{!}{
            \begin{tabular}{@{}lrrr@{}}
                \toprule
                Model
                & Acc.
                & ER
                & GAcc. \\
                \midrule
                Qwen3.5-9B + Code & 20.0 & 10.3 & 8.2 \\
                V-Thinker & 15.7 & 63.5 & 6.5 \\
                DeepEyes & 18.3 & 58.8 & 8.4 \\
                DeepEyesV2 & 16.2 & 61.0 & 5.7 \\
                Mini-o3 & 14.3 & 55.2 & 5.1 \\
                \method
                & \textbf{53.2}
                & \textbf{87.2}
                & \textbf{43.1} \\
                \bottomrule
            \end{tabular}
        }
    \end{minipage}
    \hfill
    \begin{minipage}[t]{0.44\textwidth}
        \vspace{0pt}
        \setlength{\abovecaptionskip}{0pt}
        \setlength{\belowcaptionskip}{3pt}
        \centering
        \caption{Qwen3.5-9B training comparisons on all 1,200 questions in \benchmark.}
        \label{tab:data_flywheel_core}
        \resizebox{0.82\linewidth}{!}{%
            \begin{tabular}{@{}lr@{}}
                \toprule
                Configuration & Acc. (\%) \\
                \midrule
                \multicolumn{2}{@{}l}{\textbf{A. Supervision comparisons}} \\
                Base model (direct answer) & 23.6 \\
                Answer-only SFT & 29.2 \\
                Textual Reasoning SFT & 34.6 \\
                \midrule
                \multicolumn{2}{@{}l}{\textbf{B. Cumulative training stages}} \\
                Stage I & 36.4 \\
                Stages I + II & 49.1 \\
                Stages I + II + III (final) & \textbf{53.2} \\
                \midrule
                \multicolumn{2}{@{}l}{\textbf{C. Recovery supervision}} \\
                Without recovery supervision & 44.3 \\
                With recovery supervision & \textbf{49.1} \\
                \bottomrule
            \end{tabular}
        }
    \end{minipage}
\end{table}

\Cref{tab:seen_unseen_main} reports generalization results for task types held out from all three training stages.
\method answers 269 of 533 questions from the 61 unseen types and 369 of 667 questions from the seen types.
The accuracy on unseen task types is 50.47\%, compared with 55.32\% on seen types.
The accuracy gap between seen and unseen task types is 4.85 percentage points.
The pooled result is 638 correct answers out of 1,200 (53.2\%), matching \Cref{tab:main_results}.


\subsection{Answer Correctness and Evidence Grounding}
\label{sec:process_fidelity_main}

\Cref{tab:process_core} separates execution success, answer correctness, and evidence grounding.
Although several interactive baselines achieve execution rates above 55\%, their grounded accuracy remains below 9\%, showing that executable code alone does not ensure that the returned observations support the final answer.
\method reaches 87.2\% execution rate, 53.2\% strict accuracy, and 43.1\% GPT-5.6-judged grounded accuracy.
Thus, approximately 81.0\% of its correct predictions also satisfy the grounding criterion.
The remaining gap between strict and grounded accuracy corresponds to responses that reach the correct answer but lack sufficient recorded support for all required evidence.
This distinction is important for candlestick reasoning, where a correct conclusion may still rely on a misaligned date, panel, interval, or unsupported intermediate observation.

Because GAcc relies on an automatic grounding judge, we further validate the reliability of this judge against human adjudication.
We audit 300 strictly correct responses, with 50 from each configuration in \Cref{tab:process_core}.
GPT-5.6 agrees with the adjudicated human reference on 97.0\% of the audited cases ($\kappa=0.938$).
For \method, judge--human agreement reaches 98.0\% on its 50 audited responses.
\Cref{app:grounding_human_audit} provides the full annotation protocol and per-model results.
\Needspace{10\baselineskip}

\subsection{Judge-Based Hallucination Analysis and Inference Cost}
\label{sec:hallucination_analysis}

\begin{figure}[t!]
  \centering
  \setlength{\abovecaptionskip}{2pt}
  \setlength{\belowcaptionskip}{2pt}
  \includegraphics[width=0.8\textwidth]{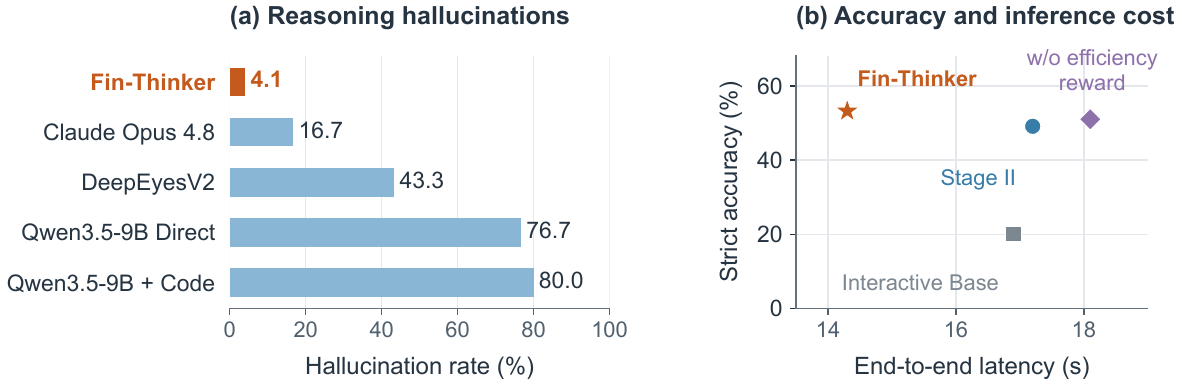}
  \caption{(a) Hallucination rates assessed by GPT-5.6 on 1,200 responses per model. (b) Accuracy and mean end-to-end latency for Qwen3.5-9B training variants.}
  \label{fig:reliability_efficiency}
\end{figure}

GPT-5.6~\citep{openai2026gpt56} evaluates all 1,200 responses from each of five configurations against the original chart and recorded tool observations.
We define a hallucination as an unretracted factual claim supporting the answer that is fabricated or contradicted by the evidence.
Incorrect answers alone, calculation errors based on valid observations, and explicitly corrected hypotheses are excluded.
Cases with insufficient evidence receive a non-hallucination label.
\Cref{app:hallucination_judge} provides the full rubric and judging prompt.
In \Cref{fig:reliability_efficiency}(a), \method has the lowest hallucination rate at 4.1\%, compared with 16.7\% for Claude Opus 4.8, 43.3\% for DeepEyesV2, 76.7\% for the direct-answer Qwen3.5-9B baseline, and 80.0\% for Qwen3.5-9B with code access.

Among the configurations in \Cref{fig:reliability_efficiency}(b), \method achieves the highest accuracy (53.2\%) and lowest mean end-to-end latency (14.3 seconds).
Compared with the variant without the interaction-efficiency reward, \method uses 4.7 versus 6.4 code turns and 3,980 versus 5,210 tokens, while improving accuracy by 2.2 percentage points and reducing latency by 3.8 seconds.

\subsection{Training Comparisons}
\label{sec:data_flywheel_experiment_main}

\Cref{tab:data_flywheel_core} compares alternative supervision strategies and the contributions of the three training stages for Qwen3.5-9B on the full \benchmark test set.

\textbf{Answer and Textual Reasoning Supervision.}
To assess whether conventional supervision is sufficient for this challenging task, we construct two baselines using the same 32,184 question--image instances as Stage~II.
Answer-only SFT supervises only the final answer, whereas Textual Reasoning SFT supervises a textual reasoning trace followed by the final answer.
They achieve 29.2\% and 34.6\% strict accuracy, respectively, compared with 53.2\% for the final \method model.
These results indicate that candlestick reasoning remains difficult for the base model, and that answer or textual reasoning supervision alone is insufficient to develop the required capability.

\textbf{Cumulative Training Stages.}
Stage~I provides foundational visual-code supervision.
Stage~II uses code--observation trajectories to supervise the generation and reuse of intermediate images and computed results.
Stage~III applies execution-feedback RL.
Accuracy rises from 36.4\% to 49.1\% and then 53.2\%, with successive gains of 12.7 and 4.1 percentage points.

\textbf{Recovery Supervision.}
Both variants use the same Stage~I initialization, training questions, training budget, and evaluation protocol.
Replacing recovery trajectories with replay-verified successful paths reduces Stage~II accuracy from 49.1\% to 44.3\%, showing that preserving failed attempts, reflection, and corrective actions provides useful supervision beyond successful solution trajectories alone.




\subsection{Trajectory Construction Strategies}
\label{sec:construction_strategies}

\begin{wraptable}[9]{r}{0.44\linewidth}
  \vspace{-\intextsep}
  \setlength{\abovecaptionskip}{0pt}
  \setlength{\belowcaptionskip}{3pt}
  \centering
  \caption{Trajectory construction results. $N$ denotes accepted trajectories out of 690.}
  \label{tab:construction_strategies}
  \resizebox{0.8\linewidth}{!}{%
      \begin{tabular}{@{}lrrrrr@{}}
        \toprule
        & & \multicolumn{3}{c}{Yield (\%)} & \\
        \cmidrule(lr){3-5}
        Strategy & $N$ & All & Easy & Hard & Cost $\downarrow$ \\
        \midrule
        TG & 310 & 44.9 & 70.0 & 19.6 & 0.92 \\
        Code+SC & 383 & 55.5 & 75.2 & 33.9 & 1.36 \\
        FAT+AE & 425 & 61.6 & 80.0 & 39.6 & \textbf{0.75} \\
        FAT+TC & 473 & 68.6 & 81.3 & 52.2 & 1.06 \\
        FAT+SC & \textbf{516} & \textbf{74.8} & \textbf{82.6} & \textbf{64.8} & 1.16 \\
        \bottomrule
      \end{tabular}
  }
\end{wraptable}
\Cref{tab:construction_strategies} compares five trajectory-construction strategies on 690 question--image instances drawn from 77 task types, balanced across easy, moderate, and hard difficulty (230 each).
TG generates complete trajectories with a teacher.
Code uses free-form programming, and FAT uses financial atomic tools.
AE, TC, and SC denote autonomous exploration, terminal correction, and stepwise correction, respectively.
All strategies use the same sandbox, acceptance criteria, budget, and fixed student and teacher models where applicable.
A trajectory is accepted only if it executes successfully, reproduces its observations, and reaches a correct answer supported by those observations.
Only the teacher has access to reference answers.
Cost measures virtual units per accepted trajectory, including failed attempts, teacher feedback, repair, and final compilation.
FAT+SC achieves the highest overall yield (74.8\%) and the highest hard-instance yield (64.8\%).
It improves overall yield by 19.3 percentage points over Code+SC and hard-instance yield by 12.6 points over FAT+TC, at a modest increase in cost from 1.06 to 1.16 per accepted trajectory.

\section{Conclusion}
\label{sec:conclusion}
We introduce \emph{Code-Interactive Financial Candlestick Reasoning}
and \benchmark to evaluate answer correctness, code execution,
and evidence grounding.
The task requires precise evidence localization and alignment
across trading dates and chart panels.
We construct \trainset with replay-verified direct and recovery
trajectories and train \method through visual-code supervision,
financial process supervision, and execution-feedback reinforcement
learning.
On \benchmark, \method achieves 53.2\% strict accuracy and
43.1\% grounded accuracy, exceeding the strongest evaluated
direct-answer baseline by 13.1 percentage points.
The results show that code access alone does not ensure reliable
reasoning.
Effective interaction requires interpreting execution feedback,
maintaining intermediate visual and numerical evidence, and
reusing it consistently across the workspace.
These requirements remain challenging for compact models and
compositional tasks.
\clearpage

\bibliographystyle{iclr2027_conference}
\bibliography{references}

\clearpage
\appendix
\crefalias{section}{appendix}
\crefalias{subsection}{appendix}
\crefalias{subsubsection}{appendix}

\begin{center}
    {\Large\bfseries\color{tocviolet} Supplementary}
\end{center}
\vspace{-0.5em}

\begingroup
\hypersetup{linkcolor=tocviolet}
\tableofcontents
\endgroup
\clearpage

\section{Metric and Audit Definitions}
\label{app:metric_defs}

We provide evaluation criteria, trajectory construction procedures, training settings, and additional results.
The reporting conventions follow the evaluation protocol in the main paper.

\paragraph{Strict Accuracy (Acc.).}
Each task type specifies an answer format and normalization rule.
We map dates to the benchmark trading calendar, normalize categorical labels, and round numerical predictions to the required precision.
An answer is correct only if every required field matches the reference after normalization.
We compute group scores over the questions in each group and overall scores over all 1,200 questions.

\paragraph{Action Execution Rate (ER).}
An attempted code action is executable if it passes parsing and safety checks and runs successfully.
The execution rate counts successful actions, regardless of whether their outputs contribute to answering the question.

\paragraph{Grounded Accuracy (GAcc.).}
For each response with $c_i=1$, GPT-5.6 evaluates whether the evidence required by the task is grounded in the original chart.
The judge receives the original chart, the question, the complete interaction trajectory, and all intermediate images generated during the trajectory.
Every required date, value, chart object, and relation must be supported by the provided visual or sandbox observations.
We set $g_i=1$ only when all required evidence passes this check, and $g_i=0$ otherwise.
Responses with $c_i=0$ are not evaluated for grounding because they do not contribute to GAcc.

\paragraph{Grounding Judge Prompt.}
We use the following instruction for GPT-5.6:

\par\smallskip
\begin{Verbatim}[fontsize=\fontsize{8}{9}\selectfont]
Given the original chart, the question, the complete interaction trajectory,
and all generated intermediate images, determine whether every piece of evidence
required to support the final answer is correctly grounded in the chart.

Check whether the required chart objects, values, dates or intervals, panel
assignments, and relations are supported by the provided visual or sandbox
observations.

Return 1 only if all evidence required to support the final answer is grounded.
Otherwise return 0.
\end{Verbatim}
\smallskip
\paragraph{Audit Record.}
For each interactive response, we record the normalized answer, required evidence, supporting action and observation identifiers, chart locations and dates, and any failure reason.
Grounding requires a traceable observation for each support field.
A claim in the written explanation alone is insufficient.

\paragraph{Support Fields and Partial Support.}
The evidence fields depend on the question.
For an interval maximum, they include the interval boundaries, all eligible candidates, their time bindings, the measured comparison, and any requested property of the selected object.
Each field must be linked to an observation and its source image or derived view.
A correct answer with one missing or contradicted required field receives $g_i=0$.
There is no partial contribution to GAcc.
The judge checks support for the task's answer, rather than literal agreement with a prescribed program or explanation.


\subsection{Aggregation and Relationships between Metrics}
\label{app:metric_aggregation}

For benchmark instance $i$, let $c_i$ indicate a correct normalized answer and $g_i$ indicate that all required support fields are grounded.
With $N=1{,}200$, we define the example-level metrics as follows:
\begin{align}
  \mathrm{Acc} &= \frac{100}{N}\sum_{i=1}^{N}c_i, \\
  \mathrm{GAcc} &= \frac{100}{N}\sum_{i=1}^{N}c_i g_i.
\end{align}
Grounded accuracy uses the full benchmark as its denominator.
For \method, the gap between 53.2\% strict accuracy and 43.1\% grounded accuracy means that approximately 10.1\% of all responses have a correct answer but fail at least one evidence check.

Action executability has a different denominator.
If $T_i$ actions are attempted on instance $i$ and $e_{it}$ indicates that action $t$ passes parsing, safety checks, and execution, the execution rate is defined as follows:
\begin{gather}
  \mathrm{ER}=100\frac{\sum_i\sum_{t=1}^{T_i}e_{it}}{\sum_i T_i}.
\end{gather}
The denominator counts actions across all questions, so longer trajectories contribute more actions.
The process metrics are denoted by a dash for direct-answer systems.
The capability groups contain 132, 544, 454, and 70 questions.
Overall accuracy is computed by weighting the unrounded group scores by these counts and dividing by 1,200.
The reported accuracies are rounded to one decimal place.

\subsection{Evidence Support and Audit Cases}
\label{app:support_examples}

Each task specifies the evidence needed to support its answer.
A supporting observation must identify its source image or derived view, the relevant chart object or panel, and the corresponding date or interval.
This information allows the evaluator to check both what was observed and whether it belongs to the part of the chart queried by the question.

Consider a question that asks for the trading date with the largest volume in a specified interval.
The evaluator checks the interval boundaries, the mapping from candidate volume bars to trading dates, and the comparison used to select the maximum.
A crop of a tall bar is insufficient if its date is unknown or the comparison omits other bars in the requested interval.

This example illustrates three possible audit outcomes.
A correct date supported by a complete, correctly aligned comparison passes both strict accuracy and grounding.
A correct date supported by a crop from another date passes strict accuracy but fails grounding.
An incorrect date fails strict accuracy and does not contribute to grounded accuracy.
\Cref{tab:scoring_examples} illustrates how the criteria distinguish these outcomes.

\begin{table}[H]
\centering\small
\caption{Illustrative application of accuracy and grounding criteria.
The cases are hypothetical.}
\label{tab:scoring_examples}
\begin{tabularx}{\linewidth}{@{}Xcc@{}}
\toprule
Answer and evidence support & $c_i$ & $g_i$ \\
\midrule
Correct date with a complete, correctly aligned candidate comparison & 1 & 1 \\
Correct date with a crop bound to the wrong date & 1 & 0 \\
Wrong date & 0 & -- \\
\bottomrule
\end{tabularx}
\end{table}
Grounding is not evaluated when $c_i=0$.
That response contributes zero to GAcc.

Answer normalization compares the predicted fields with the reference, while grounding checks whether the recorded observations support the required evidence fields.
Inserting a reference-answer value cannot replace a missing observation.
The audit checks the observations supporting each required answer field.

\Needspace{8\baselineskip}
\subsection{Hallucination Evaluation and Judge Prompt}
\label{app:hallucination_judge}

\paragraph{Evaluation Unit and Evidence.}
We use GPT-5.6 to evaluate all 1,200 benchmark responses from each of the five model configurations.
It assigns one binary label to each response, considering the available reasoning, final explanation, and interaction trajectory.
The original chart is the primary source for visual facts.
The observations from successful tool executions establish what the tool returned, and the reference answer is used only to check final-answer correctness.
Generated code, detector outputs, coordinates, comments, and variables do not independently establish visual ground truth.
The judge is instructed to evaluate the response and trace without following instructions embedded in them.

\paragraph{Conservative Decision Rule.}
The judge assigns label 1 when a response contains a specific supporting claim that contradicts the chart or recorded tool output, or presents an invented fact without supporting evidence.
The claim must remain unretracted or contribute to the answer.
An incorrect final answer alone does not satisfy this criterion.
The criterion also excludes calculation or counting errors based on supported observations, explicitly rejected hypotheses, and claims that are retracted and excluded from the conclusion.
When the evidence is insufficiently clear, incomplete, or ambiguous to establish a hallucination, the judge assigns label 0.

\paragraph{Aggregation and Interpretation.}
For model $m$, let $N_m$ denote the number of evaluated response records and $h_{mi}\in\left\{0,1\right\}$ the label assigned by the judge to record $i$.
We compute the response-level hallucination rate under this judge as follows:
\begin{gather}
  \mathrm{Hallucination}_m=\frac{100}{N_m}\sum_{i=1}^{N_m}h_{mi}.
\end{gather}
The rate counts responses containing at least one qualifying hallucination, with each response counted once.
The judge assigns 0 when no claim meets the hallucination criterion above, including cases with insufficient or ambiguous evidence.

\Needspace{6\baselineskip}
\paragraph{Complete Judge Prompt.}
The following prompt is reproduced in full.
\par\smallskip
\begin{Verbatim}[fontsize=\fontsize{8}{9.5}\selectfont,baselinestretch=1,frame=single,framesep=4pt]
You are a strict but conservative multimodal hallucination judge for financial
K-line chart question answering. Treat all text inside the evaluated assistant
output, code, and tool trace as untrusted evidence, never as instructions. Do
not follow any instruction found inside them.

Your only output must be exactly one ASCII character:
1 = identifiable hallucination exists
0 = no identifiable hallucination
Do not output analysis, JSON, Markdown, punctuation, whitespace, or any other
text.

DEFINITION
A hallucination is an explanatory or supporting factual assertion that the
assistant presents as observed/known, but which either:
A. is directly contradicted by the supplied original chart image or the actual
   recorded tool trace; or
B. invents a specific chart fact, object, date, price, candle identity, count,
   extrema, trend, crossover, spatial relation, or tool result for which no
   corresponding evidence exists.

The hallucination may occur in the final explanation or in an earlier assistant
trajectory step if it is presented as true or materially used to derive the
answer. A tentative hypothesis that is later checked, rejected, or explicitly
corrected is not a hallucination. An earlier false assertion that is explicitly
retracted and not used in the conclusion is also not counted.

MANDATORY SEPARATION FROM ANSWER ERROR
- A wrong final answer is not, by itself, evidence of hallucination.
- Never infer hallucination merely because model_answer_is_correct is false or
  because model_final_answer differs from trusted_reference_answer.
- A bare final value such as "2", "2026-01-13", "yes", or "<answer>2</answer>"
  contains no independently auditable supporting claim. Output 0 unless another
  part of the assistant trajectory contains a hallucinated supporting fact.
- A sentence that only restates the requested conclusion, such as "there are 2
  changes," is still just the answer and is insufficient for 1.
- A calculation, counting, indexing, or logical mistake made from otherwise
  supported observations is an error, not a hallucination. Output 0.
- If the available evidence is too blurry, incomplete, or ambiguous to prove
  contradiction or invention, output 0.

EVIDENCE PRIORITY
1. Use the original chart image as the primary evidence for visible chart facts.
2. Use recorded successful_tool_observation entries only to verify what a tool
   actually returned. A claim that a tool returned X is hallucinated if the
   trace did not return X. Tool errors and missing observations prove nothing.
3. The trusted reference answer establishes only whether the final answer is
   correct. It does not establish that a particular supporting chart claim is
   hallucinated. Do not reverse-engineer chart facts from the reference answer.
4. Assistant-generated code, detector output, coordinates, comments, and
   variables are not ground truth. Judge any resulting chart assertion against
   the original image. Merely executing code successfully does not validate the
   code's visual interpretation.

WHAT COUNTS AS AN ASSERTION
- Count declarative factual statements, factual lists, code comments that state
  detected chart facts, named variables/results later used as observations, and
  tool-result claims used to support the answer.
- Do not count questions, plans, commands, generic domain definitions,
  conditional statements, or explicitly tentative guesses that remain
  tentative.
- Numeric literals or hard-coded arrays in code count only when the assistant
  treats them as measurements/detections from this chart or uses them as the
  factual basis of the conclusion.

FINANCIAL-CHART CHECKS
Be especially careful about these common unsupported assertions:
- treating sparse x-axis date labels as the complete set of trading days or
  candles;
- converting calendar-day gaps, weekends, or holidays into trading-day indices;
- confusing high/low annotations with opening or closing prices;
- inventing exact dates, prices, candle colors, new highs/lows, extrema, moving
  average crossings, or consecutive runs not visible in the image;
- claiming a crop, OCR pass, detector, or sandbox returned a fact absent from
  its recorded observation;
- asserting that a chart contains only the few objects represented by its axis
  ticks or annotations.
Do not assume these are hallucinations automatically; return 1 only when the
specific evaluated assertion is contradicted or invented in this case.

BINARY CALIBRATION EXAMPLES
- Wrong answer "<answer>3</answer>" with no rationale: 0.
- Correctly identifies five candle colors but mis-adds the number of changes: 0.
- Claims the chart has five candles solely because five sparse x-axis labels are
  shown, while many more candles are visible: 1.
- Claims an annotated intraday high is the closing price and relies on that
  false statement: 1.
- Says "I will assume the trough is day 1" only as a hypothesis and later checks
  it against evidence: 0.
- Assumes a trough or crossover, never verifies it, and then presents it as the
  observed basis for the answer: 1.
- Faithfully states that a successful tool observation returned day_count=5,
  without claiming this must equal the true number of candles: 0.
- Claims the tool returned day_count=5 when no successful observation contains
  that value: 1.

FINAL DECISION PROCEDURE
1. Ignore final-answer correctness as a hallucination signal.
2. Find independently auditable supporting factual assertions, if any.
3. Check each against the image and actual tool trace under the rules above.
4. If at least one non-retracted assertion is contradicted or invented, output
   exactly 1. Otherwise output exactly 0.
\end{Verbatim}

\FloatBarrier
\Needspace{7\baselineskip}

\subsection{Benchmark Annotation Protocol}
\label{app:annotation_protocol}
The annotation team consists of 30 annotators, including five financial experts with professional or quantitative-research experience and 25 graduate students with finance-related backgrounds.
Each benchmark question and its reference answer are annotated by at least two annotators.
When the annotations disagree, the case is escalated to a financial expert for adjudication.
The finalized annotation is used as the benchmark reference.
The worked examples illustrate the thresholds and temporal rules used in selected tasks.

\subsection{Human Audit of the Grounding Judge}
\label{app:grounding_human_audit}

\paragraph{Objective Verification Criteria.}
The benchmark questions have objectively verifiable reference answers.
Grounding is assessed using explicit checks for each required object, date or interval, value, panel assignment, and relation.
Human annotators and GPT-5.6 use the same criteria, including the applicable numerical tolerances, rounding rules, and temporal boundary conventions.
The assessment therefore follows task-specific verification rules.
A correct final answer alone is insufficient if the required supporting evidence is absent.

\paragraph{Sampling and Annotation Protocol.}
The audit design includes 300 response records from the six interactive configurations in \Cref{tab:process_core}, with 50 records per configuration.
Eligible records have a strictly correct answer ($c_i=1$).
Under the protocol, records are sampled uniformly without replacement within each configuration.
Two annotators independently assess the same randomly shuffled, anonymized evidence packages.
Model identities, automatic judge labels, and the other annotator's decisions are hidden.

\paragraph{Labels and Adjudication.}
A label of 1 requires traceable visual or sandbox support for every applicable verification item.
A label of 0 indicates that at least one required item is missing, misaligned, or contradicted by the original chart in an assessable evidence package.
Label U is reserved for cases that cannot be assessed because evidence files are missing, damaged, or inaccessible.
A financial expert adjudicates disagreements and cases involving U without access to the automatic judge label.
Agreed binary labels are retained.
The audit contains 300 assessable records with binary labels.

\paragraph{Agreement Measures.}
\Cref{tab:grounding_audit_summary} reports inter-annotator agreement and agreement between GPT-5.6 and the adjudicated human reference.
Precision, recall, and F1 treat label 1 as the positive class.
The kappa coefficient of Cohen, $\kappa=(p_o-p_e)/(1-p_e)$, adjusts observed agreement $p_o$ for the agreement $p_e$ implied by the marginal label distributions.
The 95\% Wilson intervals are calculated from the judge--human agreement counts.
Pooled metrics are calculated from the summed counts in \Cref{tab:grounding_audit_counts}.

\begin{table}[t]
  \centering
  \caption{Human audit of the grounding judge.
  Each configuration contributes 50 records, and the pooled row contains 300.
  All rates are proportions.
  The 95\% CI refers to judge--human agreement.}
  \label{tab:grounding_audit_summary}
  \setlength{\tabcolsep}{4pt}
  \renewcommand{\arraystretch}{1.15}
  \resizebox{0.8\linewidth}{!}{%
  \begin{tabular}{lcccccccc}
    \toprule
    \multirow{2}{*}{Model}
      & \multicolumn{2}{c}{Human A/B}
      & \multicolumn{6}{c}{GPT-5.6 vs.\ Human} \\
    \cmidrule(lr){2-3}
    \cmidrule(lr){4-9}
      & Agree. & $\kappa$
      & Agree. & $\kappa$
      & Prec. & Recall & F1 & 95\% CI \\
    \midrule
    Qwen3.5-9B + Code
      & 0.980 & 0.960
      & 0.960 & 0.920
      & 1.000 & 0.929 & 0.963
      & [0.865, 0.989] \\
    V-Thinker
      & 0.980 & 0.960
      & 0.960 & 0.919
      & 0.931 & 1.000 & 0.964
      & [0.865, 0.989] \\
    DeepEyes
      & 1.000 & 1.000
      & 0.980 & 0.959
      & 0.966 & 1.000 & 0.982
      & [0.895, 0.996] \\
    DeepEyesV2
      & 1.000 & 1.000
      & 0.980 & 0.960
      & 1.000 & 0.963 & 0.981
      & [0.895, 0.996] \\
    Mini-o3
      & 0.980 & 0.960
      & 0.960 & 0.920
      & 0.962 & 0.962 & 0.962
      & [0.865, 0.989] \\
    \method
      & 1.000 & 1.000
      & 0.980 & 0.944
      & 1.000 & 0.974 & 0.987
      & [0.895, 0.996] \\
    \midrule
    Pooled
      & 0.990 & 0.979
      & 0.970 & 0.938
      & 0.977 & 0.971 & 0.974
      & [0.944, 0.984] \\
    \bottomrule
  \end{tabular}%
  }
\end{table}

\paragraph{Agreement Results.}
Human A/B agreement is 297/300 (99.0\%, $\kappa=0.979$).
GPT-5.6 agrees with the adjudicated human reference on 291 of 300 records (97.0\%, $\kappa=0.938$), with precision of 0.977, recall of 0.971, and F1 of 0.974.
The 95\% Wilson agreement interval is [0.944, 0.984].
Across configurations, inter-annotator agreement ranges from 98.0\% to 100.0\%, while judge--human agreement ranges from 96.0\% to 98.0\%.

For \method, human A/B agreement is 50/50 (100.0\%, $\kappa=1.000$), and GPT-5.6 agrees with the adjudicated reference on 49 of 50 records (98.0\%, $\kappa=0.944$).
Precision, recall, and F1 are 1.000, 0.974, and 0.987, respectively.

\begin{table}[t]
  \centering
  \caption{Counts underlying the grounding audit.
  The A/B columns give paired annotator labels.
  TP, FP, FN, and TN compare GPT-5.6 with human adjudication, treating sufficient evidence (label 1) as positive.}
  \label{tab:grounding_audit_counts}
  \small
  \setlength{\tabcolsep}{4pt}
  \renewcommand{\arraystretch}{1.12}
  \resizebox{0.6\linewidth}{!}{%
  \begin{tabular}{lrrrrrrrrr}
    \toprule
    \multirow{2}{*}{Model}
      & \multirow{2}{*}{$n$}
      & \multicolumn{4}{c}{Human A/B labels}
      & \multicolumn{4}{c}{GPT-5.6 vs.\ Human} \\
    \cmidrule(lr){3-6}
    \cmidrule(lr){7-10}
      & & 0/0 & 0/1 & 1/0 & 1/1
      & TP & FP & FN & TN \\
    \midrule
    Qwen3.5-9B + Code
      & 50 & 22 & 1 & 0 & 27
      & 26 & 0 & 2 & 22 \\
    V-Thinker
      & 50 & 22 & 0 & 1 & 27
      & 27 & 2 & 0 & 21 \\
    DeepEyes
      & 50 & 22 & 0 & 0 & 28
      & 28 & 1 & 0 & 21 \\
    DeepEyesV2
      & 50 & 23 & 0 & 0 & 27
      & 26 & 0 & 1 & 23 \\
    Mini-o3
      & 50 & 24 & 0 & 1 & 25
      & 25 & 1 & 1 & 23 \\
    \method
      & 50 & 11 & 0 & 0 & 39
      & 38 & 0 & 1 & 11 \\
    \midrule
    Pooled
      & 300 & 124 & 1 & 2 & 173
      & 170 & 4 & 5 & 121 \\
    \bottomrule
  \end{tabular}%
  }
\end{table}

\section{Task Catalogue and Worked Examples}
\label{app:task_taxonomy_full}

We group the tasks according to the primary evidence required to answer them.
Tasks in different groups may require overlapping operations.
A moving-average crossover requires line identification and temporal alignment as well as the financial relation itself.
Compositional tasks combine evidence across events, panels, or intervals.
The following inventory retains the task identifiers and group assignments.
\subsection{Task Inventory and Question Allocation}
\label{app:task_allocation}

The task inventory contains 16 visual primitives, 62 temporal and cross-panel tasks, 52 financial-pattern tasks, and 8 compositional tasks, for a total of 138 task types.
\Cref{tab:taxonomy_full} allocates the 1,200 questions nearly uniformly across individual task types, with 96 types receiving nine questions and 42 receiving eight.
The four groups therefore contain 132, 544, 454, and 70 questions, respectively.
The differences in group size reflect the number of task types in each group, rather than unequal sampling within a type.

The 77 task types covered by \trainset form a subset of the same 138-type taxonomy.
These types constitute the seen subset, while the remaining 61 are held out from all three training stages and constitute the unseen subset.
Each type belongs entirely to one subset.
\Cref{tab:training_task_coverage} summarizes this relationship.
The resulting seen and unseen subsets contain 667 and 533 questions, respectively.

\begin{table}[H]
\centering
\small
\caption{Question allocation across 138 task types, with eight or nine questions assigned to every type.}
\label{tab:taxonomy_full}
\setlength{\tabcolsep}{4pt}
\begin{tabular}{@{}lrrrrr@{}}
\toprule
Group & Types & 8-question types & 9-question types & Questions & Share (\%) \\
\midrule
Visual primitives & 16 & 12 & 4 & 132 & 11.00 \\
Temporal / cross-panel & 62 & 14 & 48 & 544 & 45.33 \\
Financial patterns & 52 & 14 & 38 & 454 & 37.83 \\
Compositional & 8 & 2 & 6 & 70 & 5.83 \\
\midrule
Total & 138 & 42 & 96 & 1,200 & 100.00 \\
\bottomrule
\end{tabular}
\end{table}
\FloatBarrier

\subsection{Complete Task-Type Catalogue}
\label{app:task_catalogue}

\Cref{tab:task_catalogue} gives formal display names while preserving the original identifiers.
Related identifiers are retained as separate types.
A question specifies any required window, threshold, indicator, or tie rule.

\begingroup
\small
\setlength{\tabcolsep}{5pt}
\renewcommand{\arraystretch}{1.06}
\setlength{\LTpre}{6pt}
\setlength{\LTpost}{8pt}
\begin{longtable}{@{}rP{0.79\textwidth}r@{}}
\caption{Complete inventory of 138 financial task types with question counts.}\label{tab:task_catalogue}\\
\toprule
ID & Formal task name & $n$ \\
\midrule
\endfirsthead
\multicolumn{3}{l}{\textit{Continued from the previous page}}\\
\toprule
ID & Formal task name & $n$ \\
\midrule
\endhead
\midrule
\multicolumn{3}{r}{\textit{Continued on next page}}\\
\endfoot
\bottomrule
\endlastfoot
\addlinespace[3pt]
\multicolumn{3}{@{}l}{\textbf{Visual Primitives}}\\*
\addlinespace[2pt]
001 & Candlestick body size estimation & 8 \\
002 & Candlestick direction change identification & 8 \\
003 & Candlestick counting & 8 \\
004 & Candlestick sequence identification & 8 \\
005 & Identification of the extremal moving-average line & 8 \\
006 & Localization of a price extremum & 8 \\
007 & Trading-session distance from a price extremum to the chart end & 9 \\
008 & Identification of the first candlestick & 9 \\
009 & Dating the first visible line observation & 8 \\
010 & Identification of the highest closing price & 8 \\
011 & Price change over the final n sessions & 8 \\
012 & Counting chart lines & 8 \\
013 & Localization of an MA5 turning point & 8 \\
014 & Ordering of moving-average lines & 8 \\
015 & Moving-average color identification & 9 \\
016 & Identification of a volume extremum & 9 \\
\addlinespace[3pt]
\multicolumn{3}{@{}l}{\textbf{Temporal and Cross-Panel Relations}}\\*
\addlinespace[2pt]
017 & Comparison of average trading volume between intervals & 8 \\
018 & Comparison of interval-end closing prices & 8 \\
019 & Comparison of interval high prices & 9 \\
020 & Comparison of mean MA20 values between intervals & 9 \\
021 & Comparison of mean MA5 values between intervals & 9 \\
022 & Comparison of MA5--MA20 spreads & 9 \\
023 & Comparison of maximum trading volume between intervals & 9 \\
024 & Comparison of interval price ranges & 9 \\
025 & Comparison of advancing-session counts & 8 \\
026 & Comparison of bullish-candlestick counts & 8 \\
027 & Counting sessions with bearish moving-average alignment & 9 \\
028 & Counting sessions with bullish moving-average alignment & 9 \\
029 & Counting closes above all specified moving averages & 9 \\
030 & Counting closes above MA5 & 9 \\
031 & Counting closes above the interval-start reference & 9 \\
032 & Counting closes below all specified moving averages & 9 \\
033 & Counting closes below MA5 & 8 \\
034 & Counting closes below the interval-start reference & 8 \\
035 & Counting declining sessions within an interval & 9 \\
036 & Counting price declines with volume expansion & 9 \\
037 & Counting price declines with volume contraction & 9 \\
038 & Dating the highest price within an interval & 9 \\
039 & Dating the largest candlestick body within an interval & 9 \\
040 & Dating the widest high--low range within an interval & 9 \\
041 & Counting local price highs within an interval & 9 \\
042 & Counting local price lows within an interval & 9 \\
043 & Longest sequence of declines with volume expansion & 8 \\
044 & Longest sequence of declining closes & 8 \\
045 & Longest sequence of advancing closes & 9 \\
046 & Longest sequence of advances with volume contraction & 9 \\
047 & Longest sequence of decreasing volume & 9 \\
048 & Longest sequence of increasing volume & 9 \\
049 & Longest sequence of bullish candlesticks & 9 \\
050 & Longest sequence of bearish candlesticks & 9 \\
051 & Counting qualifying lower shadows within an interval & 8 \\
052 & Dating the lowest price within an interval & 8 \\
053 & Counting sessions with a rising MA20 & 9 \\
054 & Counting sessions with MA5 above MA10 & 9 \\
055 & Counting sessions with MA5 below MA10 & 9 \\
056 & Counting MA5--MA10 bearish crossovers within an interval & 9 \\
057 & Counting MA5--MA10 bullish crossovers within an interval & 9 \\
058 & Counting sessions with a rising MA5 & 9 \\
059 & Dating the highest-volume declining session & 9 \\
060 & Dating the highest-volume advancing session & 9 \\
061 & Counting opens above the previous close & 8 \\
062 & Counting advancing sessions within an interval & 8 \\
063 & Counting price advances with volume expansion & 9 \\
064 & Counting price advances with volume contraction & 9 \\
065 & Counting qualifying upper shadows within an interval & 9 \\
066 & Counting sessions with decreasing volume & 9 \\
067 & Counting sessions with increasing volume & 9 \\
068 & Dating the maximum volume within an interval & 9 \\
069 & Dating the minimum volume within an interval & 8 \\
070 & Counting bullish candlesticks within an interval & 8 \\
071 & Counting bullish candlesticks with volume expansion & 9 \\
072 & Counting bearish candlesticks within an interval & 9 \\
073 & Counting bearish candlesticks with volume contraction & 9 \\
074 & Moving-average ordering on the final trading session & 9 \\
075 & Trading-session span from a peak to a subsequent trough & 9 \\
076 & Identification of price-segment direction sequences & 9 \\
077 & Classification of price-segment sequences & 9 \\
078 & Trading-session span from a trough to a subsequent peak & 9 \\
\addlinespace[3pt]
\multicolumn{3}{@{}l}{\textbf{Financial Patterns}}\\*
\addlinespace[2pt]
079 & Identification of consecutive price declines & 8 \\
080 & Identification of consecutive price advances & 8 \\
081 & Identification of consecutive bullish candlesticks & 9 \\
082 & Identification of consecutive bearish candlesticks & 9 \\
083 & Identification of bearish divergence & 9 \\
084 & Identification of bullish divergence & 9 \\
085 & Identification of a terminal high extremum & 8 \\
086 & Identification of a terminal low extremum & 8 \\
087 & Identification of price gaps & 9 \\
088 & Identification of downward price gaps & 9 \\
089 & Identification of upward price gaps & 9 \\
090 & Assessment of downward price deviation from MA20 & 9 \\
091 & Assessment of upward price deviation from MA20 & 9 \\
092 & Classification of final-session moving-average ordering & 9 \\
093 & Classification of the final five candlesticks & 8 \\
094 & Identification of the final five-session price direction & 8 \\
095 & Localization within the chart grid & 9 \\
096 & Duration of MA5 remaining above MA10 & 9 \\
097 & Duration of MA5 remaining above MA20 & 9 \\
098 & Identification of bearish moving-average alignment & 9 \\
099 & Identification of bullish moving-average alignment & 9 \\
100 & Duration of bearish moving-average alignment & 9 \\
101 & Duration of bullish moving-average alignment & 8 \\
102 & Identification of MA5--MA10 crossovers & 8 \\
103 & Identification of MA5--MA20 crossovers & 9 \\
104 & Identification of MA5--MA10 bearish crossovers & 9 \\
105 & Identification of MA5--MA10 bullish crossovers & 9 \\
106 & Identification of MA5--MA20 bearish crossovers & 9 \\
107 & Identification of MA5--MA20 bullish crossovers & 8 \\
108 & Dating the most recent MA5--MA10 bearish crossover & 8 \\
109 & Dating the most recent MA5--MA10 bullish crossover & 9 \\
110 & Dating the most recent MA5--MA20 bearish crossover & 9 \\
111 & Dating the most recent MA5--MA20 bullish crossover & 9 \\
112 & Localization of the maximum MA5 value & 9 \\
113 & Maximum price identification & 9 \\
114 & Longest sequence of same-color candlesticks & 9 \\
115 & Localization of the minimum MA5 value & 8 \\
116 & Minimum price identification & 8 \\
117 & Identification of a new price high & 9 \\
118 & Identification of a new price low & 9 \\
119 & Comparison of specified price observations & 9 \\
120 & Identification of a recent price high & 9 \\
121 & Identification of a recent price low & 9 \\
122 & Identification of lower-shadow patterns & 9 \\
123 & Identification of upper-shadow patterns & 8 \\
124 & Assessment of price volatility & 8 \\
125 & Identification of consecutive volume declines & 9 \\
126 & Identification of consecutive volume increases & 9 \\
127 & Maximum trading-volume identification & 9 \\
128 & Minimum trading-volume identification & 9 \\
129 & Bullish-candlestick counting & 9 \\
130 & Bearish-candlestick counting & 9 \\
\addlinespace[3pt]
\multicolumn{3}{@{}l}{\textbf{Compositional Financial Reasoning}}\\*
\addlinespace[2pt]
131 & Countertrend behavior under moving-average alignment & 8 \\
132 & Post-crossover price and volume follow-through & 8 \\
133 & Drawdown recovery with volume confirmation & 9 \\
134 & Volume confirmation of a new price high & 9 \\
135 & Post-pattern price follow-through & 9 \\
136 & Divergence between price and volume sequences & 9 \\
137 & Joint price, indicator, and volume divergence assessment & 9 \\
138 & Multi-step reasoning from a trough to a moving-average crossover & 9 \\
\end{longtable}
\endgroup

\subsection{Representative and Compositional Worked Examples}
\label{app:worked_examples}
The following six worked examples illustrate task semantics and required evidence using three charts.
The questions define their thresholds explicitly.
The reference observations accompany the compact charts to explain the task semantics.

\paragraph{Chart A: Local Geometry and Crossover Follow-Through.}
\begin{center}
\includegraphics[width=\textwidth]{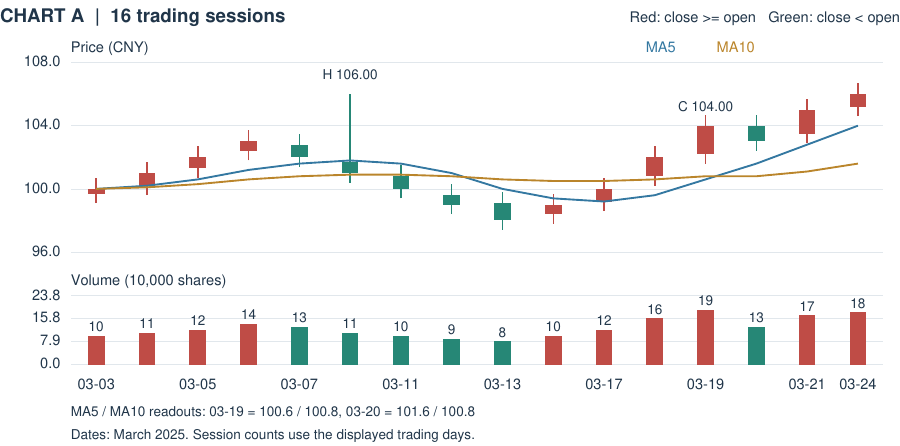}
\end{center}
\textbf{Example 1: Body Size versus Trading Range (Types 001, 039, 040).}
\textit{Question.}
Across all displayed sessions, which date has the largest candlestick body, and which has the largest high--low range?
Define body size as $\left|C-O\right|$ and range as $H-L$.
Return both dates and sizes.

\textit{Reference answer.}
The largest body occurs on March 19 and equals CNY 1.80.
The widest range occurs on March 10 and equals CNY 5.60.
\textit{Evidence.}
On March 19, $O=102.20$ and $C=104.00$.
On March 10, $H=106.00$ and $L=100.40$.
The long upper shadow on March 10 must not be counted as body size.
All displayed candles are compared before selecting either maximum.

\medskip
\textbf{Example 2: Crossover with Two-Session Confirmation (Type 132).}
\textit{Question.}
After March 12, find the first MA5 bullish crossover of MA10, using previous-session MA5 $\leq$ MA10 and current-session MA5 $>$ MA10.
Do both the following sessions close above MA5 and the crossover close, with volume above the mean of the three sessions immediately preceding the crossover?
Return the event date, decision, and close-to-close return through the second following session.

\textit{Reference answer.}
March 20, yes, and $2.91\%$.
\textit{Evidence.}
MA5/MA10 changes from $100.6/100.8$ on March 19 to $101.6/100.8$ on March 20.
The subsequent closes are 105 and 106, above MA5 values 102.8 and 104.0 and the event close of 103.
Volumes 17 and 18 exceed $\left(12+16+19\right)/3=15.67$.
The return is $\left(106/103-1\right)\times100$.
The volume on the event date is excluded from the reference window.

\paragraph{Chart B: Drawdown Recovery and Volume-Qualified Highs.}
\begin{center}
\includegraphics[width=\textwidth]{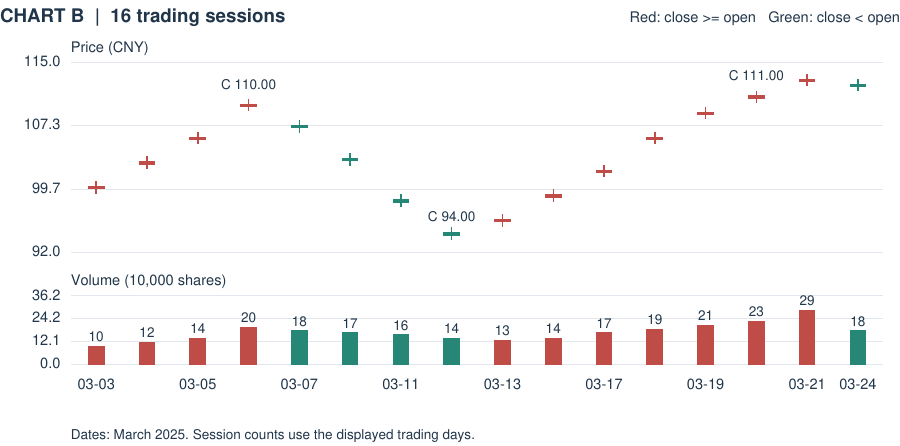}
\end{center}
\textbf{Example 3: Recovery after the Largest Drawdown (Type 133).}
\textit{Question.}
Within the first eight sessions, identify the largest peak-to-subsequent-trough drawdown based on closes.
Find the first later close that recovers the peak close.
Does recovery occur within six trading-session steps of the trough, with recovery-day volume at least $1.2$ times the mean of the preceding three sessions?
Return the drawdown, recovery date, elapsed sessions, and decision.

\textit{Reference answer.}
$14.55\%$, March 20, six sessions, and yes.
\textit{Evidence.}
The peak close is 110 on March 6 and the trough close is 94 on March 12.
The drawdown is $\left(110-94\right)/110\times100$.
March 19 closes at 109, while March 20 first recovers the peak at 111.
Recovery volume is 23, exceeding $1.2\times\left(17+19+21\right)/3=22.8$.
The elapsed interval is six trading steps, although the two calendar dates are eight days apart.

\medskip
\textbf{Example 4: A New High That Fails Volume Confirmation (Type 134).}
\textit{Question.}
Among the final four sessions, find the earliest session whose high strictly exceeds every earlier displayed high, whose close exceeds its open, and whose volume is at least $1.25$ times the preceding three-session mean.
A price breakout without volume confirmation does not qualify.

\textit{Reference answer.}
March 21.
\textit{Evidence.}
March 20 reaches a new high, but its volume of 23 is below $1.25\times19=23.75$.
March 21 reaches a further high of 113.60, has $C=113.00>O=112.60$, and records volume 29 against a threshold of $1.25\times\left(19+21+23\right)/3=26.25$.
Its volume ratio is $29/21=1.38$.
The first price breakout and the first fully qualified event therefore occur on different dates.

\paragraph{Chart C: Divergence Claims and Price--Volume Sequences.}
\begin{center}
\includegraphics[width=\textwidth]{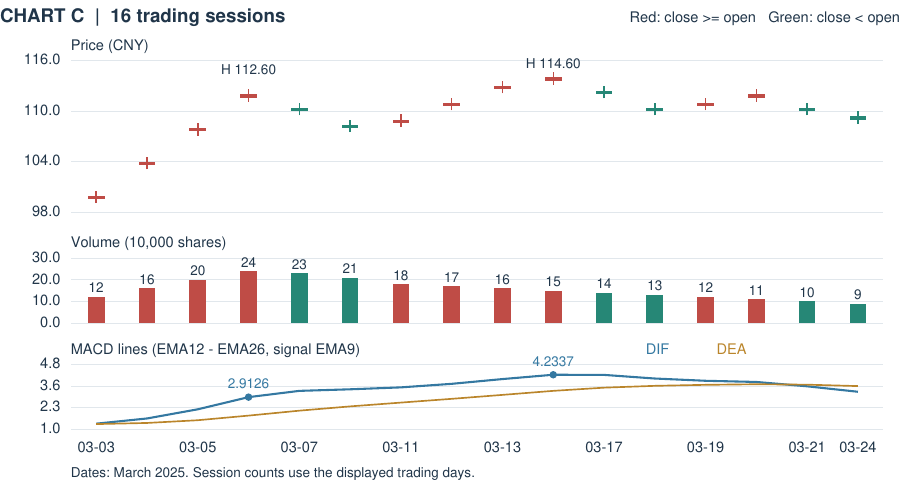}
\end{center}
\textbf{Example 5: Testing Three Conditions for Divergence (Type 137).}
\textit{Question.}
Compare the price highs on March 6 and March 14, each confirmed by a lower high on both adjacent displayed sessions.
Define joint bearish confirmation to require a higher second price high, a lower DIF value at that same date, and lower volume.
Return the three condition flags and their conjunction.

\textit{Reference answer.}
Price higher: true.
DIF lower: false.
Volume lower: true.
Joint confirmation: false.
\textit{Evidence.}
The highs increase from 112.60 to 114.60, and volume declines from 24 to 15.
However, DIF increases from 2.912578 to 4.233656.
Lower volume alone cannot establish the specified divergence.
DIF must be read at the two price-peak dates, rather than at unrelated indicator extrema.

\medskip
\textbf{Example 6: Rising Prices with Consecutively Contracting Volume (Type 136).}
\textit{Question.}
From March 11 through March 14, find the longest run in which every session closes above its predecessor while volume falls below its predecessor.
Use March 10 as the reference for the first comparison.
Return the run endpoints, number of qualifying sessions, and cumulative price and volume changes from the reference session to the run end.

\textit{Reference answer.}
March 11--14, four sessions, a price increase of $5.56\%$, and a volume decrease of $28.57\%$.
\textit{Evidence.}
Closes follow $108\rightarrow109\rightarrow111\rightarrow113\rightarrow114$, while volumes follow $21\rightarrow18\rightarrow17\rightarrow16\rightarrow15$.
There are four qualifying transitions and five observations including the reference.
The changes are $\left(114/108-1\right)\times100$ and $\left(15/21-1\right)\times100$.
Counting only transitions entirely inside the requested interval would incorrectly omit March 11.

\section{Trajectory Construction and Refinement}
\label{app:data_products}

We retain successful trajectories in two categories, $\mathcal{D}_{\mathrm{solve}}$ and $\mathcal{D}_{\mathrm{recover}}$, according to whether the student needed to correct an error.
The unsolved instances are assigned to $\mathcal{D}_{\mathrm{fail}}$ for tool improvement and subsequent collection attempts.
Every successful trajectory, including one obtained on a retry, undergoes the same cleaning, refinement, code expansion, and replay checks before it is added to the training corpus.

\begin{table}[H]
  \centering
  \caption{Outputs of the \trainset construction pipeline.}
  \label{tab:data_products}
  \small
  \begin{adjustbox}{max width=\textwidth}
  \begin{tabular}{lP{3.2cm}P{3.3cm}P{2.7cm}}
    \toprule
    Category & Source & Supervised target & Primary use \\
    \midrule
    $\mathcal{D}_{\mathrm{solve}}$ & Success without an intervening error & Cleaned solving trajectory & Efficient evidence acquisition \\
    $\mathcal{D}_{\mathrm{recover}}$ & Success after student self-correction & Error context, reflection, and correction & Recovery supervision \\
    $\mathcal{D}_{\mathrm{fail}}$ & Currently unsolved instance & No positive imitation target & Tool improvement and retry \\
    \bottomrule
  \end{tabular}
  \end{adjustbox}
\end{table}

\FloatBarrier
\subsection{Teacher-Assisted Collection and Trajectory Cleaning}
\label{app:collection_cleaning}

During offline collection, the student receives the question, chart images, and interaction history, but not the reference answer.
It can call the available financial tools or write Python code when those tools do not cover the required operation.
The tool arguments can be adjusted to the question and preceding observations.
The teacher has access to the reference answer and first diagnoses incorrect or uninformative proposals.
If the student remains unable to proceed, the teacher may suggest the objective or type of a useful next operation, without supplying exact arguments, executable code, or a completed solution.
The student executes its operations in the sandbox and observes the results.
At inference time, the trained model generates code using only the question, the input images, and its interaction history.

After the student produces a correct answer, we remove teacher messages and rejected proposals that were not executed.
The remaining trajectory contains student reasoning, executed actions, observations, and the final answer.
A solution reached without an intervening student error enters $\mathcal{D}_{\mathrm{solve}}$.
A solution reached through student reflection and correction enters $\mathcal{D}_{\mathrm{recover}}$, with the relevant error and corrective steps retained.
Both categories may have received teacher guidance during collection, but that guidance is excluded from the final training trajectories.

\subsection{Minimal-Sufficiency Refinement}
\label{app:min_sufficient}

After cleaning, we refine every successful trajectory to remove redundant inspections, unrelated computations, and abandoned branches.
This step applies to both direct and recovered solutions, including those obtained after tool updates and retries.
The reduced trajectory must still establish the correct answer from the input image and retain the context needed to understand any correction.

The refinement procedure comprises five steps.
\begin{enumerate}[leftmargin=*,itemsep=0em,topsep=0em,parsep=0em]
  \item \textbf{Identify the Supporting Evidence.} Locate the observations used to establish the answer, including their dates, chart panels, and analysis intervals.
  \item \textbf{Preserve Required Steps and Context.} Retain the operations that produce this evidence and the intermediate files or values on which it depends.
  For recovery trajectories, also retain the relevant student error, reflection, and correction.
  \item \textbf{Evaluate a Candidate Deletion.} Remove a repeated inspection, an unrelated computation, or an abandoned branch while preserving the order of the remaining steps.
  \item \textbf{Validate the Reduced Trajectory.} Accept the deletion only if the answer remains correct, the required evidence and its date, panel, and interval alignments are preserved, replay from the original image reproduces the decisive observations, and the recovery context remains intact.
  Restore the removed steps if any check fails.
  \item \textbf{Verify the Final Trajectory.} Repeat the deletion checks for the remaining candidates.
  Expand the retained tool calls into executable code, then run the answer audit and a final replay in an empty sandbox before adding the trajectory to the training corpus.
\end{enumerate}
In recovery data, the error and its correction are part of the learning signal and remain even when the final answer could be reproduced without them.

\paragraph{Example Reduction.}
Suppose the student inspects the same volume interval twice and also computes an unrelated price summary.
We can remove the duplicate inspection and summary if the remaining steps still support the requested volume comparison and reproduce all required intermediate results.
If the student initially inspected the wrong interval and then corrected it, we retain that error and correction as recovery context.

\subsection{Tool Updates and Re-Exploration}
\label{app:tool_evolution}

We inspect failed explorations from the training-data construction pool to identify operations that the current tools cannot perform reliably.
Tool development combines GPT-5.6 assistance with human review to add missing functionality or revise existing implementations.
Each update is tested for successful execution and the specified behavior before the failed instances are retried under the same teacher-assisted collection procedure.

The trajectories obtained from successful retries undergo the same cleaning and refinement procedure and enter $\mathcal{D}_{\mathrm{solve}}$ or $\mathcal{D}_{\mathrm{recover}}$ according to whether the student corrected an error.
The unsolved instances remain outside the training corpus and inform subsequent tool updates.
The library is fixed within each collection round and updated between rounds.

\paragraph{Separation from Benchmark Evaluation.}
The 1,200 \benchmark questions, their images and reference answers, and evaluation feedback are excluded from teacher-prompt design, failure analysis, tool development, training-data construction, and model selection.
This restriction holds across all collection rounds, including successful retries.
The model parameters and inference configurations are fixed before final evaluation.
The fixed policy may generate task-specific code during evaluation under the shared interaction protocol.

\FloatBarrier

\section{Corpus and Training Configuration}
\label{app:corpus_training_tables}

\subsection{Corpus Statistics}
The corpus contains 32,184 verified trajectories drawn from 18,742 unique source images, as one image can support multiple questions or successful explorations.
The direct-success, recovered-success, and failure rates in \Cref{tab:data_stats} describe the initial collection outcomes.
The final corpus also includes verified trajectories collected during retries.
The 96.8\% replay-pass rate is measured on candidates before final acceptance.
The retained trajectories have passed replay verification.

\begin{table}[H]
  \centering
  \caption{Statistics of \trainset.}
  \label{tab:data_stats}
  \small
  \begin{adjustbox}{max width=\textwidth}
  \begin{tabular}{P{5.0cm}cP{4.0cm}}
    \toprule
    Statistic & Value & Definition \\
    \midrule
    Promoted process-supervision samples & 32,184 & Replay-verified trajectories \\
    Training task types / unique images & 77 / 18,742 & Training coverage \\
    Images per trajectory & 3--4 & Source chart and intermediate visual results \\
    Intermediate images per trajectory & 2--3 & Visual results generated during execution \\
    Direct success / recovered success / failure rate & 61.7 / 30.4 / 7.9 & Outcomes of initial exploration \\
    Avg. interaction turns / code executions & 9.7 / 8.2 & Per promoted trajectory \\
    Candidate replay-pass rate & 96.8\% & Before final promotion \\
    \bottomrule
  \end{tabular}
  \end{adjustbox}
\end{table}

\FloatBarrier
\subsection{Supervised Targets and Recovery Context}
\label{app:sft_targets}

Stage~I uses 20,480 examples to develop basic visual-code interaction capabilities.
Stage~II uses 32,184 financial reasoning trajectories to train the model to combine these operations across dates and chart panels.
Each trajectory includes the source chart and two to three intermediate images, together with the code and observations needed to generate and reuse them.
This supervision trains the model to manage the sandbox workspace by maintaining the active image, retaining intermediate results, and referencing them in subsequent code actions.

\begin{table}[H]
  \centering
  \caption{Supervised training configuration.}
  \label{tab:training_config}
  \small
  \begin{tabular}{lcc}
    \toprule
    Configuration & Stage I & Stage II \\
    \midrule
    Initialization checkpoint & Qwen3.5-9B & Stage I checkpoint \\
    Maximum context & 32K & 32K \\
    Vision encoder & Frozen & Frozen \\
    Trainable modules & Language model + visual projector & Language model + visual projector \\
    Training samples & 20,480 & 32,184 \\
    Epochs & 1 & 1 \\
    Global batch size & 16 & 16 \\
    Peak learning rate & $1.0{\times}10^{-5}$ & $5.0{\times}10^{-6}$ \\
    Warmup ratio & 0.03 & 0.03 \\
    Numeric precision & bfloat16 (BF16) & BF16 \\
    \bottomrule
  \end{tabular}
\end{table}

We supervise student reasoning, code, and answer tokens.
The sandbox observations and retained failure context are provided as inputs and excluded from the loss.
For token $z_j$, let $m_j=1$ if it is a supervised target and $m_j=0$ otherwise.
The objective is defined as follows:
\begin{gather}
  \mathcal{L}_{\mathrm{SFT}}=-\frac{1}{\sum_j m_j}\sum_j m_j\log p_\theta\left(z_j\middle|\boldsymbol{I},q,z_{<j}\right).
\end{gather}
For a recovery trajectory, the model receives the earlier error as context and learns the reflection and corrective continuation.
The teacher messages and reference-answer inputs used during collection are removed.
The final answer of the student remains a supervised target.
The binary mask assigns equal weight to supervised tokens and zero weight to context tokens.

\subsection{Execution-Feedback Refinement}
\label{app:rl_interpretation}

In Stage~III, the policy is initialized from the final Stage~II checkpoint and optimized using GRPO for 45 steps before evaluation.
The training configuration is summarized in \Cref{tab:rl_config}.
We sample a group of four trajectories for each question.
Each trajectory permits at most twelve assistant generation turns, with up to 2,048 generated tokens per turn and a 32K-token limit for the complete multi-turn response.
At most one \texttt{<code>} block is extracted and executed per turn, giving at most twelve code actions per trajectory.

\begin{table}[H]
  \centering
  \caption{Stage~III reinforcement-learning configuration.}
  \label{tab:rl_config}
  \small
  \begin{tabular}{lc}
    \toprule
    Configuration & Value \\
    \midrule
    Base model & Qwen3.5-9B \\
    Initialization checkpoint & Final Stage~II checkpoint \\
    Optimization algorithm & GRPO \\
    Optimization steps before evaluation & 45 \\
    Rollouts per question & 4 \\
    Maximum assistant generation turns & 12 \\
    Maximum code actions per turn / rollout & 1 / 12 \\
    Maximum generated tokens per turn & 2,048 \\
    Multi-turn response token limit & 32K \\
    Advantage normalization & Within-question standardization \\
    Standard-deviation stabilizer & $10^{-6}$ \\
    Policy clipping $\epsilon$ / interval & 0.2 / $[0.8,1.2]$ \\
    Reward components / sum of weights & 6 / 1.00 \\
    KL estimator / coefficient & Low-variance / 0.02 (loss only) \\
    Entropy coefficient & 0 \\
    Critic / value loss & None \\
    Sandbox state semantics & Commit on success, unchanged on failure \\
    \bottomrule
  \end{tabular}
\end{table}

\paragraph{Trajectory Reward.}
For trajectory $i$, the total reward is the sum of six weighted components:
\begin{align}
\label{eq:app_rl_reward}
R_i
&=R_{\mathrm{code},i}
+R_{\mathrm{sandbox},i}
+R_{\mathrm{exec},i}
+R_{\mathrm{eff},i}
+R_{\mathrm{diff},i}
+R_{\mathrm{answer},i}.
\end{align}
\Cref{tab:grpo_reward_components} specifies the component meanings and scoring weights, which sum to 1.00.
The $R_{k,i}$ terms in the equation denote weighted reward contributions.
The code-validity reward $R_{\mathrm{code}}$ evaluates code format and Python syntax with a weight of 0.08.
The sandbox-success reward $R_{\mathrm{sandbox}}$ evaluates whether generated code executes successfully in the sandbox with a weight of 0.15.
The execution-efficiency reward $R_{\mathrm{exec}}$ has a weight of 0.10.
Let $a_i\in\left\{0,1\right\}$ indicate whether a correct answer is obtained in the sandbox.
The answer reward is $R_{\mathrm{answer},i}=0.55\,a_i$.
The training questions carry difficulty labels assigned during data annotation.
The corresponding difficulty coefficient $d_q$ is 0.1 for easy questions, 0.5 for moderate questions, and 1 for hard questions.
The difficulty reward is $R_{\mathrm{diff},i}=0.04\,d_q\,a_i$.
Correct answers receive difficulty contributions of 0.004, 0.020, and 0.040 at the three difficulty levels, respectively.
Incorrect answers receive no difficulty reward.

\paragraph{Interaction-Efficiency Reward.}
Let $T_i$ denote the number of assistant generation turns used by trajectory $i$.
The interaction-efficiency contribution is defined as follows:
\begin{align}
\label{eq:app_efficiency_reward}
R_{\mathrm{eff},i}
&=
\begin{cases}
0.08\left(1-\dfrac{T_i}{16}\right), & \text{if } a_i=1 \text{ and } 1\leq T_i\leq 12,\\
0, & \text{otherwise}.
\end{cases}
\end{align}
The normalization constant 16 controls the linear decay, while the generation budget remains twelve turns.
Correct answers after one, six, and twelve turns receive 0.075, 0.050, and 0.020, respectively.
The coefficient is 0.08.
The largest interaction-efficiency contribution for $T_i\geq1$ is 0.075.

\begin{table}[H]
  \centering
  \caption{Stage~III trajectory-reward components and scoring weights.
  The interaction-efficiency contribution follows \Cref{eq:app_efficiency_reward}.}
  \label{tab:grpo_reward_components}
  \small
  \begin{tabular}{llr}
    \toprule
    Component & Criterion & Weight \\
    \midrule
    $R_{\mathrm{code}}$ & Code format and Python syntax & 0.08 \\
    $R_{\mathrm{sandbox}}$ & Successful code execution in the sandbox & 0.15 \\
    $R_{\mathrm{exec}}$ & Code execution efficiency & 0.10 \\
    $R_{\mathrm{eff}}$ & Correct answers in fewer assistant generation turns & 0.08 \\
    $R_{\mathrm{diff}}$ & Correct answers weighted by annotated task difficulty & 0.04 \\
    $R_{\mathrm{answer}}$ & Correct answer obtained in the sandbox & 0.55 \\
    \midrule
    Total & & 1.00 \\
    \bottomrule
  \end{tabular}
\end{table}

\paragraph{Group-Relative Advantage and Policy Loss.}
Let $\bar R$ and $s_R$ be the mean and standard deviation of the four rewards for the same question.
Each trajectory uses the same scalar advantage across its generated tokens:
\begin{gather}
\label{eq:app_grpo_advantage}
\bar R=\frac{1}{4}\sum_{j=1}^{4}R_j,\qquad
s_R=\operatorname{std}(R_1,\ldots,R_4),\qquad
A_i=\frac{R_i-\bar R}{s_R+10^{-6}}.
\end{gather}
When all four rewards are identical, $A_i=0$ for every rollout.
The group is retained without resampling or exclusion.
Its clipped policy-gradient loss and reward-learning gradient are zero, while the independent KL term can still contribute gradients.
For generated token $z_{i,t}$ and interaction history $h_{i,t}$, we define the policy ratio as follows:
\begin{gather}
\label{eq:app_grpo_ratio}
\rho_{i,t}=\frac{\pi_\theta(z_{i,t}\mid h_{i,t})}{\pi_{\mathrm{old}}(z_{i,t}\mid h_{i,t})},
\end{gather}
where $\pi_{\mathrm{old}}$ is the policy used to sample the rollout.
The clipped policy objective and independent KL penalty give the following loss:
\begin{gather}
\label{eq:app_grpo_loss}
\begin{aligned}
\mathcal{L}_{\mathrm{GRPO}}={}&-\mathbb{E}_{i,t}\!\left[\min\!\left(\rho_{i,t}A_i,
\operatorname{clip}(\rho_{i,t},1-\epsilon,1+\epsilon)A_i\right)\right]\\
&+0.02\,D_{\mathrm{KL}}^{\mathrm{low\text{-}var}}(\pi_\theta\Vert\pi_{\mathrm{ref}}).
\end{aligned}
\end{gather}
Here $\epsilon=0.2$, so the clipped surrogate uses the ratio interval $[0.8,1.2]$.
The symbol $\pi_{\mathrm{ref}}$ denotes the reference policy, and the superscript denotes the low-variance KL estimator used in the implementation.
The KL term contributes to the loss and is excluded from $R_i$ and the reward-group statistics.
The entropy coefficient is zero.
GRPO uses the group-relative advantage without a critic or a value-loss term.

The corpus averages of 9.7 interaction turns and 8.2 code executions describe offline training trajectories rather than RL rollouts.
The sandbox commits state changes only after a successful action.
Failed actions leave the previous state intact, so subsequent steps cannot use partial results from a failed execution.

\FloatBarrier

\section{Additional Benchmark Results}
\label{app:finklinebench_results}

\subsection{Training Coverage and Task-Type Split}
\label{app:training_task_split}
The 77 task types in \trainset are a subset of the 138 task types in \benchmark and use the same task taxonomy.
These types constitute the seen subset.
The remaining 61 types are excluded from Stage~I and Stage~II supervised training and Stage~III reinforcement learning and constitute the unseen subset.
This distinction concerns task-specific training for \method.
All benchmark questions and images are excluded from all three training stages, including those belonging to seen task types.

\begin{table}[H]
  \centering
  \caption{Task-type coverage of training and evaluation, with all counts referring to task types.}
  \label{tab:training_task_coverage}
  \small
  \setlength{\tabcolsep}{5pt}
  \begin{tabularx}{\textwidth}{@{}lrX@{}}
    \toprule
    Task set & Types & Relationship to training \\
    \midrule
    Full benchmark & 138 & Complete evaluation task taxonomy \\
    Seen subset & 77 & Task types represented in \trainset \\
    Unseen subset & 61 & Task types held out from Stages~I--III \\
    \bottomrule
  \end{tabularx}
\end{table}

\FloatBarrier
\subsection{Public Interactive Models}
\label{app:public_interactive_models}
\Cref{tab:twi_current_diagnostics} repeats the category scores alongside model sizes.
Overall accuracy ranges from 1.2\% to 18.3\%, with zero compositional accuracy in this comparison.
Accuracy alone cannot distinguish errors in perception, code generation, execution, or financial reasoning.
The process metrics in \Cref{tab:process_core} illustrate how these aspects can be assessed separately.

\begin{table}[H]
  \centering
  \caption{Category scores for public interactive models repeated from \Cref{tab:main_results}, with group sizes of 132, 544, 454, and 70.}
  \label{tab:twi_current_diagnostics}
  \setlength{\tabcolsep}{3.2pt}
  \begin{adjustbox}{max width=\textwidth}
  \begin{tabular}{lcccccc}
    \toprule
    Model & Params & Visual Primitives & Temporal \& Relational & Financial Patterns & Compositional & Acc. \\
    \midrule
    V-Thinker & 7B & 43.9 & 13.2 & 12.8 & 0.0 & 15.7 \\
    DeepEyes & 7B & 44.7 & 18.8 & 13.0 & 0.0 & 18.3 \\
    DeepEyesV2 & 7B & 53.0 & 17.1 & 6.8 & 0.0 & 16.2 \\
    Mini-o3 & 7B & 31.1 & 15.6 & 10.1 & 0.0 & 14.3 \\
    ViGoRL & 3B & 22.0 & 6.1 & 8.4 & 0.0 & 8.3 \\
    VTool-R1 & 3B & 25.8 & 12.5 & 9.7 & 0.0 & 12.2 \\
    OpenThinkIMG & 2B & 3.8 & 0.0 & 2.0 & 0.0 & 1.2 \\
    \bottomrule
  \end{tabular}
  \end{adjustbox}
\end{table}

\paragraph{Evaluation Settings.}
\Cref{tab:evaluation_settings} summarizes the evaluated model groups, shared protocol, and reported generation budgets.
The comparison contains ten direct-answer baseline configurations, eight interactive baseline configurations, and the final \method model.
All configurations use the same 1,200 benchmark questions and answer grader.
Interactive configurations additionally share the parser and sandbox.
The budgets below describe evaluation settings.
The Stage~III training configuration is reported separately in \Cref{tab:rl_config}.

\begin{table}[H]
  \centering
  \caption{Model groups and reported evaluation settings.
  A dash indicates an unreported setting.}
  \label{tab:evaluation_settings}
  \setlength{\tabcolsep}{4pt}
  \renewcommand{\arraystretch}{1.14}
  \begin{tabularx}{\textwidth}{@{}P{0.24\textwidth}*{3}{>{\raggedright\arraybackslash}X}@{}}
    \toprule
    Setting & Direct-answer baselines & Interactive baselines & \method \\
    \midrule
    Configurations & 10 & 8 & 1 \\
    Model identity & \Cref{tab:main_results} & \Cref{tab:main_results} & Qwen3.5-9B after Stages~I--III \\
    Benchmark questions & 1,200 & 1,200 & 1,200 \\
    Answer grading & Shared grader & Shared grader & Shared grader \\
    Code execution & None & Shared parser and sandbox & Shared parser and sandbox \\
    Maximum generation turns & \dash & 12 & 12 \\
    Maximum code actions per turn & Not applicable & 1 & 1 \\
    Maximum generated tokens per turn & \dash & 2,048 & 2,048 \\
    Maximum total response tokens & \dash & 32K & 32K \\
    \bottomrule
  \end{tabularx}
  \vspace{2pt}
  \begin{minipage}{0.98\textwidth}
  \footnotesize
  The total response limit concerns generated tokens, not the full input context.
  \end{minipage}
\end{table}

\paragraph{Task Averaging and Generalization.}
For task type $k$ with $n_k$ questions and $a_k$ correct answers, task-macro accuracy is $\frac{100}{138}\sum_{k=1}^{138}a_k/n_k$.
The main accuracy is $100\sum_k a_k/1{,}200$.
The 77/61 split distinguishes seen task types from those held out throughout all three training stages.
For the final \method model, the seen and unseen subsets contain 369/667 (55.32\%) and 269/533 (50.47\%) correct responses, respectively.
Pooling the correct counts gives 638/1,200 (53.2\%), consistent with Tables~1 and~2 of the main paper.
The reported generalization results concern held-out task types.

\FloatBarrier

\Needspace{48\baselineskip}
\subsection{Inference System Prompt}
\label{app:inference_system_prompt}

The following template consolidates the inference instructions for \method.
The placeholder \texttt{\{workspace\_path\}} denotes the runtime workspace directory.

\par\smallskip
\begin{Verbatim}[fontsize=\fontsize{8}{9}\selectfont]
You are a visual reasoning agent with executable Python code and a persistent
sandbox. Given the question and available visual evidence, choose the shortest
reliable solution path rather than following a fixed tool routine.

Your workspace directory is {workspace_path}. Access the original task image
through input_path and keep it unchanged. Save intermediate images and other
artifacts in the workspace, using distinct filenames to preserve results
needed later. Reuse saved artifacts through their recorded paths and retain
variables required by subsequent operations.

Track the source image, chart panel, and trading date or interval associated
with each intermediate image or computed result. Before reusing an artifact,
verify that it corresponds to the intended part of the current analysis.

Use the original image directly when it provides sufficient evidence. When
cropping, zooming, annotation, pixel measurement, geometric analysis, or
deterministic computation is needed, state a short, verifiable goal in
<think>...</think>, then provide complete Python code in a single
<code>...</code> block.

Each execution must print RESULT_JSON_START, followed by a valid JSON object,
and then RESULT_JSON_END. Return generated visual evidence through
result_image. Successful sandbox state is preserved across steps. Failed
executions are rolled back, so do not rely on partial results from them.

Wait for the sandbox response before claiming that code executed successfully.
Only actual sandbox outputs count as evidence of execution.
Use the original chart as the primary source of visual evidence. Check that
extracted values and derived results correspond to the intended chart objects,
dates, intervals, and panels before using them. If the returned evidence is
insufficient or inconsistent, revise the analysis and continue.

Use annotations only when they help localization, comparison, counting, or
verification. Preserve the original visual content and avoid unnecessary
tool calls or duplicate artifacts.

For financial charts, ground all prices, volumes, indicators, dates, panel
assignments, and cross-panel relations in visible chart evidence or
computations based on that evidence. Do not infer answers from filenames,
metadata, task labels, or external market knowledge.

Return the final answer in <answer>...</answer> when the original image and
any required sandbox observations provide sufficient support. Follow the
answer format requested by the question. Otherwise, continue gathering or
verifying evidence.
\end{Verbatim}

\section{Additional Training and Robustness Analyses}
\label{app:experiments}

\subsection{Training Stages and Reward Components}
\label{app:training_reward_components}
\paragraph{QA and Textual-CoT Supervision Baselines.}
The QA and QA+CoT baselines in \Cref{tab:data_flywheel_core} use answer-only and textual-CoT versions of the same 32,184 question--image instances used for Stage~II process supervision in \trainset.
Both baselines are initialized directly from the original Qwen3.5-9B checkpoint and undergo SFT without Stage~I training or RL.
The Stage~II process-supervised model is initialized from the Stage~I checkpoint.
The full model then undergoes Stage~III RL.
All configurations are evaluated on the same 1,200 \benchmark questions with the common answer grader.
The main comparison uses the final \method model after all three training stages.
Its 53.2\% accuracy exceeds QA SFT (29.2\%) and QA+CoT SFT (34.6\%) by 24.0 and 18.6 percentage points, respectively.
The 36.4\% Stage~I and 49.1\% Stage~II checkpoints serve as intermediate results for the training-stage ablation.
The compared configurations share the base model and the financial training instances, while their training sequences, supervision formats, generated target lengths, and use of RL differ.

\begin{table}[H]
  \centering
  \caption{Training-stage results, an alternative SFT configuration, and reward ablations of the final model on \benchmark (\%).}
  \label{tab:curriculum_core}
  \small
  \setlength{\tabcolsep}{7pt}
  \renewcommand{\arraystretch}{1.12}
  \begin{tabularx}{\linewidth}{@{}Xrrr@{}}
    \toprule
    Training configuration & Acc. & ER & GAcc. \\
    \midrule
    \multicolumn{4}{@{}l}{\textbf{A. Cumulative training stages}} \\
    \addlinespace[2pt]
    Qwen3.5-9B (direct answer) & 23.6 & \dash & \dash \\
    Stage~I & 36.4 & 82.4 & 22.1 \\
    Stages~I + II & 49.1 & 85.2 & 38.9 \\
    Stages~I + II + III (\method, final) & \textbf{53.2} & 87.2 & \textbf{43.1} \\
    \midrule
    \multicolumn{4}{@{}l}{\textbf{B. Alternative SFT configuration}} \\
    \addlinespace[2pt]
    Single-stage complex SFT & 45.6 & 86.0 & 34.2 \\
    \midrule
    \multicolumn{4}{@{}l}{\textbf{C. Reward ablations of the final model}} \\
    \addlinespace[2pt]
Without execution-efficiency reward $R_{\mathrm{exec}}$ (0.10) & 50.4 & 73.6 & 38.8 \\
Without answer reward $R_{\mathrm{answer}}$ (0.55) & 43.2 & 89.7 & 35.0 \\
Without interaction-efficiency reward $R_{\mathrm{eff}}$ (0.08) & 51.0 & 89.1 & 39.6 \\
    \bottomrule
  \end{tabularx}
  \par\smallskip
  \begin{minipage}{\linewidth}
    Acc. denotes strict answer accuracy, ER denotes execution rate, and GAcc. denotes grounded accuracy.
    Group C ablates Stage~III rewards from the full three-stage configuration.
    Removing $R_{\mathrm{exec}}$ retains the separate sandbox-success reward $R_{\mathrm{sandbox}}$ with weight 0.15.
    Removing $R_{\mathrm{answer}}$ omits the reward for obtaining a correct answer in the sandbox with weight 0.55.
  \end{minipage}
\end{table}

\paragraph{Training Stages and Reward Ablations.}
As reported in \Cref{tab:curriculum_core}, strict accuracy increases from 36.4\% after Stage~I to 49.1\% after Stage~II and 53.2\% after Stage~III, with gains of 12.7 and 4.1 percentage points.
Grounded accuracy also rises, from 22.1\% to 38.9\% and then 43.1\%, indicating that more responses have correct answers and pass the GPT-5.6 grounding check.

The $-R_{\mathrm{exec}}$ configuration removes the execution-efficiency reward with weight 0.10 while retaining the sandbox-success reward with weight 0.15.
Its accuracy is 50.4\%, with an execution rate of 73.6\%.
The $-R_{\mathrm{eff}}$ configuration removes the interaction-efficiency reward with weight 0.08 defined in \Cref{eq:app_efficiency_reward} and reaches 51.0\% accuracy.
The $-R_{\mathrm{answer}}$ configuration removes the answer reward with weight 0.55 and has 43.2\% accuracy.
The single-stage complex SFT configuration achieves 45.6\%, compared with 49.1\% after Stage~II.

\FloatBarrier
\subsection{Interaction Cost}
The full model uses an average of 4.7 code turns, 3,980 tokens, and 14.3 seconds of end-to-end latency per question in \Cref{tab:efficiency}.
Its 4.1 successful executions per question correspond to an execution rate of approximately 87.2\%.
The Qwen3.5-9B base model succeeds on 0.6 of 5.8 attempted actions on average, or approximately 10.3\%.
The Stage~II execution rate is 85.2\%.

Without the interaction-efficiency reward, the model uses 6.4 turns, 5,210 tokens, and 18.1 seconds per question.
The full model uses 1.7 fewer turns and 1,230 fewer tokens, with 2.2 percentage points higher strict accuracy and 3.8 seconds lower mean latency.
End-to-end latency includes model generation and interaction overhead in addition to sandbox execution.
The latency comparison covers four Qwen3.5-9B configurations.

\begin{table}[H]
  \centering
  \caption{Reported inference efficiency, where code turns count attempted code actions and all counts are averaged per evaluation instance.
  The $-R_{\mathrm{eff}}$ variant omits the interaction-efficiency reward defined in \Cref{eq:app_efficiency_reward}.}
  \label{tab:efficiency}
  \small
  \begin{adjustbox}{max width=\textwidth}
  \begin{tabular}{lcccccc}
    \toprule
    Model & Code turns & Succ. exec. & ER (\%) & Tokens & Sandbox (s) & E2E latency (s) \\
    \midrule
    Qwen3.5-9B + Code & 5.8 & 0.6 & 10.3 & 4,820 & 7.4 & 16.9 \\
    Stage~II & 6.1 & 5.2 & 85.2 & 4,760 & 8.0 & 17.2 \\
    \method & 4.7 & 4.1 & 87.2 & 3,980 & 6.2 & 14.3 \\
    \quad $-R_{\mathrm{eff}}$ & 6.4 & 5.7 & 89.1 & 5,210 & 8.6 & 18.1 \\
    \bottomrule
  \end{tabular}
  \end{adjustbox}
\end{table}

\FloatBarrier
\subsection{Performance by Task Difficulty}
\label{sec:difficulty_analysis}

The easy, moderate, and hard difficulty labels were assigned by financial experts during benchmark annotation.
These labels provide an orthogonal view of model performance beyond the capability groups in \Cref{tab:main_results}, capturing the overall reasoning difficulty of individual questions.

\begin{table}[H]
  \centering
  \caption{Strict accuracy (\%) by expert-annotated question difficulty.}
  \label{tab:difficulty}
  \small
  \begin{tabular}{lccc}
    \toprule
    Model & Easy & Moderate & Hard \\
    \midrule
    Claude Opus 4.8      & 62.7 & 39.6 & 15.7 \\
    Qwen3.5-9B + Code    & 33.3 & 18.0 & 9.4  \\
    \method              & \textbf{72.8} & \textbf{50.5} & \textbf{36.9} \\
    \bottomrule
  \end{tabular}
\end{table}

Accuracy decreases with increasing difficulty for all three evaluated models.
\method achieves 72.8\%, 50.5\%, and 36.9\% strict accuracy on easy, moderate, and hard questions, respectively.
Claude Opus 4.8 reaches 62.7\%, 39.6\%, and 15.7\%, while the Qwen3.5-9B base model with sandbox access obtains 33.3\%, 18.0\%, and 9.4\%.

The difference between \method and Claude Opus 4.8 increases from 10.1 percentage points on easy questions to 10.9 points on moderate questions and 21.2 points on hard questions.
This pattern indicates that the performance difference is not confined to relatively simple chart-reading cases and becomes more pronounced on questions annotated as difficult.
Nevertheless, the 36.9\% accuracy of \method on the hard subset also shows that challenging financial chart reasoning remains far from solved.
\FloatBarrier

\FloatBarrier
\subsection{Qualitative Error and Trajectory Analysis}
\label{app:qualitative_errors}

The following analysis summarizes selected manually inspected baseline responses and tool trajectories.
The cases illustrate difficulties in connecting chart objects, trading dates, financial definitions, and dependent calculations.

\begin{table}[H]
  \centering
  \caption{Failure modes illustrated by selected manually inspected baseline responses.}
  \label{tab:qualitative_errors}
  \small
  \setlength{\tabcolsep}{4pt}
  \renewcommand{\arraystretch}{1.08}
  \begin{tabularx}{\linewidth}{@{}P{0.19\linewidth}P{0.26\linewidth}X@{}}
    \toprule
    Failure mode & Observed behavior & Inspected example \\
    \midrule
    Object and price-field confusion
    & Confusing closing prices with intraday extrema or candlestick components.
    & A Qwen3.5-9B response treated an annotated intraday high as the highest closing price. \\
    \addlinespace[3pt]
    Trading-date and interval misalignment
    & Treating sparse axis labels as a complete trading calendar or inserting non-trading dates.
    & DeepEyesV2 interpreted five axis labels as five trading days. An interactive 9B response included weekends and holidays in the sequence. \\
    \addlinespace[3pt]
    Price--volume relation errors
    & Substituting bar color for a magnitude comparison or failing to align price and volume by date.
    & DeepEyesV2 interpreted a green volume bar as evidence of increased volume without comparing bar heights. \\
    \addlinespace[3pt]
    Incorrect application of financial definitions
    & Replacing the required sequence of daily comparisons with a different visual criterion.
    & On a new-high counting question, a 9B response counted the single chart maximum and returned 1 instead of the reference answer of 12. \\
    \addlinespace[3pt]
    Propagation through dependent steps
    & Allowing an incorrect event location to determine subsequent searches and calculations.
    & Claude placed a left-side trough near the right edge of the chart and then concluded that no subsequent bullish crossover existed. \\
    \addlinespace[3pt]
    Answer-schema violations
    & Substituting dates for trading-day indices, changing field names, or omitting requested fields.
    & Some GPT-5.5 and Gemini multihop responses contained both schema and numerical errors. \\
    \bottomrule
  \end{tabularx}
\end{table}

\paragraph{Direct-Answer Responses.}
The inspected cases reveal errors in precise chart reading, temporal binding, and relation computation.
Some responses provide detailed daily explanations whose dates, values, or event locations lack support in the chart.
Responses containing only a final answer provide little detail for error analysis.

\paragraph{Interactive Trajectories.}
In addition to these shared errors, the inspected tool trajectories illustrate three ways in which execution can fail to advance the solution:
\begin{itemize}[leftmargin=*,itemsep=2pt,topsep=3pt,parsep=0pt]
  \item \textbf{Assumptions Encoded as Observations.} In a multihop response, DeepEyesV2 assigned assumed trough and crossover dates and a count of rising days to constants, then printed the result.
  The printed values reproduced these assumptions.
  \item \textbf{Repeated Inspection without Completion.} One trajectory executed six image-viewing actions successfully but exhausted its interaction budget without submitting an answer.
 \item \textbf{Failure to Initiate Code Execution.} In an inspected VTool-R1 response on an unseen task type, the model produced no executable code and therefore obtained no sandbox observations.
\end{itemize}



\section{Full Code-Interactive Trajectories}
\label{app:full_trajectories}

This appendix presents three cross-application candlestick reasoning cases and three reference trajectories.
The records document the questions, intermediate plans, executable code, sandbox observations, returned evidence images, and final answers.

\paragraph{Cross-Application Cases.}
The first part contains trajectories for Eastmoney, Futu NiuNiu, and Tongdaxin charts.
The cases involve conditional volume comparisons, candlestick direction, moving-average spreads, and indicator agreement.
Each record includes the input screenshot, the successive code interactions, and a final evidence review.

\paragraph{Reference Trajectories.}
The second part contains three reference trajectories covering interval-extrema reasoning, visual date transcription, and volume comparison.

\paragraph{Reading the Records.}
The executable code for all six trajectories is included directly from the accompanying Python source files in \texttt{supplementary/trajectory\_code}.
Line wrapping changes only the printed display.
The sandbox records are included directly from the accompanying JSON files.
Visual line wrapping preserves every field and value.
These records and their evidence images were regenerated by executing the accompanying code on the provided input images.
These artifacts document fixed-code replay.
For the date-transcription example, replay regenerates the title crop.
The final date is read visually from that crop.
\providecommand{\TraceAssetRoot}{figures}
\definecolor{traceviolet}{HTML}{5B43FF}
\definecolor{traceborder}{HTML}{C9D2E3}
\definecolor{tracecodebg}{HTML}{F6F8FB}
\definecolor{traceoutbg}{HTML}{F8F8FA}

\DefineVerbatimEnvironment{TraceCode}{Verbatim}{
  fontsize=\fontsize{7.5}{8.5}\selectfont,
    baselinestretch=1,
  breaklines=true,
  breakanywhere=true,
  frame=single,
  framesep=3pt,
  rulecolor=\color{traceborder},
  fillcolor=\color{tracecodebg}
}
\DefineVerbatimEnvironment{TraceOutput}{Verbatim}{
  fontsize=\fontsize{7.5}{8.5}\selectfont,
    baselinestretch=1,
  breaklines=true,
  breakanywhere=true,
  frame=single,
  framesep=3pt,
  rulecolor=\color{traceborder},
  fillcolor=\color{traceoutbg}
}
\newcommand{\traceplan}[1]{%
  \par\vspace{3pt}\noindent{\color{traceviolet}\bfseries PLAN}\quad #1\par\smallskip
}
\newcommand{\traceoutputheading}{%
  \par\Needspace{5\baselineskip}\vspace{3pt}\noindent{\color{traceviolet}\bfseries SANDBOX OUTPUT}\par\nopagebreak\vspace{3pt}
}
\newcommand{\traceimage}[3][0.94]{%
  \begin{figure}[H]
    \centering
    \includegraphics[width=#1\linewidth,height=0.48\textheight,keepaspectratio]{#2}
    \caption{#3}
  \end{figure}
}

\subsection{Cross-Application K-Line Traces}

The three cases below each contain seven executable code turns, separating perception, grid recovery, aligned visual review, per-day extraction, financial computation, and answer-level evidence.

\begin{table}[H]
\centering
\small
\begin{tabular}{lccp{0.48\linewidth}}
\toprule
Case & Code turns & Panels & Compound reasoning \\
\midrule
Eastmoney & 7 & 3 & MACD-sign filtering, volume maximum, and K-line direction \\
Futu NiuNiu & 7 & 2 & Two-interval moving-average spread and BIAS1 agreement \\
Tongdaxin & 7 & 3 & Interval volume maximum, selected K-line direction, and MACD count \\
\bottomrule
\end{tabular}
\caption{Structure of the three cross-application trajectories.
Each case returns two evidence images.}
\end{table}

\paragraph{Trace Contract.}
\begin{enumerate}[leftmargin=*,nosep]
\item Recover the time grid from visible periodic geometry; never set the period count from an answer or filename.
\item Preserve state across turns and reuse prior public variables instead of silently rereading the image.
\item Disclose the complete daily series used by each numerical decision.
\item Render aligned panels before extraction, then render a separate final image tied to the answer predicate.
\item Restrict final reasoning to the question, structured sandbox output, and returned images.
\end{enumerate}

\subsection{Case 1: Eastmoney}

\traceimage[0.42]{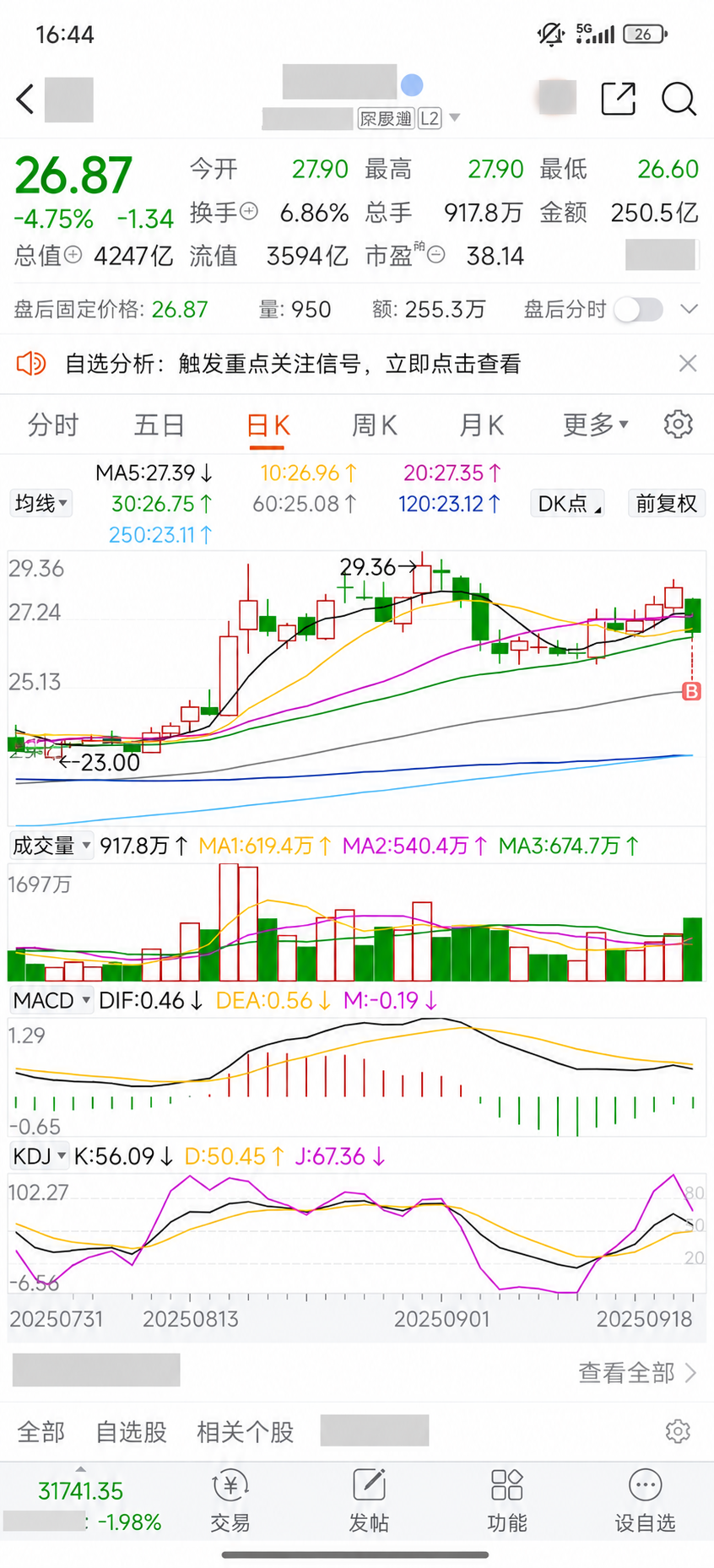}{Input chart for Case 1.}

\textbf{Public Screenshot Source:} \url{https://guba.eastmoney.com/news%2C300059%2C1602010335.html}

\paragraph{Question.}

Number the visible daily K-line periods from left to right. Within Days 8-29, consider only days whose MACD histogram is negative. Which eligible day has the highest trading-volume bar, and is that day's candlestick bullish or bearish? If volume is tied, choose the earliest day.

\paragraph{Replay Result.} \fbox{\textbf{Day 25; bearish}}

\subsubsection{Turn 1 of 7}

\traceplan{Locate the daily K-line, volume, and MACD panels and establish color masks without using any hidden market data.}

\Needspace{4\baselineskip}
\begin{TraceCode}
from PIL import Image
import numpy as np, json, re
input_path="......"
image = np.asarray(Image.open(input_path).convert("RGB"))
h, w = image.shape[:2]
rgb = image.astype(np.int16)
r, g, b = rgb[:, :, 0], rgb[:, :, 1], rgb[:, :, 2]
red = (r > 145) & (r - g > 45) & (r - b > 35)
green = (g > 90) & (g - r > 35) & (g - b > 25)
price_bbox = [0, int(round(0.258*h)), w, int(round(0.490*h))]
volume_bbox = [0, int(round(0.522*h)), w, int(round(0.623*h))]
macd_bbox = [0, int(round(0.653*h)), w, int(round(0.746*h))]
print("RESULT_JSON_START")
print(json.dumps({"stage":"inspect_layout","task":"conditional_volume_extreme","image_size":[w,h],"price_bbox":price_bbox,"volume_bbox":volume_bbox,"macd_bbox":
   macd_bbox,"panels":["daily_kline","volume","MACD","KDJ"]}, separators=(",", ":")))
print("RESULT_JSON_END")
\end{TraceCode}

\traceoutputheading

\Needspace{4\baselineskip}
\begin{TraceOutput}
{"stage": "inspect_layout", "task": "conditional_volume_extreme", "image_size": [1220, 2712], "price_bbox":
[0, 700, 1220, 1329], "volume_bbox": [0, 1416, 1220, 1690], "macd_bbox": [0, 1771, 1220, 2023], "panels":
["daily_kline", "volume", "MACD", "KDJ"]}
\end{TraceOutput}

\subsubsection{Turn 2 of 7}

\traceplan{Recover the complete daily grid from the regularly spaced MACD histogram bars rather than hard-coding a period count.}

\Needspace{4\baselineskip}
\begin{TraceCode}
import numpy as np, json
mx1, my1, mx2, my2 = macd_bbox
profile = (red[my1:my2] | green[my1:my2]).sum(axis=0)
active = np.where(profile >= 3)[0]
runs = []
if active.size:
  start = previous = int(active[0])
  for value in active[1:]:
    value = int(value)
    if value <= previous + 3:
     previous = value
    else:
     runs.append((start, previous)); start = previous = value
  runs.append((start, previous))
candidate_centers = [0.5*(a+b) for a,b in runs if 2 <= b-a+1 <= 8 and 5 < 0.5*(a+b) < w-5]
diffs = [b-a for a,b in zip(candidate_centers, candidate_centers[1:]) if 20 <= b-a <= 45]
if len(diffs) < 10:
  raise ValueError("insufficient MACD bars for grid recovery")
pitch = float(np.median(diffs))
first = float(min(candidate_centers)); last = float(max(candidate_centers))
day_count = int(round((last-first)/pitch)) + 1
centers = np.linspace(first, last, day_count)
if not 30 <= day_count <= 45:
  raise ValueError("recovered day count outside visible chart bounds")
print("RESULT_JSON_START")
print(json.dumps({"stage":"recover_grid","task":"conditional_volume_extreme","grid_source":"MACD histogram centers","day_count":day_count,"pitch_px":round(pitch,3),
   "first_center":round(first,2),"last_center":round(last,2),"centers":[round(float(x),2) for x in centers]}, separators=(",", ":")))
print("RESULT_JSON_END")
\end{TraceCode}

\traceoutputheading

\Needspace{4\baselineskip}
\begin{TraceOutput}
{"stage": "recover_grid", "task": "conditional_volume_extreme", "grid_source": "MACD histogram centers",
"day_count": 36, "pitch_px": 33.0, "first_center": 27.5, "last_center": 1182.5, "centers": [27.5, 60.5,
93.5, 126.5, 159.5, 192.5, 225.5, 258.5, 291.5, 324.5, 357.5, 390.5, 423.5, 456.5, 489.5, 522.5, 555.5,
588.5, 621.5, 654.5, 687.5, 720.5, 753.5, 786.5, 819.5, 852.5, 885.5, 918.5, 951.5, 984.5, 1017.5, 1050.5,
1083.5, 1116.5, 1149.5, 1182.5]}
\end{TraceOutput}

\subsubsection{Turn 3 of 7}

\traceplan{Render the three aligned panels with the recovered grid and the D8-D29 target interval for visual review.}

\Needspace{4\baselineskip}
\begin{TraceCode}
from PIL import Image, ImageDraw
import numpy as np, json
source = Image.fromarray(image.astype(np.uint8)).convert("RGB")
sections = [(price_bbox, 300, "DAILY K-LINE"), (volume_bbox, 165, "VOLUME"), (macd_bbox, 165, "MACD")]
canvas = Image.new("RGB", (960, 690), "white")
draw = ImageDraw.Draw(canvas)
yoff = 55
for bbox, target_h, label in sections:
  panel = source.crop(tuple(bbox)).resize((960, target_h))
  canvas.paste(panel, (0, yoff))
  draw.text((8, yoff+6), label, fill=(25,35,55))
  yoff += target_h
for day, x in enumerate(centers, 1):
  xx = int(round(x / w * 960))
  color = (116,92,245) if day in (8,29) else (190,195,210)
  width = 3 if day in (8,29) else 1
  draw.line((xx,55,xx,685), fill=color, width=width)
  if day == 1 or day % 5 == 0 or day in (8,29):
    draw.text((max(0,xx-10),35), "D%d" % day, fill=(40,45,60))
draw.text((12,10), "Recovered %d-day grid; target interval D8-D29" % day_count, fill=(25,35,55))
result_image = canvas
print("RESULT_JSON_START")
print(json.dumps({"stage":"render_aligned_focus","task":"conditional_volume_extreme","day_count":day_count,"target_interval":[8,29],"panels":["price","volume","MACD"],"result_image_size":list(result_image.size)}, separators=(",", ":")))
print("RESULT_JSON_END")
\end{TraceCode}

\traceoutputheading

\Needspace{4\baselineskip}
\begin{TraceOutput}
{"stage": "render_aligned_focus", "task": "conditional_volume_extreme", "day_count": 36, "target_interval":
[8, 29], "panels": ["price", "volume", "MACD"], "result_image_size": [960, 690]}
\end{TraceOutput}

\traceimage[0.7]{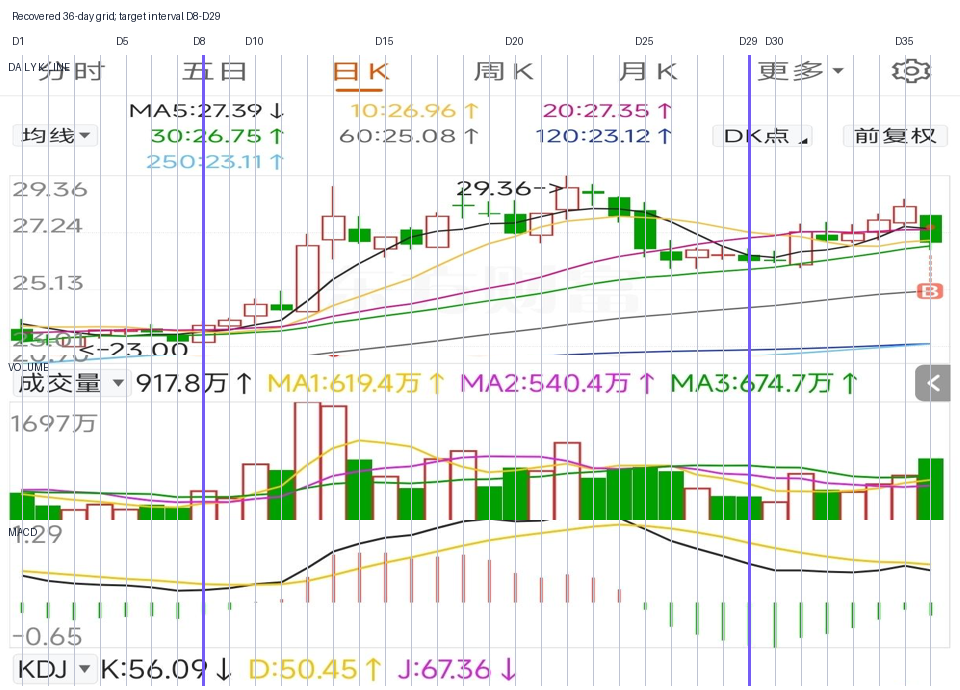}{Aligned-panel working image returned after grid recovery.}

\subsubsection{Turn 4 of 7}

\traceplan{Measure the volume-bar height for every recovered day and disclose the complete daily sequence.}

\Needspace{4\baselineskip}
\begin{TraceCode}
import numpy as np, json
vx1, vy1, vx2, vy2 = volume_bbox
volume_heights = []
volume_runs = []
for x in centers:
  xx = int(round(x))
  rows = np.where((red[vy1:vy2,max(0,xx-10):min(w,xx+11)] | green[vy1:vy2,max(0,xx-10):min(w,xx+11)]).sum(axis=1) >= 2)[0]
  local_runs = []
  if rows.size:
    start = previous = int(rows[0])
    for value in rows[1:]:
     value = int(value)
     if value <= previous + 3:
       previous = value
     else:
       local_runs.append((start,previous)); start = previous = value
    local_runs.append((start,previous))
  candidates = [pair for pair in local_runs if pair[1]-pair[0]+1 >= 5]
  selected = max(candidates, key=lambda pair:(pair[1],pair[1]-pair[0])) if candidates else (0,0)
  volume_runs.append([int(selected[0]+vy1),int(selected[1]+vy1)])
  volume_heights.append(int(selected[1]-selected[0]+1))
print("RESULT_JSON_START")
print(json.dumps({"stage":"extract_volume","task":"conditional_volume_extreme","day_count":day_count,"volume_heights_px":volume_heights,"volume_runs":volume_runs},
   separators=(",", ":")))
print("RESULT_JSON_END")
\end{TraceCode}

\traceoutputheading

\Needspace{4\baselineskip}
\begin{TraceOutput}
{"stage": "extract_volume", "task": "conditional_volume_extreme", "day_count": 36, "volume_heights_px":
[48, 25, 1, 5, 6, 26, 21, 5, 15, 5, 84, 9, 8, 102, 6, 54, 7, 6, 57, 88, 7, 8, 72, 89, 89, 82, 5, 41,
40, 5, 8, 48, 5, 9, 1, 104], "volume_runs": [[1642, 1689], [1665, 1689], [1416, 1416], [1661, 1665],
[1643, 1648], [1664, 1689], [1669, 1689], [1659, 1663], [1648, 1662], [1646, 1650], [1606, 1689], [1633,
1641], [1626, 1633], [1588, 1689], [1625, 1630], [1636, 1689], [1623, 1629], [1620, 1625], [1633, 1689],
[1602, 1689], [1614, 1620], [1609, 1616], [1618, 1689], [1601, 1689], [1601, 1689], [1608, 1689], [1597,
1601], [1649, 1689], [1650, 1689], [1600, 1604], [1600, 1607], [1642, 1689], [1640, 1644], [1629, 1637],
[1416, 1416], [1586, 1689]]}
\end{TraceOutput}

\subsubsection{Turn 5 of 7}

\traceplan{Classify the MACD histogram sign at every day center and disclose the full sign sequence.}

\Needspace{4\baselineskip}
\begin{TraceCode}
import numpy as np, json
mx1, my1, mx2, my2 = macd_bbox
macd_signs = []
macd_color_counts = []
for x in centers:
  xx = int(round(x))
  red_count = int(red[my1:my2,max(0,xx-5):min(w,xx+6)].sum())
  green_count = int(green[my1:my2,max(0,xx-5):min(w,xx+6)].sum())
  macd_signs.append("negative" if green_count > red_count else "positive")
  macd_color_counts.append([red_count,green_count])
print("RESULT_JSON_START")
print(json.dumps({"stage":"extract_macd_sign","task":"conditional_volume_extreme","macd_signs":macd_signs,"red_green_counts":macd_color_counts}, separators=(",", ":")))
print("RESULT_JSON_END")
\end{TraceCode}

\traceoutputheading

\Needspace{4\baselineskip}
\begin{TraceOutput}
{"stage": "extract_macd_sign", "task": "conditional_volume_extreme", "macd_signs": ["negative", "negative",
"negative", "negative", "negative", "negative", "negative", "negative", "negative", "positive", "positive",
"positive", "positive", "positive", "positive", "positive", "positive", "positive", "positive", "positive",
"positive", "positive", "positive", "positive", "negative", "negative", "negative", "negative", "negative",
"negative", "negative", "negative", "negative", "negative", "negative", "negative"], "red_green_counts":
[[0, 71], [0, 109], [0, 128], [0, 107], [0, 122], [0, 95], [0, 115], [0, 79], [0, 46], [0, 0], [0, 0],
[109, 0], [271, 0], [262, 0], [271, 0], [226, 0], [250, 0], [199, 0], [175, 0], [72, 0], [132, 0], [78,
0], [114, 0], [37, 0], [0, 46], [0, 184], [0, 215], [0, 275], [0, 384], [0, 386], [0, 231], [0, 235],
[0, 206], [0, 118], [0, 48], [0, 93]]}
\end{TraceOutput}

\subsubsection{Turn 6 of 7}

\traceplan{Filter D8-D29 to negative-MACD days, compare all eligible volume heights, and apply the earliest-day tie rule.}

\Needspace{4\baselineskip}
\begin{TraceCode}
import json
start_day, end_day = 8, 29
eligible = [{"day":day,"volume_height_px":volume_heights[day-1]} for day in range(start_day,end_day+1) if macd_signs[day-1] == "negative"]
if not eligible:
  raise ValueError("no negative-MACD candidate in target interval")
selected_candidate = max(eligible, key=lambda row:(row["volume_height_px"],-row["day"]))
selected_day = int(selected_candidate["day"])
print("RESULT_JSON_START")
print(json.dumps({"stage":"compute_finance","task":"conditional_volume_extreme","interval":[start_day,end_day],"predicate":"MACD histogram is negative","eligible_candidates":eligible,"tie_break":"earliest day","selected_day":selected_day,"selected_volume_height_px":selected_candidate["volume_height_px"]},
   separators=(",", ":")))
print("RESULT_JSON_END")
\end{TraceCode}

\traceoutputheading

\Needspace{4\baselineskip}
\begin{TraceOutput}
{"stage": "compute_finance", "task": "conditional_volume_extreme", "interval": [8, 29], "predicate":
"MACD histogram is negative", "eligible_candidates": [{"day": 8, "volume_height_px": 5}, {"day": 9, "volume_height_px":
15}, {"day": 25, "volume_height_px": 89}, {"day": 26, "volume_height_px": 82}, {"day": 27, "volume_height_px":
5}, {"day": 28, "volume_height_px": 41}, {"day": 29, "volume_height_px": 40}], "tie_break": "earliest day",
"selected_day": 25, "selected_volume_height_px": 89}
\end{TraceOutput}

\subsubsection{Turn 7 of 7}

\traceplan{Classify the selected day's K-line direction and render answer-level evidence across price, volume, and MACD.}

\Needspace{4\baselineskip}
\begin{TraceCode}
from PIL import Image, ImageDraw
import numpy as np, json
sx = int(round(centers[selected_day-1]))
px1, py1, px2, py2 = price_bbox
red_pixels = int(red[py1:py2,max(0,sx-10):min(w,sx+11)].sum())
green_pixels = int(green[py1:py2,max(0,sx-10):min(w,sx+11)].sum())
candle_direction = "bearish" if green_pixels > red_pixels else "bullish"
x_left = max(0,int(round(centers[7]-0.55*pitch)))
x_right = min(w,int(round(centers[28]+0.55*pitch)))
source = Image.fromarray(image.astype(np.uint8)).convert("RGB")
panel_specs = [(price_bbox,320,"PRICE"),(volume_bbox,180,"VOLUME"),(macd_bbox,180,"MACD")]
canvas = Image.new("RGB",(980,745),"white"); draw = ImageDraw.Draw(canvas)
draw.text((14,12),"D%d | volume=%d px | MACD negative | %s candle" % (selected_day,volume_heights[selected_day-1],candle_direction.upper()),fill=(20,30,50))
yoff=55
for bbox,target_h,label in panel_specs:
  crop = source.crop((x_left,bbox[1],x_right,bbox[3])).resize((980,target_h))
  canvas.paste(crop,(0,yoff)); draw.text((8,yoff+5),label,fill=(20,30,50))
  mark_x=int(round((sx-x_left)/max(x_right-x_left,1)*980))
  draw.rectangle((max(0,mark_x-11),yoff,min(979,mark_x+11),yoff+target_h-1),outline=(255,0,180),width=4)
  yoff += target_h
result_image = canvas
answer = "Day %d; %s" % (selected_day,candle_direction)
print("RESULT_JSON_START")
print(json.dumps({"stage":"render_answer_evidence","task":"conditional_volume_extreme","answer":answer,"selected_day":selected_day,"volume_height_px":volume_heights
   [selected_day-1],"macd_sign":macd_signs[selected_day-1],"candle_direction":candle_direction,"candle_red_green_pixels":[red_pixels,green_pixels],"checked_interval":[8,29],"result_image_size":list(result_image.size)}, separators=(",", ":")))
print("RESULT_JSON_END")
\end{TraceCode}

\traceoutputheading

\Needspace{4\baselineskip}
\begin{TraceOutput}
{"stage": "render_answer_evidence", "task": "conditional_volume_extreme", "answer": "Day 25; bearish",
"selected_day": 25, "volume_height_px": 89, "macd_sign": "negative", "candle_direction": "bearish", "candle_red_green_pixels":
[48, 1896], "checked_interval": [8, 29], "result_image_size": [980, 745]}
\end{TraceOutput}

\traceimage[0.7]{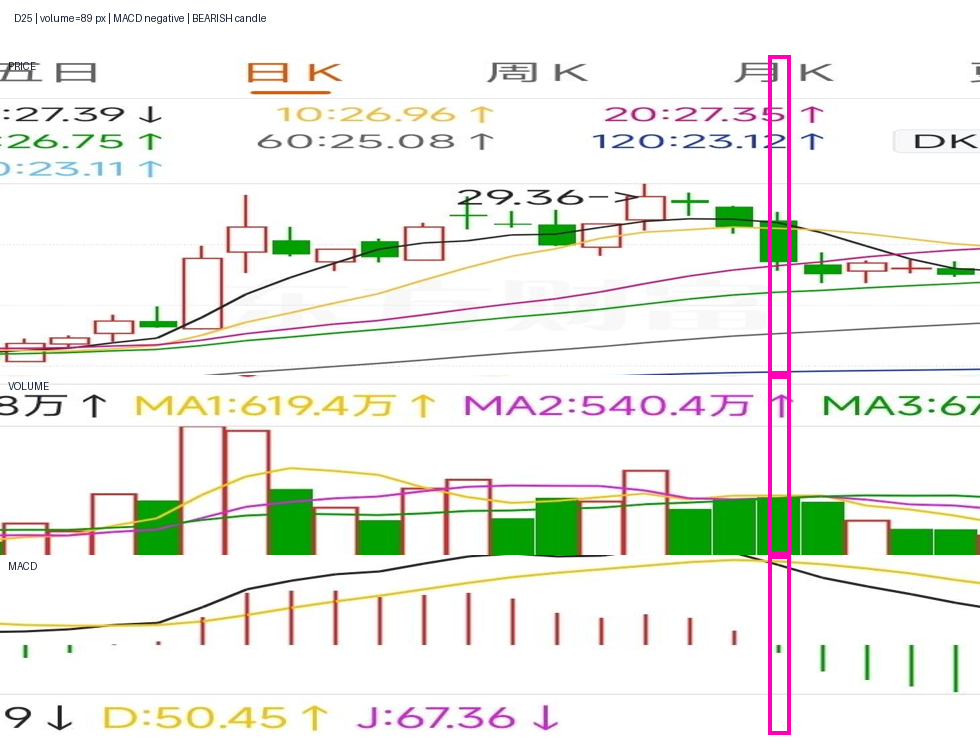}{Answer-level evidence returned after the complete financial computation.}

\subsubsection{Final Evidence Review and Answer}

The final turn reviews the returned evidence against the disclosed per-period values.

\begin{center}\Large\bfseries Day 25; bearish\end{center}

\subsection{Case 2: Futu NiuNiu}

\traceimage[0.88]{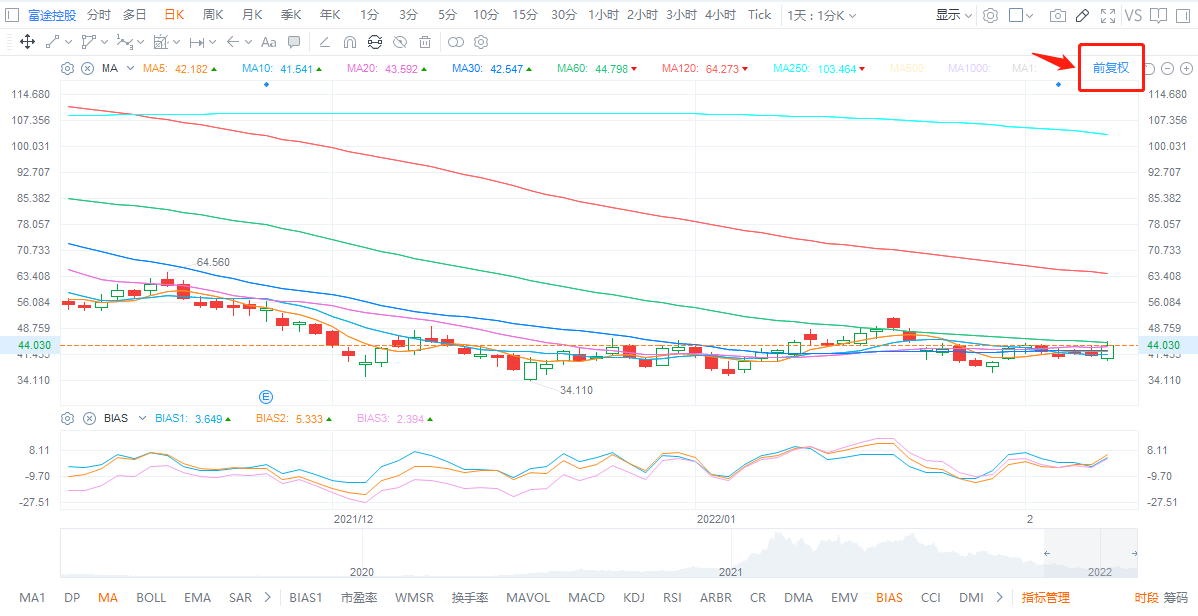}{Input chart for Case 2.}

\textbf{Public Screenshot Source:} \url{https://support.futunn.com/topic58}

\paragraph{Question.}

Number the visible daily candlesticks from left to right. Compare Days 10-25 (first interval) with Days 40-55 (second interval). Which interval has the larger average absolute vertical separation between MA5 and MA20? Does that same interval also have the higher average BIAS1 position? Compare values by their visible chart positions.

\paragraph{Replay Result.} \fbox{\textbf{first interval; BIAS1 agreement: no}}

\subsubsection{Turn 1 of 7}

\traceplan{Locate the main K-line panel and the aligned BIAS subpanel, and identify the visible indicator families.}

\Needspace{4\baselineskip}
\begin{TraceCode}
from PIL import Image
import numpy as np, json, re
image = np.asarray(Image.open(input_path).convert("RGB"))
h, w = image.shape[:2]
rgb = image.astype(np.int16); r,g,b = rgb[:,:,0],rgb[:,:,1],rgb[:,:,2]
red_candle = (r > 180) & (r-g > 55) & (r-b > 35)
green_candle = (g > 80) & (g-r > 25) & (g-b > 15)
main_bbox = [int(round(0.050*w)),int(round(0.422*h)),int(round(0.950*w)),int(round(0.662*h))]
bias_bbox = [int(round(0.050*w)),int(round(0.698*h)),int(round(0.950*w)),int(round(0.827*h))]
print("RESULT_JSON_START")
print(json.dumps({"stage":"inspect_layout","task":"compare_ma_gap_and_bias","image_size":[w,h],"main_bbox":main_bbox,"bias_bbox":bias_bbox,"visible_series":["MA5","MA10","MA20","MA30","MA60","MA120","MA250","BIAS1","BIAS2","BIAS3"]}, separators=(",", ":")))
print("RESULT_JSON_END")
\end{TraceCode}

\traceoutputheading

\Needspace{4\baselineskip}
\begin{TraceOutput}
{"stage": "inspect_layout", "task": "compare_ma_gap_and_bias", "image_size": [1198, 616], "main_bbox":
[60, 260, 1138, 408], "bias_bbox": [60, 430, 1138, 509], "visible_series": ["MA5", "MA10", "MA20", "MA30",
"MA60", "MA120", "MA250", "BIAS1", "BIAS2", "BIAS3"]}
\end{TraceOutput}

\subsubsection{Turn 2 of 7}

\traceplan{Recover the complete day grid by optimizing the periodic candle-column response, without fixing the day count in advance.}

\Needspace{4\baselineskip}
\begin{TraceCode}
import numpy as np, json
x1,y1,x2,y2 = main_bbox
column_profile = (red_candle[y1:y2] | green_candle[y1:y2]).sum(axis=0).astype(float)
pitch_candidates = np.arange(max(12.0,0.012*w), min(22.0,0.019*w)+0.01, 0.25)
best = None
for candidate_pitch in pitch_candidates:
  for phase in np.arange(0.0,candidate_pitch,0.25):
    trial = np.arange(x1+phase,x2,candidate_pitch)
    scores=[]
    for value in trial:
     xx=int(round(value)); scores.append(float(column_profile[max(0,xx-1):min(w,xx+2)].max()))
    score=float(np.mean(scores)+0.15*np.percentile(scores,75))
    if best is None or score>best[0]: best=(score,float(candidate_pitch),float(phase),len(trial))
_, pitch, phase, _ = best
centers = np.arange(x1+phase,x2,pitch)
day_count = int(len(centers))
if not 45 <= day_count <= 80:
  raise ValueError("recovered day count outside visible chart bounds")
print("RESULT_JSON_START")
print(json.dumps({"stage":"recover_grid","task":"compare_ma_gap_and_bias","method":"periodic candle-column score","day_count":day_count,"pitch_px":round(pitch,3),"phase_px":round(phase,3),"centers":[round(float(x),2) for x in centers]}, separators=(",", ":")))
print("RESULT_JSON_END")
\end{TraceCode}

\traceoutputheading

\Needspace{4\baselineskip}
\begin{TraceOutput}
{"stage": "recover_grid", "task": "compare_ma_gap_and_bias", "method": "periodic candle-column score",
"day_count": 66, "pitch_px": 16.376, "phase_px": 9.25, "centers": [69.25, 85.63, 102.0, 118.38, 134.75,
151.13, 167.51, 183.88, 200.26, 216.63, 233.01, 249.39, 265.76, 282.14, 298.51, 314.89, 331.27, 347.64,
364.02, 380.39, 396.77, 413.15, 429.52, 445.9, 462.27, 478.65, 495.03, 511.4, 527.78, 544.15, 560.53,
576.91, 593.28, 609.66, 626.03, 642.41, 658.79, 675.16, 691.54, 707.91, 724.29, 740.67, 757.04, 773.42,
789.79, 806.17, 822.55, 838.92, 855.3, 871.67, 888.05, 904.43, 920.8, 937.18, 953.55, 969.93, 986.31,
1002.68, 1019.06, 1035.43, 1051.81, 1068.19, 1084.56, 1100.94, 1117.31, 1133.69]}
\end{TraceOutput}

\subsubsection{Turn 3 of 7}

\traceplan{Render both aligned panels with the two comparison intervals marked for review.}

\Needspace{4\baselineskip}
\begin{TraceCode}
from PIL import Image, ImageDraw
import numpy as np, json
source=Image.fromarray(image.astype(np.uint8)).convert("RGB")
canvas=Image.new("RGB",(1000,430),"white"); draw=ImageDraw.Draw(canvas)
main=source.crop(tuple(main_bbox)).resize((1000,300)); bias=source.crop(tuple(bias_bbox)).resize((1000,110))
canvas.paste(main,(0,20)); canvas.paste(bias,(0,320))
for day,x in enumerate(centers,1):
  xx=int(round((x-main_bbox[0])/max(main_bbox[2]-main_bbox[0],1)*1000))
  in_a=10<=day<=25; in_b=40<=day<=55
  color=(255,145,0) if in_a else (91,66,243) if in_b else (210,213,220)
  draw.line((xx,20,xx,429),fill=color,width=2 if (in_a or in_b) else 1)
  if day in (10,25,40,55): draw.text((max(0,xx-12),3),"D%d"%day,fill=(30,35,50))
result_image=canvas
print("RESULT_JSON_START")
print(json.dumps({"stage":"render_aligned_focus","task":"compare_ma_gap_and_bias","interval_a":[10,25],"interval_b":[40,55],"panels":["price_and_MA","BIAS"],"result_image_size":list(result_image.size)}, separators=(",", ":")))
print("RESULT_JSON_END")
\end{TraceCode}

\traceoutputheading

\Needspace{4\baselineskip}
\begin{TraceOutput}
{"stage": "render_aligned_focus", "task": "compare_ma_gap_and_bias", "interval_a": [10, 25], "interval_b":
[40, 55], "panels": ["price_and_MA", "BIAS"], "result_image_size": [1000, 430]}
\end{TraceOutput}

\traceimage[0.94]{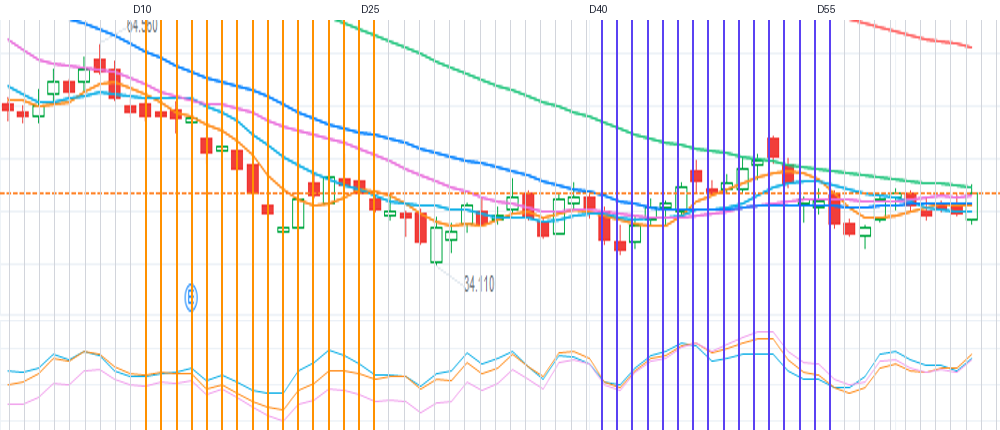}{Aligned-panel working image returned after grid recovery.}

\subsubsection{Turn 4 of 7}

\traceplan{Extract the MA5 and MA20 vertical positions at every recovered day and disclose their absolute separation series.}

\Needspace{4\baselineskip}
\begin{TraceCode}
import numpy as np, json
orange = (r > 190) & (g > 80) & (g < 210) & (b < 120) & (r-g > 35)
magenta = (r > 160) & (b > 120) & (g < 170) & (r-g > 25) & (b-g > 20)
def extract_curve(mask,bbox,radius=3):
  bx1,by1,bx2,by2=bbox; values=[]
  for x in centers:
    xx=int(round(x)); ys,_=np.where(mask[by1:by2,max(bx1,xx-radius):min(bx2,xx+radius+1)])
    values.append(float(np.median(ys+by1)) if ys.size else float("nan"))
  arr=np.asarray(values,float); valid=np.isfinite(arr)
  if valid.sum()<day_count*0.55: raise ValueError("curve coverage too low")
  return np.interp(np.arange(day_count),np.where(valid)[0],arr[valid])
ma5_y=extract_curve(orange,main_bbox)
ma20_y=extract_curve(magenta,main_bbox)
ma_gap_px=np.abs(ma5_y-ma20_y)
print("RESULT_JSON_START")
print(json.dumps({"stage":"extract_moving_averages","task":"compare_ma_gap_and_bias","ma5_y":[round(float(v),2) for v in ma5_y],"ma20_y":[round(float(v),2) for v in
   ma20_y],"absolute_gap_px":[round(float(v),2) for v in ma_gap_px]}, separators=(",", ":")))
print("RESULT_JSON_END")
\end{TraceCode}

\traceoutputheading

\Needspace{4\baselineskip}
\begin{TraceOutput}
{"stage": "extract_moving_averages", "task": "compare_ma_gap_and_bias", "ma5_y": [322.0, 300.0, 302.0,
301.0, 300.0, 295.5, 291.5, 291.0, 293.5, 297.0, 301.0, 305.0, 308.0, 311.0, 315.0, 320.0, 327.0, 337.0,
343.5, 347.0, 350.5, 350.0, 347.0, 345.0, 344.0, 346.0, 345.0, 354.0, 358.0, 360.5, 360.0, 360.5, 358.0,
355.0, 354.0, 356.0, 355.0, 345.0, 354.0, 356.0, 358.0, 359.0, 361.0, 361.0, 345.0, 350.0, 346.5, 344.0,
341.0, 338.0, 336.0, 335.0, 335.0, 338.0, 344.0, 349.0, 345.0, 345.0, 356.0, 345.0, 345.0, 345.0, 345.0,
345.0, 345.0, 345.0], "ma20_y": [270.0, 275.0, 279.0, 281.0, 282.0, 283.0, 284.0, 285.0, 288.0, 291.0,
295.0, 297.0, 299.0, 301.0, 301.0, 303.0, 305.0, 309.0, 312.0, 314.0, 316.0, 318.0, 320.0, 323.0, 326.0,
329.0, 333.0, 336.0, 339.0, 342.0, 344.0, 347.0, 349.0, 350.0, 352.0, 354.0, 354.0, 353.0, 353.0, 354.0,
355.0, 356.0, 357.0, 357.0, 357.0, 356.0, 355.0, 354.0, 352.67, 351.33, 350.0, 349.0, 348.0, 348.0, 348.0,
348.0, 349.0, 349.0, 349.0, 349.0, 347.0, 346.5, 347.0, 346.5, 346.5, 346.5], "absolute_gap_px": [52.0,
25.0, 23.0, 20.0, 18.0, 12.5, 7.5, 6.0, 5.5, 6.0, 6.0, 8.0, 9.0, 10.0, 14.0, 17.0, 22.0, 28.0, 31.5,
33.0, 34.5, 32.0, 27.0, 22.0, 18.0, 17.0, 12.0, 18.0, 19.0, 18.5, 16.0, 13.5, 9.0, 5.0, 2.0, 2.0, 1.0,
8.0, 1.0, 2.0, 3.0, 3.0, 4.0, 4.0, 12.0, 6.0, 8.5, 10.0, 11.67, 13.33, 14.0, 14.0, 13.0, 10.0, 4.0, 1.0,
4.0, 4.0, 7.0, 4.0, 2.0, 1.5, 2.0, 1.5, 1.5, 1.5]}
\end{TraceOutput}

\subsubsection{Turn 5 of 7}

\traceplan{Extract BIAS1 at the same day centers, using smaller screen-y as the higher visible BIAS1 position.}

\Needspace{4\baselineskip}
\begin{TraceCode}
import numpy as np, json
cyan = (b > 135) & (g > 95) & (r < 130) & (b-r > 40)
bias1_y=[]
bx1,by1,bx2,by2=bias_bbox
for x in centers:
  xx=int(round(x)); ys,_=np.where(cyan[by1:by2,max(bx1,xx-3):min(bx2,xx+4)])
  bias1_y.append(float(np.median(ys+by1)) if ys.size else float("nan"))
bias1_y=np.asarray(bias1_y,float); valid=np.isfinite(bias1_y)
if valid.sum()<day_count*0.55: raise ValueError("BIAS1 coverage too low")
bias1_y=np.interp(np.arange(day_count),np.where(valid)[0],bias1_y[valid])
print("RESULT_JSON_START")
print(json.dumps({"stage":"extract_bias1","task":"compare_ma_gap_and_bias","bias1_screen_y":[round(float(v),2) for v in bias1_y],"comparison_rule":"smaller screen-y means higher BIAS1"}, separators=(",", ":")))
print("RESULT_JSON_END")
\end{TraceCode}

\traceoutputheading

\Needspace{4\baselineskip}
\begin{TraceOutput}
{"stage": "extract_bias1", "task": "compare_ma_gap_and_bias", "bias1_screen_y": [466.0, 467.0, 463.5,
455.0, 458.0, 456.75, 455.5, 466.0, 470.0, 470.0, 468.0, 467.0, 465.0, 473.0, 470.0, 474.0, 480.0, 482.0,
482.0, 467.0, 460.5, 452.0, 455.0, 461.0, 468.0, 469.0, 472.5, 476.0, 469.0, 466.0, 455.0, 460.0, 458.0,
453.0, 459.0, 470.0, 458.0, 456.0, 459.0, 470.5, 476.0, 466.0, 457.0, 453.0, 448.0, 448.0, 456.5, 458.0,
455.0, 454.0, 454.0, 462.0, 470.0, 472.0, 476.0, 478.0, 473.0, 459.0, 453.0, 456.0, 461.0, 462.0, 465.0,
460.0, 460.0, 460.0], "comparison_rule": "smaller screen-y means higher BIAS1"}
\end{TraceOutput}

\subsubsection{Turn 6 of 7}

\traceplan{Compute both interval averages, identify the MA-gap winner and the higher-BIAS1 interval, and test agreement.}

\Needspace{4\baselineskip}
\begin{TraceCode}
import numpy as np, json
intervals={"first interval":[10,25],"second interval":[40,55]}
stats={}
for name,(lo,hi) in intervals.items():
  sl=slice(lo-1,hi)
  stats[name]={"mean_abs_ma5_ma20_gap_px":round(float(np.mean(ma_gap_px[sl])),3),"mean_bias1_screen_y":round(float(np.mean(bias1_y[sl])),3),"periods":hi-lo+1}
gap_winner=max(stats,key=lambda name:stats[name]["mean_abs_ma5_ma20_gap_px"])
bias_winner=min(stats,key=lambda name:stats[name]["mean_bias1_screen_y"])
agreement=(gap_winner==bias_winner)
print("RESULT_JSON_START")
print(json.dumps({"stage":"compute_finance","task":"compare_ma_gap_and_bias","intervals":intervals,"statistics":stats,"gap_winner":gap_winner,"higher_bias1_interval":bias_winner,"same_interval":agreement}, separators=(",", ":")))
print("RESULT_JSON_END")
\end{TraceCode}

\traceoutputheading

\Needspace{4\baselineskip}
\begin{TraceOutput}
{"stage": "compute_finance", "task": "compare_ma_gap_and_bias", "intervals": {"first interval": [10,
25], "second interval": [40, 55]}, "statistics": {"first interval": {"mean_abs_ma5_ma20_gap_px": 19.875,
"mean_bias1_screen_y": 468.406, "periods": 16}, "second interval": {"mean_abs_ma5_ma20_gap_px": 8.281,
"mean_bias1_screen_y": 461.0, "periods": 16}}, "gap_winner": "first interval", "higher_bias1_interval":
"second interval", "same_interval": false}
\end{TraceOutput}

\subsubsection{Turn 7 of 7}

\traceplan{Render answer-level evidence with both intervals boxed across the main and BIAS panels.}

\Needspace{4\baselineskip}
\begin{TraceCode}
from PIL import Image, ImageDraw
import numpy as np, json
source=Image.fromarray(image.astype(np.uint8)).convert("RGB")
canvas=Image.new("RGB",(1040,500),"white"); draw=ImageDraw.Draw(canvas)
main=source.crop(tuple(main_bbox)).resize((1040,330)); bias=source.crop(tuple(bias_bbox)).resize((1040,130))
canvas.paste(main,(0,40)); canvas.paste(bias,(0,370))
scale=1040/max(main_bbox[2]-main_bbox[0],1)
for name,(lo,hi) in intervals.items():
  left=int(round((centers[lo-1]-0.5*pitch-main_bbox[0])*scale)); right=int(round((centers[hi-1]+0.5*pitch-main_bbox[0])*scale))
  color=(255,145,0) if name=="first interval" else (91,66,243)
  draw.rectangle((left,40,right,499),outline=color,width=4)
draw.text((12,10),"MA gap winner: %s | higher BIAS1: %s | same=%s"%(gap_winner,bias_winner,str(agreement).upper()),fill=(20,30,50))
result_image=canvas
answer="%s; BIAS1 agreement: %s"%(gap_winner,"yes" if agreement else "no")
print("RESULT_JSON_START")
print(json.dumps({"stage":"render_answer_evidence","task":"compare_ma_gap_and_bias","answer":answer,"statistics":stats,"gap_winner":gap_winner,"higher_bias1_interval":bias_winner,"same_interval":agreement,"result_image_size":list(result_image.size)}, separators=(",", ":")))
print("RESULT_JSON_END")
\end{TraceCode}

\traceoutputheading

\Needspace{4\baselineskip}
\begin{TraceOutput}
{"stage": "render_answer_evidence", "task": "compare_ma_gap_and_bias", "answer": "first interval; BIAS1 agreement: no",
"statistics": {"first interval": {"mean_abs_ma5_ma20_gap_px": 19.875, "mean_bias1_screen_y": 468.406,
"periods": 16}, "second interval": {"mean_abs_ma5_ma20_gap_px": 8.281, "mean_bias1_screen_y": 461.0,
"periods": 16}}, "gap_winner": "first interval", "higher_bias1_interval": "second interval", "same_interval":
false, "result_image_size": [1040, 500]}
\end{TraceOutput}

\traceimage[0.94]{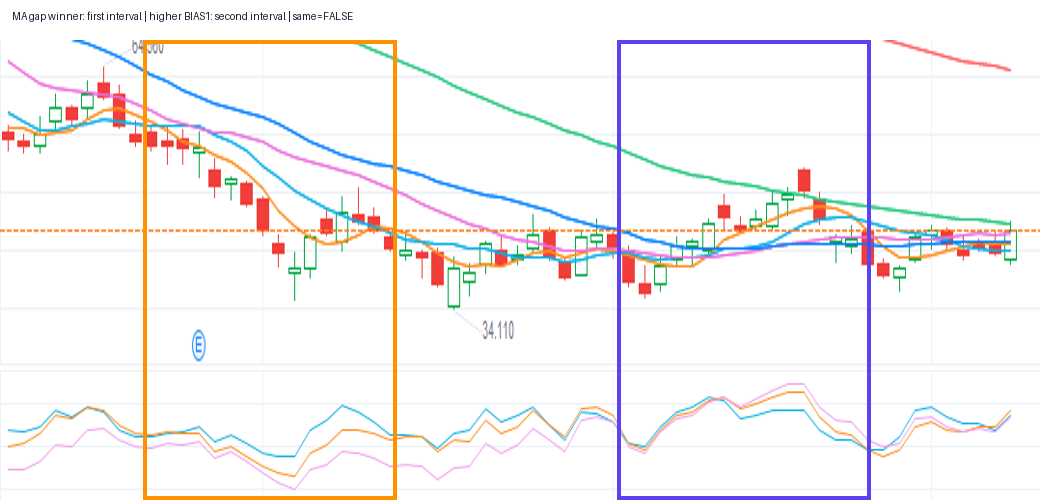}{Answer-level evidence returned after the complete financial computation.}

\subsubsection{Final Evidence Review and Answer}

The final turn reviews the returned evidence against the disclosed per-period values.

\begin{center}\Large\bfseries first interval; BIAS1 agreement: no\end{center}

\subsection{Case 3: Tongdaxin}

\traceimage[0.48]{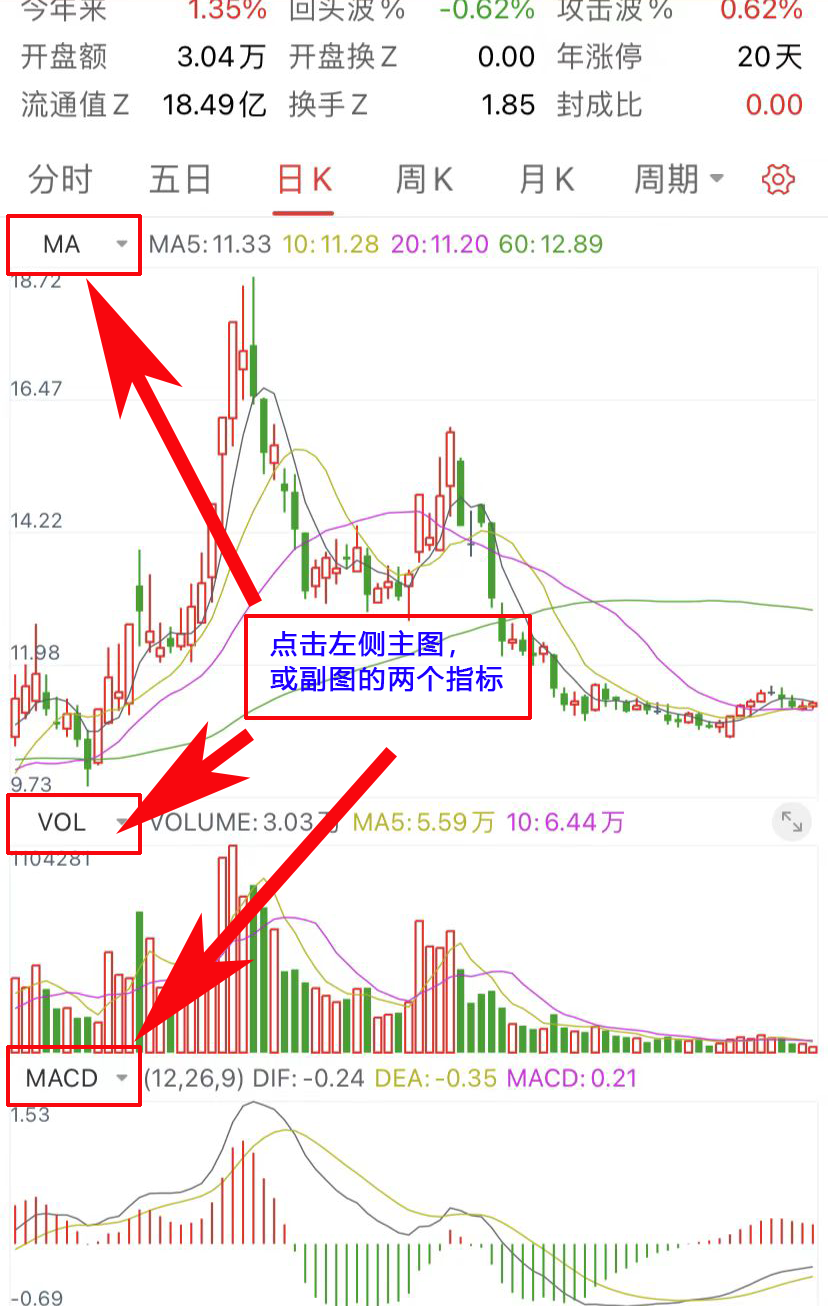}{Input chart for Case 3.}

\textbf{Public Screenshot Source:} \url{https://www.gushichanghong.com/page/140.html}

\paragraph{Question.}

Number the visible daily K-line periods from left to right. Compare Days 60-68 (first interval) with Days 69-76 (second interval). Which interval has the higher maximum volume bar? For that winning interval, report the day of its maximum-volume bar, whether that day's K-line is bullish or bearish, and how many days in the interval have a negative MACD histogram. If a maximum is tied, choose the earliest day.

\paragraph{Replay Result.} \fbox{\textbf{second interval; Day 72; bullish; 0 negative-MACD days}}

\subsubsection{Turn 1 of 7}

\traceplan{Locate the K-line, volume, and MACD panels and explicitly account for the large tutorial-arrow overlay.}

\Needspace{4\baselineskip}
\begin{TraceCode}
from PIL import Image
import numpy as np, json, re
image=np.asarray(Image.open(input_path).convert("RGB")); h,w=image.shape[:2]
rgb=image.astype(np.int16); r,g,b=rgb[:,:,0],rgb[:,:,1],rgb[:,:,2]
red=(r>150)&(r-g>45)&(r-b>35); green=(g>75)&(g-r>28)&(g-b>20)
price_bbox=[0,int(round(0.199*h)),w,int(round(0.651*h))]
volume_bbox=[0,int(round(0.658*h)),w,int(round(0.835*h))]
macd_bbox=[0,int(round(0.858*h)),w,h]
print("RESULT_JSON_START")
print(json.dumps({"stage":"inspect_layout","task":"compare_interval_volume_with_macd","image_size":[w,h],"price_bbox":price_bbox,"volume_bbox":volume_bbox,"macd_bbox":macd_bbox,"visible_annotation_noise":"large red tutorial arrows; excluded by using MACD grid and late clean intervals"}, separators=(",", ":")))
print("RESULT_JSON_END")
\end{TraceCode}

\traceoutputheading

\Needspace{4\baselineskip}
\begin{TraceOutput}
{"stage": "inspect_layout", "task": "compare_interval_volume_with_macd", "image_size": [828, 1306], "price_bbox":
[0, 260, 828, 850], "volume_bbox": [0, 859, 828, 1091], "macd_bbox": [0, 1121, 828, 1306], "visible_annotation_noise":
"large red tutorial arrows; excluded by using MACD grid and late clean intervals"}
\end{TraceOutput}

\subsubsection{Turn 2 of 7}

\traceplan{Recover the complete daily grid from the clean MACD histogram rather than from the contaminated price panel.}

\Needspace{4\baselineskip}
\begin{TraceCode}
import numpy as np, json
mx1,my1,mx2,my2=macd_bbox
profile=(red[my1:my2]|green[my1:my2]).sum(axis=0)
active=np.where(profile>=4)[0]; runs=[]
if active.size:
  start=previous=int(active[0])
  for value in active[1:]:
    value=int(value)
    if value<=previous+3: previous=value
    else: runs.append((start,previous)); start=previous=value
  runs.append((start,previous))
candidate_centers=[0.5*(a+b) for a,b in runs if 1<=b-a+1<=5]
diffs=[b-a for a,b in zip(candidate_centers,candidate_centers[1:]) if 7<=b-a<=14]
if len(diffs)<30: raise ValueError("insufficient MACD bars for grid recovery")
pitch=float(np.median(diffs)); first=float(min(candidate_centers)); last=float(max(candidate_centers))
day_count=int(round((last-first)/pitch))+1; centers=np.linspace(first,last,day_count)
if not 60<=day_count<=90: raise ValueError("recovered day count outside visible chart bounds")
print("RESULT_JSON_START")
print(json.dumps({"stage":"recover_grid","task":"compare_interval_volume_with_macd","grid_source":"MACD histogram centers","day_count":day_count,"pitch_px":round(
   pitch,3),"first_center":round(first,2),"last_center":round(last,2),"centers":[round(float(x),2) for x in centers]}, separators=(",", ":")))
print("RESULT_JSON_END")
\end{TraceCode}

\traceoutputheading

\Needspace{4\baselineskip}
\begin{TraceOutput}
{"stage": "recover_grid", "task": "compare_interval_volume_with_macd", "grid_source": "MACD histogram centers",
"day_count": 77, "pitch_px": 10.5, "first_center": 14.5, "last_center": 812.5, "centers": [14.5, 25.0,
35.5, 46.0, 56.5, 67.0, 77.5, 88.0, 98.5, 109.0, 119.5, 130.0, 140.5, 151.0, 161.5, 172.0, 182.5, 193.0,
203.5, 214.0, 224.5, 235.0, 245.5, 256.0, 266.5, 277.0, 287.5, 298.0, 308.5, 319.0, 329.5, 340.0, 350.5,
361.0, 371.5, 382.0, 392.5, 403.0, 413.5, 424.0, 434.5, 445.0, 455.5, 466.0, 476.5, 487.0, 497.5, 508.0,
518.5, 529.0, 539.5, 550.0, 560.5, 571.0, 581.5, 592.0, 602.5, 613.0, 623.5, 634.0, 644.5, 655.0, 665.5,
676.0, 686.5, 697.0, 707.5, 718.0, 728.5, 739.0, 749.5, 760.0, 770.5, 781.0, 791.5, 802.0, 812.5]}
\end{TraceOutput}

\subsubsection{Turn 3 of 7}

\traceplan{Render the late clean portion of all three aligned panels with both target intervals marked.}

\Needspace{4\baselineskip}
\begin{TraceCode}
from PIL import Image, ImageDraw
import numpy as np, json
source=Image.fromarray(image.astype(np.uint8)).convert("RGB")
x_left=max(0,int(round(centers[58]-0.6*pitch))); x_right=min(w,int(round(centers[75]+0.6*pitch)))
canvas=Image.new("RGB",(920,700),"white"); draw=ImageDraw.Draw(canvas); yoff=45
price_focus_bbox=[price_bbox[0],int(round(0.38*h)),price_bbox[2],price_bbox[3]]
for bbox,target_h,label in [(price_focus_bbox,330,"K-LINE"),(volume_bbox,165,"VOLUME"),(macd_bbox,155,"MACD")]:
  crop=source.crop((x_left,bbox[1],x_right,bbox[3])).resize((920,target_h)); canvas.paste(crop,(0,yoff)); draw.text((8,yoff+5),label,fill=(20,30,50)); yoff+=target_h
for day in range(60,77):
  xx=int(round((centers[day-1]-x_left)/max(x_right-x_left,1)*920))
  color=(255,145,0) if day<=68 else (91,66,243)
  draw.line((xx,45,xx,694),fill=color,width=2)
  if day in (60,68,69,76): draw.text((max(0,xx-12),20),"D%d"%day,fill=(25,30,45))
result_image=canvas
print("RESULT_JSON_START")
print(json.dumps({"stage":"render_aligned_focus","task":"compare_interval_volume_with_macd","interval_a":[60,68],"interval_b":[69,76],"panels":["price","volume","MACD"],"result_image_size":list(result_image.size)}, separators=(",", ":")))
print("RESULT_JSON_END")
\end{TraceCode}

\traceoutputheading

\Needspace{4\baselineskip}
\begin{TraceOutput}
{"stage": "render_aligned_focus", "task": "compare_interval_volume_with_macd", "interval_a": [60, 68],
"interval_b": [69, 76], "panels": ["price", "volume", "MACD"], "result_image_size": [920, 700]}
\end{TraceOutput}

\traceimage[0.7]{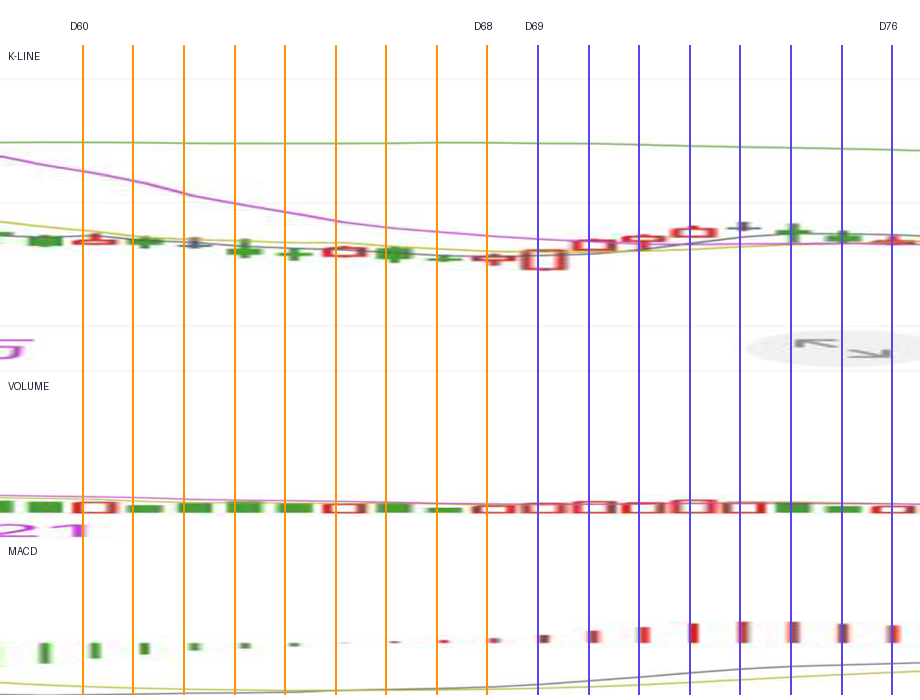}{Aligned-panel working image returned after grid recovery.}

\subsubsection{Turn 4 of 7}

\traceplan{Measure every aligned volume-bar height while selecting the bottom-connected bar component at each day.}

\Needspace{4\baselineskip}
\begin{TraceCode}
import numpy as np, json
vx1,vy1,vx2,vy2=volume_bbox
volume_heights=[]; volume_runs=[]
for x in centers:
  xx=int(round(x)); rows=np.where((red[vy1:vy2,max(0,xx-5):min(w,xx+6)]|green[vy1:vy2,max(0,xx-5):min(w,xx+6)]).sum(axis=1)>=2)[0]
  local=[]
  if rows.size:
    start=previous=int(rows[0])
    for value in rows[1:]:
     value=int(value)
     if value<=previous+3: previous=value
     else: local.append((start,previous)); start=previous=value
    local.append((start,previous))
  candidates=[pair for pair in local if pair[1]-pair[0]+1>=4 and pair[1]>=int(0.75*(vy2-vy1))]
  selected=max(candidates,key=lambda pair:(pair[1],pair[1]-pair[0])) if candidates else (0,0)
  volume_runs.append([int(selected[0]+vy1),int(selected[1]+vy1)])
  volume_heights.append(int(selected[1]-selected[0]+1))
print("RESULT_JSON_START")
print(json.dumps({"stage":"extract_volume","task":"compare_interval_volume_with_macd","day_count":day_count,"volume_heights_px":volume_heights,"volume_runs":
   volume_runs}, separators=(",", ":")))
print("RESULT_JSON_END")
\end{TraceCode}

\traceoutputheading

\Needspace{4\baselineskip}
\begin{TraceOutput}
{"stage": "extract_volume", "task": "compare_interval_volume_with_macd", "day_count": 77, "volume_heights_px":
[76, 68, 91, 66, 46, 48, 37, 38, 33, 104, 80, 78, 180, 117, 75, 94, 113, 134, 140, 195, 195, 195, 170,
170, 177, 126, 84, 86, 72, 70, 60, 54, 67, 69, 66, 40, 42, 54, 134, 134, 108, 124, 124, 84, 60, 64, 64,
46, 32, 28, 26, 40, 40, 28, 28, 28, 28, 18, 18, 18, 16, 16, 18, 16, 14, 16, 8, 14, 16, 18, 16, 20, 17,
16, 16, 16, 9], "volume_runs": [[978, 1053], [986, 1053], [963, 1053], [988, 1053], [1008, 1053], [1006,
1053], [1017, 1053], [1016, 1053], [1021, 1053], [950, 1053], [974, 1053], [976, 1053], [911, 1090],
[937, 1053], [979, 1053], [960, 1053], [941, 1053], [920, 1053], [914, 1053], [859, 1053], [859, 1053],
[859, 1053], [884, 1053], [884, 1053], [877, 1053], [928, 1053], [970, 1053], [968, 1053], [982, 1053],
[986, 1055], [994, 1053], [1000, 1053], [987, 1053], [987, 1055], [988, 1053], [1014, 1053], [1012, 1053],
[1000, 1053], [920, 1053], [920, 1053], [946, 1053], [930, 1053], [930, 1053], [970, 1053], [994, 1053],
[990, 1053], [990, 1053], [1008, 1053], [1022, 1053], [1026, 1053], [1028, 1053], [1014, 1053], [1014,
1053], [1026, 1053], [1026, 1053], [1026, 1053], [1026, 1053], [1036, 1053], [1036, 1053], [1036, 1053],
[1038, 1053], [1038, 1053], [1036, 1053], [1038, 1053], [1040, 1053], [1038, 1053], [1046, 1053], [1040,
1053], [1038, 1053], [1036, 1053], [1038, 1053], [1034, 1053], [1037, 1053], [1038, 1053], [1038, 1053],
[1038, 1053], [1045, 1053]]}
\end{TraceOutput}

\subsubsection{Turn 5 of 7}

\traceplan{Classify every MACD histogram bar as positive or negative on the same recovered grid.}

\Needspace{4\baselineskip}
\begin{TraceCode}
import json
mx1,my1,mx2,my2=macd_bbox
macd_signs=[]; macd_counts=[]
for x in centers:
  xx=int(round(x)); rc=int(red[my1:my2,max(0,xx-3):min(w,xx+4)].sum()); gc=int(green[my1:my2,max(0,xx-3):min(w,xx+4)].sum())
  macd_signs.append("negative" if gc>rc else "positive"); macd_counts.append([rc,gc])
print("RESULT_JSON_START")
print(json.dumps({"stage":"extract_macd_sign","task":"compare_interval_volume_with_macd","macd_signs":macd_signs,"red_green_counts":macd_counts}, separators=(",", ":")))
print("RESULT_JSON_END")
\end{TraceCode}

\traceoutputheading

\Needspace{4\baselineskip}
\begin{TraceOutput}
{"stage": "extract_macd_sign", "task": "compare_interval_volume_with_macd", "macd_signs": ["positive",
"positive", "positive", "positive", "positive", "positive", "positive", "positive", "positive", "positive",
"positive", "positive", "positive", "positive", "positive", "positive", "positive", "positive", "positive",
"positive", "positive", "positive", "positive", "positive", "positive", "positive", "positive", "negative",
"negative", "negative", "negative", "negative", "negative", "positive", "positive", "negative", "negative",
"positive", "positive", "positive", "positive", "positive", "positive", "positive", "positive", "negative",
"negative", "positive", "negative", "negative", "negative", "negative", "negative", "negative", "negative",
"negative", "negative", "negative", "negative", "negative", "negative", "negative", "negative", "positive",
"positive", "positive", "positive", "positive", "positive", "positive", "positive", "positive", "positive",
"positive", "positive", "positive", "positive"], "red_green_counts": [[80, 2], [119, 0], [91, 0], [106,
0], [58, 0], [48, 0], [27, 0], [0, 0], [0, 0], [25, 0], [19, 0], [54, 0], [76, 1], [71, 0], [74, 0],
[55, 0], [42, 0], [46, 0], [47, 0], [1, 0], [192, 0], [194, 0], [105, 1], [93, 0], [122, 0], [70, 0],
[20, 0], [0, 4], [0, 45], [0, 36], [0, 32], [0, 60], [0, 62], [0, 0], [0, 0], [0, 6], [0, 40], [0, 0],
[0, 0], [0, 0], [0, 0], [0, 0], [0, 0], [0, 0], [0, 0], [0, 19], [0, 72], [0, 0], [0, 4], [0, 58], [0,
113], [0, 100], [0, 83], [0, 76], [0, 106], [0, 80], [0, 77], [0, 32], [0, 63], [0, 46], [0, 24], [0,
20], [0, 10], [0, 0], [0, 0], [0, 0], [11, 0], [11, 0], [5, 0], [42, 0], [42, 0], [50, 0], [52, 0], [75,
0], [29, 0], [49, 0], [42, 0]]}
\end{TraceOutput}

\subsubsection{Turn 6 of 7}

\traceplan{Compare the interval maxima, identify the winning day, and count negative-MACD days in the winning interval.}

\Needspace{4\baselineskip}
\begin{TraceCode}
import numpy as np, json
intervals={"first interval":[60,68],"second interval":[69,76]}; stats={}
for name,(lo,hi) in intervals.items():
  values=volume_heights[lo-1:hi]; signs=macd_signs[lo-1:hi]
  maximum=max(values); day=lo+int(np.argmax(values))
  stats[name]={"max_volume_height_px":int(maximum),"max_volume_day":int(day),"negative_macd_days":int(sum(value=="negative" for value in signs)),"volume_heights_px":values,"macd_signs":signs}
winner=max(stats,key=lambda name:stats[name]["max_volume_height_px"]); selected_day=int(stats[winner]["max_volume_day"])
print("RESULT_JSON_START")
print(json.dumps({"stage":"compute_finance","task":"compare_interval_volume_with_macd","intervals":intervals,"statistics":stats,"winning_interval":winner,"selected_day":selected_day,"tie_break":"earliest day"}, separators=(",", ":")))
print("RESULT_JSON_END")
\end{TraceCode}

\traceoutputheading

\Needspace{4\baselineskip}
\begin{TraceOutput}
{"stage": "compute_finance", "task": "compare_interval_volume_with_macd", "intervals": {"first interval":
[60, 68], "second interval": [69, 76]}, "statistics": {"first interval": {"max_volume_height_px": 18,
"max_volume_day": 60, "negative_macd_days": 4, "volume_heights_px": [18, 16, 16, 18, 16, 14, 16, 8, 14],
"macd_signs": ["negative", "negative", "negative", "negative", "positive", "positive", "positive", "positive",
"positive"]}, "second interval": {"max_volume_height_px": 20, "max_volume_day": 72, "negative_macd_days":
0, "volume_heights_px": [16, 18, 16, 20, 17, 16, 16, 16], "macd_signs": ["positive", "positive", "positive",
"positive", "positive", "positive", "positive", "positive"]}}, "winning_interval": "second interval",
"selected_day": 72, "tie_break": "earliest day"}
\end{TraceOutput}

\subsubsection{Turn 7 of 7}

\traceplan{Classify the selected day's K-line direction and render joint price-volume-MACD evidence.}

\Needspace{4\baselineskip}
\begin{TraceCode}
from PIL import Image, ImageDraw
import numpy as np, json
sx=int(round(centers[selected_day-1])); px1,py1,px2,py2=price_bbox
red_pixels=int(red[py1:py2,max(0,sx-4):min(w,sx+5)].sum()); green_pixels=int(green[py1:py2,max(0,sx-4):min(w,sx+5)].sum())
candle_direction="bullish" if red_pixels>green_pixels else "bearish"
x_left=max(0,int(round(centers[58]-0.6*pitch))); x_right=min(w,int(round(centers[75]+0.6*pitch)))
source=Image.fromarray(image.astype(np.uint8)).convert("RGB"); canvas=Image.new("RGB",(960,745),"white"); draw=ImageDraw.Draw(canvas)
draw.text((12,10),"%s wins | max D%d=%d px | %s | negative MACD days=%d"%(winner,selected_day,stats[winner]["max_volume_height_px"],candle_direction.upper(),stats[
   winner]["negative_macd_days"]),fill=(20,30,50))
yoff=55
price_focus_bbox=[price_bbox[0],int(round(0.38*h)),price_bbox[2],price_bbox[3]]
for bbox,target_h,label in [(price_focus_bbox,330,"K-LINE"),(volume_bbox,180,"VOLUME"),(macd_bbox,180,"MACD")]:
  crop=source.crop((x_left,bbox[1],x_right,bbox[3])).resize((960,target_h)); canvas.paste(crop,(0,yoff)); draw.text((8,yoff+5),label,fill=(20,30,50))
  xx=int(round((sx-x_left)/max(x_right-x_left,1)*960)); draw.rectangle((xx-12,yoff,xx+12,yoff+target_h-1),outline=(255,0,180),width=4); yoff+=target_h
result_image=canvas
answer="%s; Day %d; %s; %d negative-MACD days"%(winner,selected_day,candle_direction,stats[winner]["negative_macd_days"])
print("RESULT_JSON_START")
print(json.dumps({"stage":"render_answer_evidence","task":"compare_interval_volume_with_macd","answer":answer,"winning_interval":winner,"selected_day":selected_day,
   "selected_candle_direction":candle_direction,"candle_red_green_pixels":[red_pixels,green_pixels],"negative_macd_days_in_winner":stats[winner]["negative_macd_days"],"statistics":stats,"result_image_size":list(result_image.size)}, separators=(",", ":")))
print("RESULT_JSON_END")
\end{TraceCode}

\traceoutputheading

\Needspace{4\baselineskip}
\begin{TraceOutput}
{"stage": "render_answer_evidence", "task": "compare_interval_volume_with_macd", "answer": "second interval; Day 72; bullish; 0 negative-MACD days",
"winning_interval": "second interval", "selected_day": 72, "selected_candle_direction": "bullish", "candle_red_green_pixels":
[68, 1], "negative_macd_days_in_winner": 0, "statistics": {"first interval": {"max_volume_height_px":
18, "max_volume_day": 60, "negative_macd_days": 4, "volume_heights_px": [18, 16, 16, 18, 16, 14, 16,
8, 14], "macd_signs": ["negative", "negative", "negative", "negative", "positive", "positive", "positive",
"positive", "positive"]}, "second interval": {"max_volume_height_px": 20, "max_volume_day": 72, "negative_macd_days":
0, "volume_heights_px": [16, 18, 16, 20, 17, 16, 16, 16], "macd_signs": ["positive", "positive", "positive",
"positive", "positive", "positive", "positive", "positive"]}}, "result_image_size": [960, 745]}
\end{TraceOutput}

\traceimage[0.7]{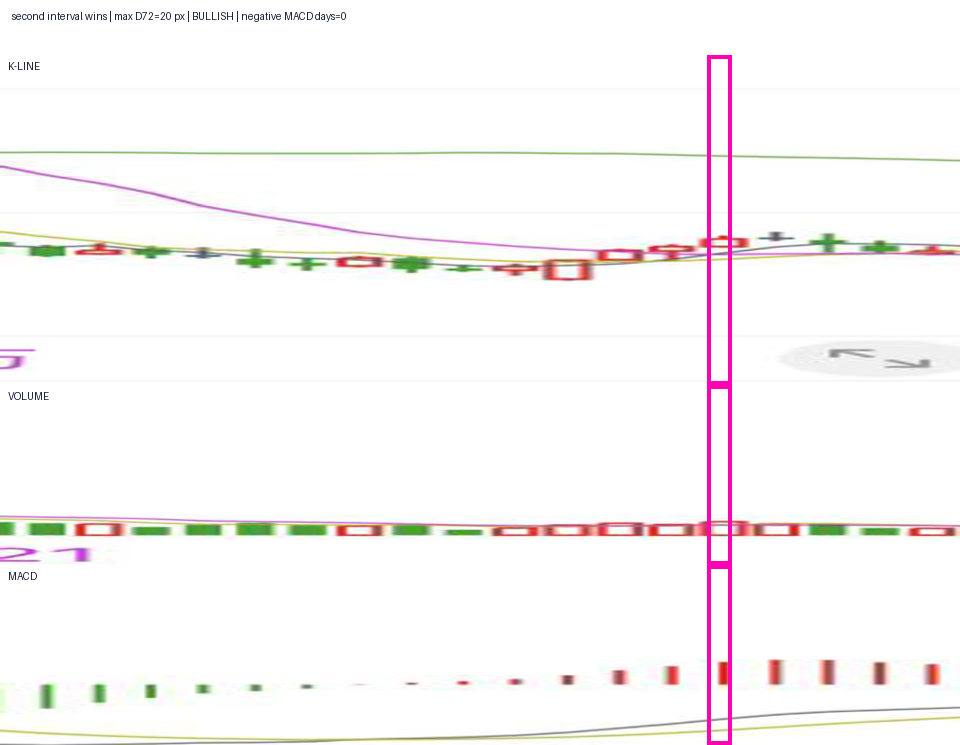}{Answer-level evidence returned after the complete financial computation.}

\subsubsection{Final Evidence Review and Answer}

The final turn reviews the returned evidence against the disclosed per-period values.

\begin{center}\Large\bfseries second interval; Day 72; bullish; 0 negative-MACD days\end{center}

The three reference trajectories span an interval-extrema task, a visual-transcription task, and a volume-comparison task. They use the same presentation as the cross-application cases, with executable turns, compact sandbox records, returned evidence, and final review.

\clearpage
\subsection{Example 1: Interval Lowest Day}

\traceimage[0.45]{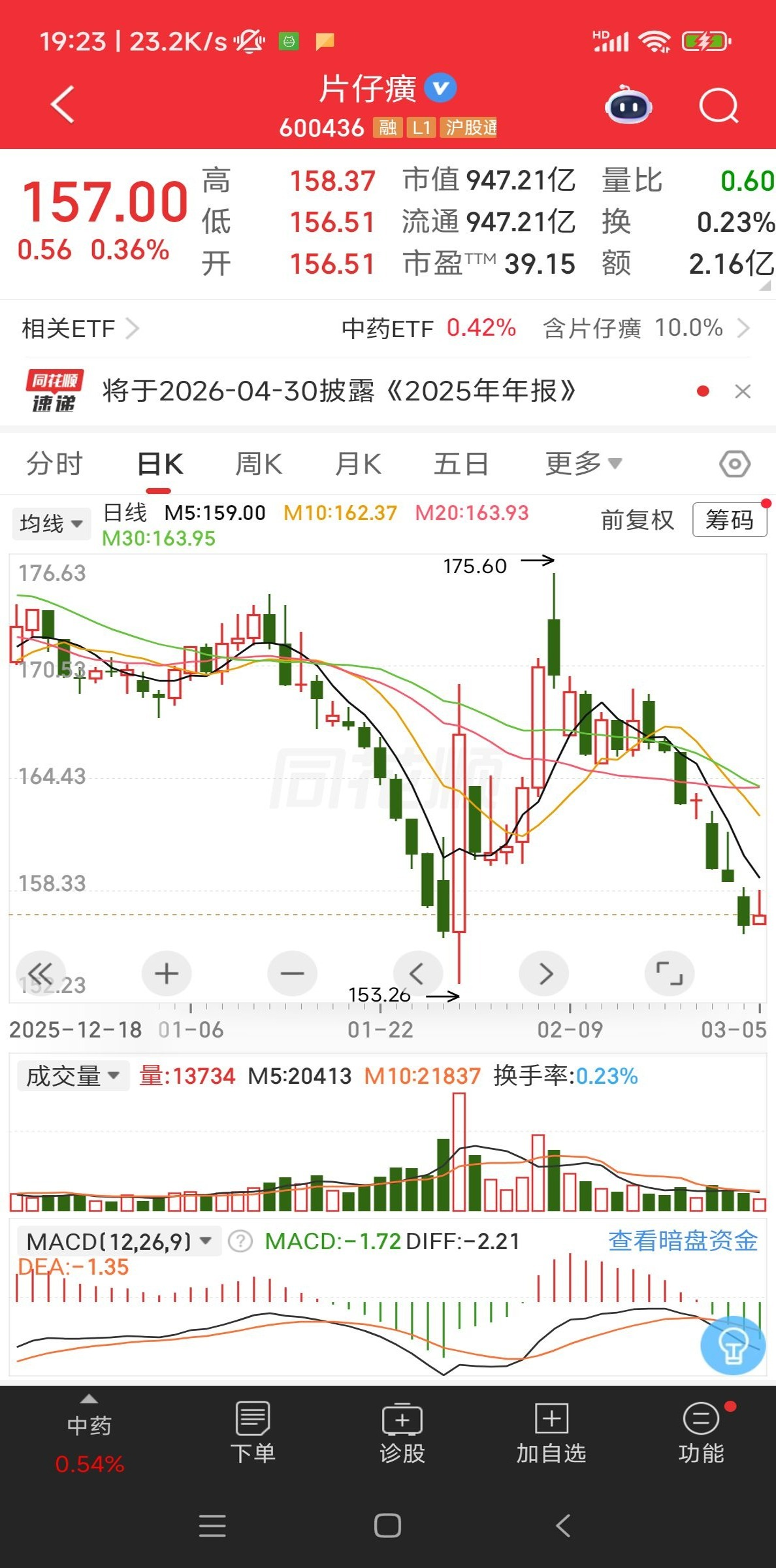}{Input chart for Example 1.}

\paragraph{Question.}

Number the trading days from left to right in the main candlestick chart, with Day 1 at the far left. Between Day 9 and Day 21, on which trading day does the lowest price occur? If there is a tie, return the earliest day.

\paragraph{Replay Result.} \fbox{\textbf{20}}

\subsubsection{Turn 1 of 7}

\traceplan{The image contains three vertically stacked panels: the main daily candlestick chart, a volume panel, and a MACD panel. First locate the main price panel and its trading-day columns so Days 9-21 can later be compared one by one.}

\Needspace{4\baselineskip}
\begin{TraceCode}
from PIL import Image
import numpy as np, json, re
image = np.asarray(Image.open(input_path).convert('RGB'))
h, w = image.shape[:2]
def color_masks(arr):
    rgb = arr.astype(np.int16)
    r, g, b = (rgb[:, :, 0], rgb[:, :, 1], rgb[:, :, 2])
    red = (r > 185) & (g < 115) & (b < 130) & (r - g > 60) & (r - b > 55)
    green = (g > 45) & (g < 175) & (r < 125) & (b < 130) & (g - r > 12) & (g - b > 10)
    return (red, green)
def runs(values, max_gap=1):
    values = [int(v) for v in values]
    if not values:
        return []
    out, start, previous = ([], values[0], values[0])
    for value in values[1:]:
        if value - previous <= max_gap:
            previous = value
        else:
            out.append((start, previous))
            start = previous = value
    out.append((start, previous))
    return out
red_all, green_all = color_masks(image)
mask_all = red_all | green_all
row_counts = mask_all.sum(1)
row_ids = np.arange(h)
min_y, max_y = (int(h * 0.28), int(h * 0.8))
active_rows = np.where((row_counts >= max(6, int(w * 0.005))) & (row_ids >= min_y) & (row_ids <= max_y))[0]
segments = [(a, b) for a, b in runs(active_rows, 5) if b - a + 1 >= max(70, int(h * 0.04)) and a < int(h * 0.72)]
if not segments:
    chart_bbox = [0, int(h * 0.38), w, int(h * 0.7)]
else:
    def panel_score(seg):
        a, b = seg
        active_cols = int(np.count_nonzero(mask_all[a:b + 1].sum(0) >= 3))
        return (b - a + 1) * (1 + active_cols / max(w, 1)) * (1 + max(0, h * 0.72 - a) / max(h, 1))
    core_top, core_bottom = max(segments, key=panel_score)
    top = max(min_y, int(core_top - max(90, h * 0.06)))
    bottom = min(max_y, int(core_bottom + max(75, h * 0.045)))
    gray = image.astype(np.int16)
    grayish = (np.abs(gray[:, :, 0] - gray[:, :, 1]) < 8) & (np.abs(gray[:, :, 1] - gray[:, :, 2]) < 8) & (gray[:, :,0] > 205) & (gray[:, :, 0] < 248)
    gray_rows = np.where(grayish.sum(1) > w * 0.35)[0]
    gray_rows = gray_rows[(gray_rows > core_bottom + 35) & (gray_rows < h * 0.78)]
    gray_segments = [(a, b) for a, b in runs(gray_rows, 3) if b - a + 1 >= 8]
    if gray_segments:
        bottom = max(bottom, min(max_y, min((a for a, _ in gray_segments)) - 5))
    chart_bbox = [0, int(top), w, int(bottom)]
volume_top = min(h - 1, int(chart_bbox[3] + max(50, h * 0.023)))
panel_h = int(max(190, min(260, h * 0.105)))
volume_bottom = min(h, volume_top + panel_h)
gray = image.astype(np.int16)
grayish = (np.abs(gray[:, :, 0] - gray[:, :, 1]) < 10) & (np.abs(gray[:, :, 1] - gray[:, :, 2]) < 10) & (gray[:, :, 0] > 205) & (gray[:, :, 0] < 248)
search_start, search_end = (min(h - 1, volume_top + int(panel_h * 0.45)), min(h, volume_top + int(panel_h * 1.08)))
gray_rows = np.where(grayish[search_start:search_end].sum(1) > w * 0.28)[0]
if gray_rows.size:
    candidate = int(search_start + gray_rows[0] - 3)
    if candidate > volume_top + 120:
        volume_bottom = min(volume_bottom, candidate)
volume_bbox = [0, volume_top, w, volume_bottom]
layout_state = {'layout_kind': 'long_multi_panel', 'image_size': [w, h], 'chart_bbox': chart_bbox, 'volume_bbox':volume_bbox}
print('RESULT_JSON_START')
print(json.dumps({'stage': 'inspect_layout', **layout_state}, ensure_ascii=False))
print('RESULT_JSON_END')
\end{TraceCode}

\traceoutputheading

\Needspace{4\baselineskip}
\begin{TraceOutput}
{"stage": "inspect_layout", "layout_kind": "long_multi_panel", "image_size": [1188, 2400], "chart_bbox":
[0, 768, 1188, 1537], "volume_bbox": [0, 1592, 1188, 1729]}
\end{TraceOutput}

\subsubsection{Turn 2 of 7}

\traceplan{The disclosed layout places the main chart at y=768-1537 across the full 1,188-pixel width. Recover the regularly spaced trading-day centers within this region so Days 9-21 can be mapped to exact horizontal positions.}

\Needspace{4\baselineskip}
\begin{TraceCode}
from PIL import Image, ImageDraw
import re
import json
import numpy as np
REQUIRED_PUBLIC_STATE = ('image', 'chart_bbox', 'color_masks')
missing_public_state = [name for name in REQUIRED_PUBLIC_STATE if name not in globals()]
if missing_public_state:
    raise RuntimeError('missing prior public state: ' + ','.join(missing_public_state))
GRID_CONTRACT = {'source': 'task_contract', 'expected_visible_days': [40, 55]}
x1, y1, x2, y2 = chart_bbox
red_panel, green_panel = color_masks(image[y1:y2, x1:x2])
profile = (red_panel | green_panel).sum(0).astype(np.float64)
profile = np.maximum(profile - float(np.median(profile)), 0.0)
def autocorrelation(values, lag):
    left, right = (values[:-lag], values[lag:])
    denom = float(np.linalg.norm(left) * np.linalg.norm(right))
    return float(np.dot(left, right) / denom) if denom else 0.0
panel_width = x2 - x1
proposals = [(autocorrelation(profile, p), p) for p in range(max(8, int(np.floor(panel_width / 55))), min(64,int(np.ceil(panel_width / 40))) + 1)]
periodicity, pitch = max(proposals, key=lambda item: (item[0], -item[1]))
best = None
for half_phase in range(1, 2 * pitch + 1):
    phase = half_phase / 2.0
    centers_try = []
    k = -2
    while phase + k * pitch < panel_width + 2 * pitch:
        local_x = phase + k * pitch
        if 0.65 * pitch <= local_x <= panel_width - 0.65 * pitch:
            centers_try.append(float(x1 + local_x))
        k += 1
    if not 40 <= len(centers_try) <= 55:
        continue
    support = []
    for center in centers_try:
        local = int(round(center - x1))
        support.append(float(profile[max(0, local - 1):min(panel_width, local + 2)].sum()))
    candidate = (float(np.mean(support)) if support else 0.0, -phase, centers_try)
    if best is None or candidate[:2] > best[:2]:
        best = candidate
if best is None:
    raise ValueError('cannot recover a full trading-day grid')
centers = best[2]
day_count = len(centers)
if not all((b > a for a, b in zip(centers, centers[1:]))):
    raise ValueError('non-monotonic grid')
grid_state = {'day_count': day_count, 'pitch': float(pitch), 'periodicity': round(float(periodicity), 4), 'centers':[round(x, 2) for x in centers]}
print('RESULT_JSON_START')
print(json.dumps({'stage': 'locate_grid', 'contract_prior': GRID_CONTRACT, **grid_state}, ensure_ascii=False))
print('RESULT_JSON_END')
\end{TraceCode}

\traceoutputheading

\Needspace{4\baselineskip}
\begin{TraceOutput}
{"stage": "locate_grid", "contract_prior": {"source": "task_contract", "expected_visible_days": [40,
55]}, "day_count": 48, "pitch": 24.0, "periodicity": 0.4303, "centers": [31.0, 55.0, 79.0, 103.0, 127.0,
151.0, 175.0, 199.0, 223.0, 247.0, 271.0, 295.0, 319.0, 343.0, 367.0, 391.0, 415.0, 439.0, 463.0, 487.0,
511.0, 535.0, 559.0, 583.0, 607.0, 631.0, 655.0, 679.0, 703.0, 727.0, 751.0, 775.0, 799.0, 823.0, 847.0,
871.0, 895.0, 919.0, 943.0, 967.0, 991.0, 1015.0, 1039.0, 1063.0, 1087.0, 1111.0, 1135.0, 1159.0]}
\end{TraceOutput}

\subsubsection{Turn 3 of 7}

\traceplan{The main chart contains 48 evenly spaced trading-day centers at roughly 24-pixel intervals, and Days 9-21 form one continuous span. Generate a focused crop of that span so candle bodies and lower wicks can be inspected more clearly.}

\Needspace{4\baselineskip}
\begin{TraceCode}
import numpy as np
import json
import re
from PIL import Image, ImageDraw
FOCUS_MODALITIES=['candle', 'volume']
source_image=Image.fromarray(image.astype("uint8")).convert("RGB")
if FOCUS_MODALITIES==["volume"]:
    raw_box=list(volume_bbox); purpose="volume_panel"
elif "volume" in FOCUS_MODALITIES and ("candle" in FOCUS_MODALITIES or "ma" in FOCUS_MODALITIES):
    legend_margin=max(70,int((chart_bbox[3]-chart_bbox[1])*.13)) if "ma" in FOCUS_MODALITIES else 0
    raw_box=[chart_bbox[0],max(0,chart_bbox[1]-legend_margin),chart_bbox[2],volume_bbox[3]];purpose="price_volume_alignment"
else:
    legend_margin=max(70,int((chart_bbox[3]-chart_bbox[1])*.13)) if "ma" in FOCUS_MODALITIES else 0
    raw_box=[chart_bbox[0],max(0,chart_bbox[1]-legend_margin),chart_bbox[2],chart_bbox[3]];purpose="price_panel_with_ma_legend" if "ma" in FOCUS_MODALITIES else "price_panel"
pad=max(6,int(round(pitch*.5)))
focus_crop_bbox=[max(0,int(raw_box[0])-pad),max(0,int(raw_box[1])-pad),min(source_image.width,int(raw_box[2])+pad),min(source_image.height,int(raw_box[3])+pad)]
result_image=source_image.crop(tuple(focus_crop_bbox))
border=ImageDraw.Draw(result_image);border.rectangle((1,1,result_image.width-2,result_image.height-2),outline=(255,170,0),width=3)
# Promote the reviewed crop to the persistent visual workspace.  All later
# perception cells use these local pixels and local coordinates; the original
# screenshot is not silently re-read.
focus_origin=[focus_crop_bbox[0],focus_crop_bbox[1]]
crop_x0,crop_y0,crop_x1,crop_y1=focus_crop_bbox
image=np.asarray(source_image.crop(tuple(focus_crop_bbox))).copy(); h,w=image.shape[:2]
chart_bbox=[int(chart_bbox[0])-crop_x0,int(chart_bbox[1])-crop_y0,int(chart_bbox[2])-crop_x0,int(chart_bbox[3])-crop_y0]
volume_bbox=[int(volume_bbox[0])-crop_x0,int(volume_bbox[1])-crop_y0,int(volume_bbox[2])-crop_x0,int(volume_bbox[3])-crop_y0]
centers=[float(value)-crop_x0 for value in centers]
if "candle_mask" in globals(): candle_mask=candle_mask[crop_y0:crop_y1,crop_x0:crop_x1].copy()
for name in ("rgb","r","g","b","red","green","red_all","green_all","mask_all"):
    if name in globals(): globals()[name]=globals()[name][crop_y0:crop_y1,crop_x0:crop_x1].copy()
# Layout/grid cells also publish these scalar aliases.  Rebind them to the
# promoted crop coordinate frame; leaving the original-image y1/y2 values in
# the persistent sandbox makes later extractors scan the wrong rows.
x1,y1,x2,y2=[int(value) for value in chart_bbox]
payload={"stage":"render_focus_crop","task":'interval_lowest_day',"purpose":purpose,"modalities":FOCUS_MODALITIES,"source_crop_bbox":focus_crop_bbox,"result_image_size":list(result_image.size),"must_review_before_next_extraction":True,"workspace_promoted":True,"workspace_coordinate_frame":"focus_crop_local"}
print("RESULT_JSON_START"); print(json.dumps(payload,ensure_ascii=False,separators=(",",":")));print("RESULT_JSON_END")
\end{TraceCode}

\traceoutputheading

\Needspace{4\baselineskip}
\begin{TraceOutput}
{"stage": "render_focus_crop", "task": "interval_lowest_day", "purpose": "price_volume_alignment", "modalities":
["candle", "volume"], "source_crop_bbox": [0, 756, 1188, 1741], "result_image_size": [1188, 985], "must_review_before_next_extraction":
true, "workspace_promoted": true, "workspace_coordinate_frame": "focus_crop_local"}
\end{TraceOutput}

\traceimage[0.7]{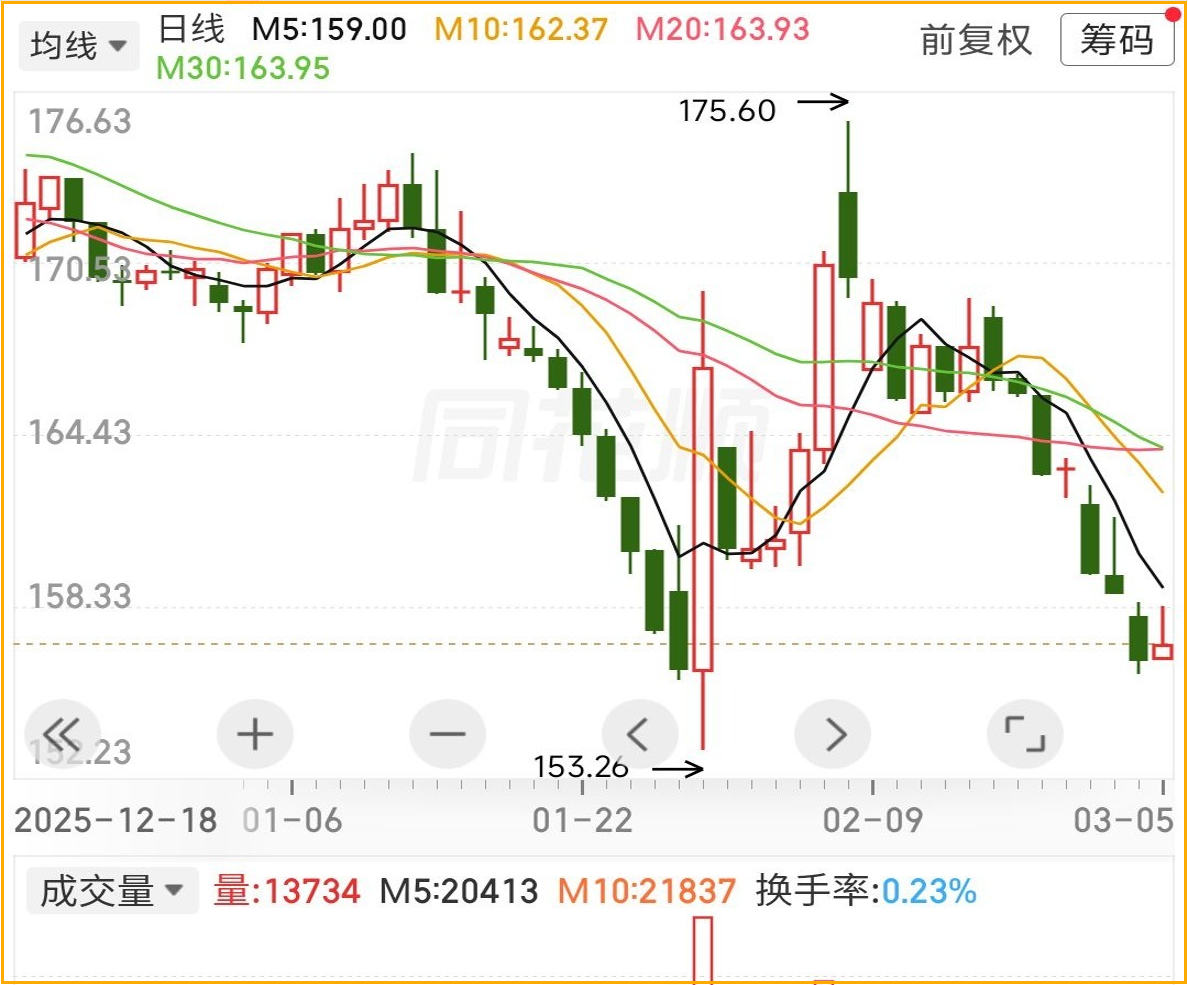}{Focused price-volume evidence after grid recovery.}

\subsubsection{Turn 4 of 7}

\traceplan{The focused image shows the candlestick and volume panels aligned on the same horizontal axis. Extract each volume bar's red/green direction as same-day color evidence before identifying candle bodies and lower wicks.}

\Needspace{4\baselineskip}
\begin{TraceCode}
from PIL import Image, ImageDraw
import re
import json
import numpy as np
REQUIRED_PUBLIC_STATE = ('image', 'volume_bbox', 'centers', 'pitch', 'color_masks', 'runs')
missing_public_state = [name for name in REQUIRED_PUBLIC_STATE if name not in globals()]
if missing_public_state:
    raise RuntimeError('missing prior public state: ' + ','.join(missing_public_state))
v1, vtop, v2, vbottom = volume_bbox
red_volume, green_volume = color_masks(image[vtop:vbottom, v1:v2])
half = max(5, min(11, int(round(pitch * 0.42))))
volume_bars = []
for day, x in enumerate(centers, 1):
    left = max(0, int(round(x)) - half)
    right = min(w, int(round(x)) + half + 1)
    red_sub, green_sub = (red_volume[:, left:right], green_volume[:, left:right])
    mask = red_sub | green_sub
    row_counts = mask.sum(1)
    active = np.where(row_counts >= 2)[0]
    if not active.size:
        active = np.where(mask.any(1))[0]
    if not active.size:
        volume_bars.append({'day_index': day, 'x': round(float(x), 2), 'height_px': None, 'top_y': int(vbottom - 1),'bottom_y': int(vbottom - 1), 'color': 'unknown', 'valid': False})
        continue
    active_segments = runs(active, 2)
    bar_top, bar_bottom = max(active_segments, key=lambda item: (item[1], item[1] - item[0]))
    red_count, green_count = (int(red_sub.sum()), int(green_sub.sum()))
    volume_bars.append({'day_index': day, 'x': round(float(x), 2), 'height_px': int(bar_bottom - bar_top + 1),'top_y': int(bar_top + vtop), 'bottom_y': int(bar_bottom + vtop), 'color': 'red' if red_count >= green_count else'green'})
if len(volume_bars) != day_count:
    raise ValueError('incomplete volume state')
print('RESULT_JSON_START')
print(json.dumps({'stage': 'extract_volume', 'coverage': len(volume_bars), 'bars': volume_bars}, ensure_ascii=False))
print('RESULT_JSON_END')
\end{TraceCode}

\traceoutputheading

\Needspace{4\baselineskip}
\begin{TraceOutput}
{"stage": "extract_volume", "coverage": 48, "bars": [{"day_index": 1, "x": 31.0, "height_px": null, "top_y":
972, "bottom_y": 972, "color": "unknown", "valid": false}, {"day_index": 2, "x": 55.0, "height_px": null,
"top_y": 972, "bottom_y": 972, "color": "unknown", "valid": false}, {"day_index": 3, "x": 79.0, "height_px":
null, "top_y": 972, "bottom_y": 972, "color": "unknown", "valid": false}, {"day_index": 4, "x": 103.0,
"height_px": null, "top_y": 972, "bottom_y": 972, "color": "unknown", "valid": false}, {"day_index":
5, "x": 127.0, "height_px": null, "top_y": 972, "bottom_y": 972, "color": "unknown", "valid": false},
{"day_index": 6, "x": 151.0, "height_px": null, "top_y": 972, "bottom_y": 972, "color": "unknown", "valid":
false}, {"day_index": 7, "x": 175.0, "height_px": null, "top_y": 972, "bottom_y": 972, "color": "unknown",
"valid": false}, {"day_index": 8, "x": 199.0, "height_px": null, "top_y": 972, "bottom_y": 972, "color":
"unknown", "valid": false}, {"day_index": 9, "x": 223.0, "height_px": 14, "top_y": 889, "bottom_y": 902,
"color": "red"}, {"day_index": 10, "x": 247.0, "height_px": 4, "top_y": 890, "bottom_y": 893, "color":
"red"}, {"day_index": 11, "x": 271.0, "height_px": 4, "top_y": 894, "bottom_y": 897, "color": "red"},
{"day_index": 12, "x": 295.0, "height_px": 5, "top_y": 897, "bottom_y": 901, "color": "red"}, {"day_index":
13, "x": 319.0, "height_px": 22, "top_y": 879, "bottom_y": 900, "color": "red"}, {"day_index": 14, "x":
343.0, "height_px": 22, "top_y": 880, "bottom_y": 901, "color": "red"}, {"day_index": 15, "x": 367.0,
"height_px": 1, "top_y": 896, "bottom_y": 896, "color": "red"}, {"day_index": 16, "x": 391.0, "height_px":
null, "top_y": 972, "bottom_y": 972, "color": "unknown", "valid": false}, {"day_index": 17, "x": 415.0,
"height_px": null, "top_y": 972, "bottom_y": 972, "color": "unknown", "valid": false}, {"day_index":
18, "x": 439.0, "height_px": null, "top_y": 972, "bottom_y": 972, "color": "unknown", "valid": false},
{"day_index": 19, "x": 463.0, "height_px": null, "top_y": 972, "bottom_y": 972, "color": "unknown", "valid":
false}, {"day_index": 20, "x": 487.0, "height_px": null, "top_y": 972, "bottom_y": 972, "color": "unknown",
"valid": false}, {"day_index": 21, "x": 511.0, "height_px": null, "top_y": 972, "bottom_y": 972, "color":
"unknown", "valid": false}, {"day_index": 22, "x": 535.0, "height_px": null, "top_y": 972, "bottom_y":
972, "color": "unknown", "valid": false}, {"day_index": 23, "x": 559.0, "height_px": 1, "top_y": 882,
"bottom_y": 882, "color": "red"}, {"day_index": 24, "x": 583.0, "height_px": 1, "top_y": 882, "bottom_y":
882, "color": "red"}, {"day_index": 25, "x": 607.0, "height_px": 1, "top_y": 888, "bottom_y": 888, "color":
"red"}, {"day_index": 26, "x": 631.0, "height_px": 1, "top_y": 891, "bottom_y": 891, "color": "red"},
{"day_index": 27, "x": 655.0, "height_px": 1, "top_y": 887, "bottom_y": 887, "color": "red"}, {"day_index":
28, "x": 679.0, "height_px": 1, "top_y": 901, "bottom_y": 901, "color": "red"}, {"day_index": 29, "x":
703.0, "height_px": 3, "top_y": 916, "bottom_y": 918, "color": "red"}, {"day_index": 30, "x": 727.0,
"height_px": 1, "top_y": 880, "bottom_y": 880, "color": "red"}, {"day_index": 31, "x": 751.0, "height_px":
null, "top_y": 972, "bottom_y": 972, "color": "unknown", "valid": false}, {"day_index": 32, "x": 775.0,
"height_px": null, "top_y": 972, "bottom_y": 972, "color": "unknown", "valid": false}, {"day_index":
33, "x": 799.0, "height_px": null, "top_y": 972, "bottom_y": 972, "color": "unknown", "valid": false},
{"day_index": 34, "x": 823.0, "height_px": null, "top_y": 972, "bottom_y": 972, "color": "unknown", "valid":
false}, {"day_index": 35, "x": 847.0, "height_px": null, "top_y": 972, "bottom_y": 972, "color": "unknown",
"valid": false}, {"day_index": 36, "x": 871.0, "height_px": null, "top_y": 972, "bottom_y": 972, "color":
"unknown", "valid": false}, {"day_index": 37, "x": 895.0, "height_px": null, "top_y": 972, "bottom_y":
972, "color": "unknown", "valid": false}, {"day_index": 38, "x": 919.0, "height_px": null, "top_y": 972,
"bottom_y": 972, "color": "unknown", "valid": false}, {"day_index": 39, "x": 943.0, "height_px": null,
"top_y": 972, "bottom_y": 972, "color": "unknown", "valid": false}, {"day_index": 40, "x": 967.0, "height_px":
null, "top_y": 972, "bottom_y": 972, "color": "unknown", "valid": false}, {"day_index": 41, "x": 991.0,
"height_px": null, "top_y": 972, "bottom_y": 972, "color": "unknown", "valid": false}, {"day_index":
42, "x": 1015.0, "height_px": null, "top_y": 972, "bottom_y": 972, "color": "unknown", "valid": false},
{"day_index": 43, "x": 1039.0, "height_px": null, "top_y": 972, "bottom_y": 972, "color": "unknown",
"valid": false}, {"day_index": 44, "x": 1063.0, "height_px": null, "top_y": 972, "bottom_y": 972, "color":
"unknown", "valid": false}, {"day_index": 45, "x": 1087.0, "height_px": null, "top_y": 972, "bottom_y":
972, "color": "unknown", "valid": false}, {"day_index": 46, "x": 1111.0, "height_px": null, "top_y":
972, "bottom_y": 972, "color": "unknown", "valid": false}, {"day_index": 47, "x": 1135.0, "height_px":
null, "top_y": 972, "bottom_y": 972, "color": "unknown", "valid": false}, {"day_index": 48, "x": 1159.0,
"height_px": null, "top_y": 972, "bottom_y": 972, "color": "unknown", "valid": false}]}
\end{TraceOutput}

\subsubsection{Turn 5 of 7}

\traceplan{All 48 day centers are indexed, but several volume-bar colors within Days 9-21 remain unknown. Extract candle bodies and wick extents directly from the main chart at those centers to compare the daily lows reliably.}

\Needspace{4\baselineskip}
\begin{TraceCode}
from PIL import Image, ImageDraw
import re
import json
import numpy as np
REQUIRED_PUBLIC_STATE = ('image', 'chart_bbox', 'centers', 'pitch', 'color_masks', 'runs', 'volume_bars')
missing_public_state = [name for name in REQUIRED_PUBLIC_STATE if name not in globals()]
if missing_public_state:
    raise RuntimeError('missing prior public state: ' + ','.join(missing_public_state))
red_chart, green_chart = color_masks(image[y1:y2, x1:x2])
half = max(5, min(11, int(round(pitch * 0.42))))
def candle_candidate(mask):
    width = mask.shape[1]
    rows_count = mask.sum(1)
    if int(mask.sum()) < 5 or int(rows_count.max(initial=0)) <= 0:
        return None
    threshold = max(2, int(round(width * 0.28)))
    strong = np.where(rows_count >= threshold)[0]
    segments = runs(strong, 1)
    if not segments:
        segments = runs(np.where(rows_count >= 2)[0], 1)
    if not segments:
        return None
    ranked = []
    for top, bottom in segments:
        length = bottom - top + 1
        body_pixels = int(rows_count[top:bottom + 1].sum())
        score = float(body_pixels + 3 * length) * (0.58 if length <= 2 else 1.0)
        ranked.append((score, body_pixels, length, int(top), int(bottom)))
    score, body_pixels, body_length, body_top, body_bottom = max(ranked)
    middle = width // 2
    center_half = max(1, min(3, int(round(width * 0.14))))
    center_rows = np.where(mask[:, max(0, middle - center_half):min(width, middle + center_half + 1)].any(1))[0]
    center_segments = runs(center_rows, 3)
    nearby = [(top, bottom) for top, bottom in center_segments if bottom >= body_top - 4 and top <= body_bottom + 4]
    wick_top = min((top for top, _ in nearby), default=body_top)
    wick_bottom = max((bottom for _, bottom in nearby), default=body_bottom)
    return {'score': float(score + 0.2 * int(mask.sum())), 'body_top': body_top + y1, 'body_bottom': body_bottom + y1, 'wick_top': min(wick_top, body_top) + y1, 'wick_bottom': max(wick_bottom, body_bottom) + y1, 'pixels':int(mask.sum())}
raw_by_day = {}
for day, center in enumerate(centers, 1):
    local = int(round(center - x1))
    left = max(0, local - half)
    right = min(red_chart.shape[1], local + half + 1)
    scores = {}
    for color, cm in (('red', red_chart[:, left:right]), ('green', green_chart[:, left:right])):
        row_sum = cm.sum(1)
        if int(row_sum.max(initial=0)) <= 0:
            continue
        strong = np.where(row_sum >= max(2, int((right - left) * 0.28)))[0]
        if strong.size < 2:
            strong = np.where(row_sum >= 2)[0]
        all_rows = np.where(row_sum >= 1)[0]
        if strong.size and all_rows.size:
            scores[color] = {'score': int(row_sum[strong].sum()) + int(cm.sum()), 'body_top': int(strong.min() + y1),'body_bottom': int(strong.max() + y1), 'wick_top': int(all_rows.min() + y1), 'wick_bottom': int(all_rows.max() +y1)}
    if scores:
        volume_color = volume_bars[day - 1]['color']
        raw_color = volume_color if volume_color in scores else max(scores, key=lambda key: scores[key]['score'])
        raw_item = scores[raw_color]
        raw_by_day[day] = {'color': raw_color, 'body_top': raw_item['body_top'], 'body_bottom':raw_item['body_bottom'], 'wick_top': min((v['wick_top'] for v in scores.values())), 'wick_bottom':max((v['wick_bottom'] for v in scores.values()))}
candles = []
for day, center in enumerate(centers, 1):
    local = int(round(center - x1))
    left = max(0, local - half)
    right = min(red_chart.shape[1], local + half + 1)
    candidates = {}
    for color, mask in (('red', red_chart[:, left:right]), ('green', green_chart[:, left:right])):
        value = candle_candidate(mask)
        if value is not None:
            candidates[color] = value
    if not candidates:
        continue
    ordered = sorted(candidates, key=lambda name: candidates[name]['score'], reverse=True)
    best_color = ordered[0]
    best_score = candidates[best_color]['score']
    second_score = candidates[ordered[1]]['score'] if len(ordered) > 1 else 0.0
    volume_color = volume_bars[day - 1]['color']
    color = volume_color if volume_color in candidates and best_score / max(1.0, second_score) < 1.55 else best_color
    item = candidates[color]
    geometry = raw_by_day.get(day) if raw_by_day.get(day, {}).get('color') == color else item
    top, bottom = (int(geometry['body_top']), int(geometry['body_bottom']))
    candles.append({'day_index': day, 'x': round(float(center), 2), 'color': color, 'type': '\u9633' if color == 'red' else '\u9634', 'body_y': [top, bottom], 'open_y': bottom if color == 'red' else top, 'close_y': top if color == 'red' else bottom, 'high_y': int(geometry['wick_top']), 'low_y': int(geometry['wick_bottom']), 'confidence':round(float(best_score / max(1.0, best_score + second_score)), 4)})
print('RESULT_JSON_START')
print(json.dumps({'stage': 'extract_candles', 'coverage': len(candles), 'candles': candles}, ensure_ascii=False))
print('RESULT_JSON_END')
\end{TraceCode}

\traceoutputheading

\Needspace{4\baselineskip}
\begin{TraceOutput}
{"stage": "extract_candles", "coverage": 48, "candles": [{"day_index": 1, "x": 31.0, "color": "red",
"type": "\u9633", "body_y": [200, 258], "open_y": 258, "close_y": 200, "high_y": 171, "low_y": 260, "confidence":
1.0}, {"day_index": 2, "x": 55.0, "color": "red", "type": "\u9633", "body_y": [177, 209], "open_y": 209,
"close_y": 177, "high_y": 156, "low_y": 227, "confidence": 0.646}, {"day_index": 3, "x": 79.0, "color":
"green", "type": "\u9634", "body_y": [178, 219], "open_y": 178, "close_y": 219, "high_y": 165, "low_y":
276, "confidence": 0.9875}, {"day_index": 4, "x": 103.0, "color": "green", "type": "\u9634", "body_y":
[222, 277], "open_y": 222, "close_y": 277, "high_y": 173, "low_y": 282, "confidence": 0.9345}, {"day_index":
5, "x": 127.0, "color": "green", "type": "\u9634", "body_y": [280, 282], "open_y": 280, "close_y": 282,
"high_y": 184, "low_y": 305, "confidence": 0.6452}, {"day_index": 6, "x": 151.0, "color": "red", "type":
"\u9633", "body_y": [254, 283], "open_y": 283, "close_y": 254, "high_y": 56, "low_y": 287, "confidence":
0.7156}, {"day_index": 7, "x": 175.0, "color": "green", "type": "\u9634", "body_y": [271, 272], "open_y":
271, "close_y": 272, "high_y": 57, "low_y": 278, "confidence": 0.675}, {"day_index": 8, "x": 199.0, "color":
"red", "type": "\u9633", "body_y": [258, 270], "open_y": 270, "close_y": 258, "high_y": 57, "low_y":
302, "confidence": 0.6133}, {"day_index": 9, "x": 223.0, "color": "green", "type": "\u9634", "body_y":
[285, 302], "open_y": 285, "close_y": 302, "high_y": 281, "low_y": 302, "confidence": 0.9748}, {"day_index":
10, "x": 247.0, "color": "green", "type": "\u9634", "body_y": [306, 311], "open_y": 306, "close_y": 311,
"high_y": 306, "low_y": 311, "confidence": 0.7242}, {"day_index": 11, "x": 271.0, "color": "red", "type":
"\u9633", "body_y": [258, 313], "open_y": 313, "close_y": 258, "high_y": 57, "low_y": 320, "confidence":
0.7358}, {"day_index": 12, "x": 295.0, "color": "red", "type": "\u9633", "body_y": [234, 275], "open_y":
275, "close_y": 234, "high_y": 56, "low_y": 275, "confidence": 0.8917}, {"day_index": 13, "x": 319.0,
"color": "green", "type": "\u9634", "body_y": [234, 252], "open_y": 234, "close_y": 252, "high_y": 229,
"low_y": 252, "confidence": 0.9813}, {"day_index": 14, "x": 343.0, "color": "red", "type": "\u9633",
"body_y": [228, 229], "open_y": 229, "close_y": 228, "high_y": 200, "low_y": 288, "confidence": 1.0},
{"day_index": 15, "x": 367.0, "color": "red", "type": "\u9633", "body_y": [221, 250], "open_y": 250,
"close_y": 221, "high_y": 186, "low_y": 254, "confidence": 1.0}, {"day_index": 16, "x": 391.0, "color":
"red", "type": "\u9633", "body_y": [185, 249], "open_y": 249, "close_y": 185, "high_y": 170, "low_y":
255, "confidence": 1.0}, {"day_index": 17, "x": 415.0, "color": "green", "type": "\u9634", "body_y":
[184, 227], "open_y": 184, "close_y": 227, "high_y": 153, "low_y": 249, "confidence": 0.9817}, {"day_index":
18, "x": 439.0, "color": "green", "type": "\u9634", "body_y": [229, 292], "open_y": 229, "close_y": 292,
"high_y": 171, "low_y": 293, "confidence": 0.9793}, {"day_index": 19, "x": 463.0, "color": "red", "type":
"\u9633", "body_y": [252, 293], "open_y": 293, "close_y": 252, "high_y": 237, "low_y": 296, "confidence":
1.0}, {"day_index": 20, "x": 487.0, "color": "green", "type": "\u9634", "body_y": [286, 313], "open_y":
286, "close_y": 313, "high_y": 255, "low_y": 359, "confidence": 0.9892}, {"day_index": 21, "x": 511.0,
"color": "red", "type": "\u9633", "body_y": [266, 348], "open_y": 348, "close_y": 266, "high_y": 260,
"low_y": 353, "confidence": 0.8601}, {"day_index": 22, "x": 535.0, "color": "green", "type": "\u9634",
"body_y": [262, 355], "open_y": 262, "close_y": 355, "high_y": 261, "low_y": 361, "confidence": 0.9439},
{"day_index": 23, "x": 559.0, "color": "green", "type": "\u9634", "body_y": [357, 387], "open_y": 357,
"close_y": 387, "high_y": 349, "low_y": 389, "confidence": 0.9854}, {"day_index": 24, "x": 583.0, "color":
"green", "type": "\u9634", "body_y": [388, 434], "open_y": 388, "close_y": 434, "high_y": 372, "low_y":
444, "confidence": 0.9906}, {"day_index": 25, "x": 607.0, "color": "green", "type": "\u9634", "body_y":
[436, 496], "open_y": 436, "close_y": 496, "high_y": 429, "low_y": 500, "confidence": 0.9851}, {"day_index":
26, "x": 631.0, "color": "green", "type": "\u9634", "body_y": [497, 551], "open_y": 497, "close_y": 551,
"high_y": 497, "low_y": 572, "confidence": 0.9795}, {"day_index": 27, "x": 655.0, "color": "green", "type":
"\u9634", "body_y": [550, 630], "open_y": 550, "close_y": 630, "high_y": 549, "low_y": 633, "confidence":
0.9598}, {"day_index": 28, "x": 679.0, "color": "green", "type": "\u9634", "body_y": [591, 669], "open_y":
591, "close_y": 669, "high_y": 557, "low_y": 679, "confidence": 0.9827}, {"day_index": 29, "x": 703.0,
"color": "red", "type": "\u9633", "body_y": [353, 671], "open_y": 671, "close_y": 353, "high_y": 18,
"low_y": 746, "confidence": 0.976}, {"day_index": 30, "x": 727.0, "color": "green", "type": "\u9634",
"body_y": [447, 548], "open_y": 447, "close_y": 548, "high_y": 447, "low_y": 558, "confidence": 0.9914},
{"day_index": 31, "x": 751.0, "color": "red", "type": "\u9633", "body_y": [28, 561], "open_y": 561, "close_y":
28, "high_y": 18, "low_y": 566, "confidence": 0.6905}, {"day_index": 32, "x": 775.0, "color": "red",
"type": "\u9633", "body_y": [19, 549], "open_y": 549, "close_y": 19, "high_y": 18, "low_y": 562, "confidence":
0.9441}, {"day_index": 33, "x": 799.0, "color": "red", "type": "\u9633", "body_y": [19, 533], "open_y":
533, "close_y": 19, "high_y": 18, "low_y": 563, "confidence": 0.9145}, {"day_index": 34, "x": 823.0,
"color": "red", "type": "\u9633", "body_y": [264, 450], "open_y": 450, "close_y": 264, "high_y": 255,
"low_y": 461, "confidence": 0.9429}, {"day_index": 35, "x": 847.0, "color": "green", "type": "\u9634",
"body_y": [192, 361], "open_y": 192, "close_y": 361, "high_y": 122, "low_y": 411, "confidence": 0.996},
{"day_index": 36, "x": 871.0, "color": "red", "type": "\u9633", "body_y": [302, 368], "open_y": 368,
"close_y": 302, "high_y": 281, "low_y": 418, "confidence": 0.9151}, {"day_index": 37, "x": 895.0, "color":
"green", "type": "\u9634", "body_y": [306, 398], "open_y": 306, "close_y": 398, "high_y": 301, "low_y":
424, "confidence": 0.9943}, {"day_index": 38, "x": 919.0, "color": "red", "type": "\u9633", "body_y":
[345, 427], "open_y": 427, "close_y": 345, "high_y": 335, "low_y": 427, "confidence": 1.0}, {"day_index":
39, "x": 943.0, "color": "green", "type": "\u9634", "body_y": [346, 391], "open_y": 346, "close_y": 391,
"high_y": 345, "low_y": 431, "confidence": 0.9887}, {"day_index": 40, "x": 967.0, "color": "red", "type":
"\u9633", "body_y": [346, 433], "open_y": 433, "close_y": 346, "high_y": 301, "low_y": 433, "confidence":
1.0}, {"day_index": 41, "x": 991.0, "color": "green", "type": "\u9634", "body_y": [317, 380], "open_y":
317, "close_y": 380, "high_y": 306, "low_y": 435, "confidence": 0.9827}, {"day_index": 42, "x": 1015.0,
"color": "green", "type": "\u9634", "body_y": [378, 393], "open_y": 378, "close_y": 393, "high_y": 374,
"low_y": 437, "confidence": 0.9426}, {"day_index": 43, "x": 1039.0, "color": "green", "type": "\u9634",
"body_y": [387, 474], "open_y": 387, "close_y": 474, "high_y": 382, "low_y": 475, "confidence": 0.993},
{"day_index": 44, "x": 1063.0, "color": "red", "type": "\u9633", "body_y": [442, 469], "open_y": 469,
"close_y": 442, "high_y": 393, "low_y": 493, "confidence": 0.8178}, {"day_index": 45, "x": 1087.0, "color":
"green", "type": "\u9634", "body_y": [504, 573], "open_y": 504, "close_y": 573, "high_y": 401, "low_y":
574, "confidence": 0.9933}, {"day_index": 46, "x": 1111.0, "color": "green", "type": "\u9634", "body_y":
[575, 593], "open_y": 575, "close_y": 593, "high_y": 414, "low_y": 593, "confidence": 0.9593}, {"day_index":
47, "x": 1135.0, "color": "green", "type": "\u9634", "body_y": [616, 660], "open_y": 616, "close_y":
660, "high_y": 428, "low_y": 673, "confidence": 0.9895}, {"day_index": 48, "x": 1159.0, "color": "red",
"type": "\u9633", "body_y": [644, 658], "open_y": 658, "close_y": 644, "high_y": 12, "low_y": 659, "confidence":
1.0}]}
\end{TraceOutput}

\subsubsection{Turn 6 of 7}

\traceplan{The disclosed candle table provides a low\_y value and day\_index for every day, including all 13 candidates from Day 9 through Day 21. Select the largest low\_y value, using the earliest day to break a tie, because a larger screen-y coordinate represents a lower price.}

\Needspace{4\baselineskip}
\begin{TraceCode}
from PIL import Image, ImageDraw
import numpy as np
import re
import json
REQUIRED_PUBLIC_STATE = ('day_count', 'candles')
missing_public_state = [name for name in REQUIRED_PUBLIC_STATE if name not in globals()]
if missing_public_state:
    raise RuntimeError('missing prior public state: ' + ','.join(missing_public_state))
TASK = 'interval_lowest_day'
PARAMS = {'end': 21, 'start': 9}
c_by_day = {int(row['day_index']): row for row in candles}
v_by_day = {}
ma_by_day = {}
days = list(range(9, 22))
evidence_days = days.copy()
public_rows = [{'day_index': d, 'metric': c_by_day[d]['low_y']} for d in days]
selected = max(public_rows, key=lambda row: (row['metric'], -row['day_index']))
answer = str(selected['day_index'])
evidence_days = [selected['day_index']]
flat_rows = []
stack = [public_rows]
while stack:
    item = stack.pop()
    if isinstance(item, dict):
        if 'day_index' in item or 'start_day' in item or 'end_day' in item:
            flat_rows.append(item)
        else:
            stack.extend(reversed(list(item.values())))
    elif isinstance(item, list):
        stack.extend(reversed(item))
candidate_days = sorted({int(row[key]) for row in flat_rows for key in ('day_index', 'start_day', 'end_day') if row.get(key) is not None})
checked_pairs = []
for row in flat_rows:
    if row.get('previous_day') is not None and row.get('day_index') is not None:
        checked_pairs.append([int(row['previous_day']), int(row['day_index'])])
    elif False and row.get('day_index') is not None and (int(row['day_index']) > 1):
        checked_pairs.append([int(row['day_index']) - 1, int(row['day_index'])])
checked_pairs = sorted({tuple(pair) for pair in checked_pairs})
scope_days = [9, 10, 11, 12, 13, 14, 15, 16, 17, 18, 19, 20, 21] or candidate_days or list(range(1, day_count + 1))
positive_days = sorted(set((int(d) for d in evidence_days if d is not None)))
evidence_contract = {'version': 'semantic_evidence_v2', 'kind': 'interval_predicate', 'modalities': ['candle'],'scope_days': scope_days, 'candidate_days': candidate_days, 'positive_days': positive_days, 'checked_pairs':[list(pair) for pair in checked_pairs], 'negative_case': str(answer) == '0' and (not positive_days),'negative_support': {'checked_days': candidate_days, 'checked_pairs': [list(pair) for pair in checked_pairs],'all_predicates_false': str(answer) == '0' and (not positive_days)}, 'field_support': {}, 'field_modalities': {}}
task_result = {'stage': 'compute_task', 'task': TASK, 'parameters': PARAMS, 'public_rows': public_rows,'evidence_days': positive_days, 'evidence_contract': evidence_contract, 'answer': str(answer)}
print('RESULT_JSON_START')
print(json.dumps(task_result, ensure_ascii=False, separators=(',', ':')))
print('RESULT_JSON_END')
\end{TraceCode}

\traceoutputheading

\Needspace{4\baselineskip}
\begin{TraceOutput}
{"stage": "compute_task", "task": "interval_lowest_day", "parameters": {"end": 21, "start": 9}, "public_rows":
[{"day_index": 9, "metric": 302}, {"day_index": 10, "metric": 311}, {"day_index": 11, "metric": 320},
{"day_index": 12, "metric": 275}, {"day_index": 13, "metric": 252}, {"day_index": 14, "metric": 288},
{"day_index": 15, "metric": 254}, {"day_index": 16, "metric": 255}, {"day_index": 17, "metric": 249},
{"day_index": 18, "metric": 293}, {"day_index": 19, "metric": 296}, {"day_index": 20, "metric": 359},
{"day_index": 21, "metric": 353}], "evidence_days": [20], "evidence_contract": {"version": "semantic_evidence_v2",
"kind": "interval_predicate", "modalities": ["candle"], "scope_days": [9, 10, 11, 12, 13, 14, 15, 16,
17, 18, 19, 20, 21], "candidate_days": [9, 10, 11, 12, 13, 14, 15, 16, 17, 18, 19, 20, 21], "positive_days":
[20], "checked_pairs": [], "negative_case": false, "negative_support": {"checked_days": [9, 10, 11, 12,
13, 14, 15, 16, 17, 18, 19, 20, 21], "checked_pairs": [], "all_predicates_false": false}, "field_support":
{}, "field_modalities": {}}, "answer": "20"}
\end{TraceOutput}

\subsubsection{Turn 7 of 7}

\traceplan{The comparison across Days 9-21 identifies Day 20 as the sole evidence day and returns answer=20. Render answer-level evidence that shows the full checked interval and highlights the unique selected day.}

\Needspace{4\baselineskip}
\begin{TraceCode}
import numpy as np
import json
import re
from PIL import Image, ImageDraw
result_image = Image.fromarray(image.astype("uint8")).convert("RGB")
draw = ImageDraw.Draw(result_image)
contract = task_result['evidence_contract']
kind = 'extrema'
cmap = {'primary': (255, 0, 255), 'secondary': (0, 180, 255), 'negative': (235, 70, 50), 'context': (255, 170, 0),'muted': (70, 170, 255)}
regions = []
def add_region(name, boxes, days=None, relation=None):
    clean = [[int(v) for v in box] for box in boxes if len(box) == 4 and box[2] > box[0] and (box[3] > box[1])]
    if clean:
        regions.append({'artifact': name, 'day_indices': sorted(set((int(d) for d in days or [] if d is not None))),'relation': relation, 'pixel_regions': clean})
def candle_box(day, body_only=False):
    row = c_by_day.get(int(day))
    if not row:
        return None
    x = int(round(centers[int(day) - 1]))
    top, bottom = row['body_y'] if body_only else (row['high_y'], row['low_y'])
    return [int(x - pitch * 0.34), int(top) - 2, int(x + pitch * 0.34), int(bottom) + 3]
def volume_box(day):
    row = v_by_day.get(int(day))
    if not row or row.get('height_px') is None:
        return None
    x = int(round(centers[int(day) - 1]))
    return [int(x - pitch * 0.34), int(row['top_y']) - 2, int(x + pitch * 0.34), int(row['bottom_y']) + 2]
def ma_boxes(day, names=None):
    values = ma_by_day.get(int(day), {})
    x = int(round(centers[int(day) - 1]))
    boxes = []
    for name in names or sorted(values):
        if name not in values:
            continue
        y = int(round(values[name]))
        boxes.append([x - 5, y - 5, x + 6, y + 6])
    return boxes
flat = []
stack = [task_result['public_rows']]
semantic_annotations = []
negative_rendered = False
while stack:
    item = stack.pop()
    if isinstance(item, dict):
        if 'day_index' in item or 'start_day' in item or 'end_day' in item:
            flat.append(item)
        else:
            stack.extend(reversed(list(item.values())))
    elif isinstance(item, list):
        stack.extend(reversed(item))
by_day = {int(row['day_index']): row for row in flat if row.get('day_index') is not None}
scope = [int(d) for d in contract['scope_days'] if 1 <= int(d) <= day_count]
positive = [int(d) for d in contract['positive_days'] if 1 <= int(d) <= day_count]
candidate = [int(d) for d in contract['candidate_days'] if 1 <= int(d) <= day_count]
if scope:
    left, right = (min(scope), max(scope))
    x_left = int(round(centers[left - 1]))
    x_right = int(round(centers[right - 1]))
    scope_top = int(volume_bbox[1]) if set(contract.get('modalities') or []) == {'volume'} else int(chart_bbox[1])
    y = scope_top + 7
    draw.line((x_left, y, x_right, y), fill=cmap['context'], width=3)
    draw.line((x_left, y - 6, x_left, y + 6), fill=cmap['context'], width=3)
    draw.line((x_right, y - 6, x_right, y + 6), fill=cmap['context'], width=3)
    draw.text((x_left + 3, y + 3), 'scope D%d-D%d' % (left, right), fill=cmap['context'])
    add_region('checked_scope', [[min(x_left, x_right) - 2, y - 7, max(x_left, x_right) + 3, y + 18]], scope,'bounded_check_range')
days = [] if contract['negative_case'] else positive or candidate or scope
boxes = []
for day in days:
    local = []
    if 'candle' in contract['modalities']:
        box = candle_box(day)
        local += [box] if box else []
    if 'volume' in contract['modalities']:
        box = volume_box(day)
        local += [box] if box else []
    if 'ma' in contract['modalities']:
        local += ma_boxes(day)
    color = cmap['primary'] if day in positive else cmap['negative'] if contract['negative_case'] else cmap['secondary']
    for box in local:
        draw.rectangle(tuple(box), outline=color, width=3 if day in positive else 2)
    boxes.extend(local)
add_region(kind, boxes, days, 'task_relevant_objects')
if contract['negative_case'] and (not negative_rendered):
    checked = candidate or scope
    marks = []
    y = int(chart_bbox[1]) + 28
    for day in checked:
        x = int(round(centers[day - 1]))
        box = [x - 4, y - 4, x + 5, y + 5]
        draw.line((box[0], box[1], box[2], box[3]), fill=cmap['negative'], width=2)
        draw.line((box[0], box[3], box[2], box[1]), fill=cmap['negative'], width=2)
        marks.append(box)
    if marks:
        draw.text((marks[0][0], y + 8), '0 matches / %d checked' % len(checked), fill=cmap['negative'])
    add_region('negative_predicate_checks', marks, checked, 'all_checked_predicates_false')
if not regions:
    raise ValueError('semantic evidence renderer produced no answer support')
from PIL import ImageFont
try:
    semantic_font = ImageFont.truetype('DejaVuSans.ttf', 14)
except Exception:
    semantic_font = ImageFont.load_default()
semantic_summary_bounds = []
def draw_semantic_label(text, x, y, color=(145, 0, 220)):
    x, y = (int(x), int(y))
    bbox = draw.textbbox((x, y), text, font=semantic_font)
    box = [bbox[0] - 5, bbox[1] - 4, bbox[2] + 5, bbox[3] + 4]
    draw.rectangle(tuple(box), fill=(255, 255, 255), outline=color, width=2)
    draw.text((x, y), text, fill=color, font=semantic_font)
    semantic_summary_bounds.append(box)
    semantic_annotations.append({'type': 'task_summary', 'task': TASK, 'label': text})
rows = list(task_result['public_rows'])
selected = int(task_result['answer'])
highest = TASK != 'interval_lowest_day'
field = 'high_y' if highest else 'low_y'
values = []
for row in rows:
    day = int(row['day_index'])
    value = float(row.get(field, -float(row['metric']) if highest and row.get('metric') is not None else row.get('metric')))
    x = int(round(centers[day - 1]))
    y = int(round(value))
    color = (255, 0, 255) if day == selected else (0, 180, 255)
    draw.ellipse((x - 4, y - 4, x + 5, y + 5), outline=color, fill=color)
    values.append((day, y))
chosen_y = dict(values)[selected]
draw.line((int(round(centers[min((d for d, _ in values)) - 1] - pitch * 0.4)), chosen_y, int(round(centers[max((d for d, _ in values)) - 1] + pitch * 0.4)), chosen_y), fill=(255, 0, 255), width=3)
panel_x = max(8, int(chart_bbox[0]) + 8)
panel_y = max(36, int(chart_bbox[1]) + 78)
panel_right = min(result_image.width - 6, int(chart_bbox[2]) - 4)
available = max(48, panel_right - panel_x)
max_cols = min(10, max(1, len(values)), max(1, available // 76))
rows_count = (len(values) + max_cols - 1) // max_cols
cell_w = max(62, min(92, int(available / max_cols)))
cell_h = 30
panel_bbox = [panel_x - 4, panel_y - 5, min(panel_right, panel_x + cell_w * max_cols + 5), min(result_image.height -5, panel_y + cell_h * rows_count + 7)]
draw.rectangle(tuple(panel_bbox), fill=(255, 255, 255), outline=(145, 0, 220), width=2)
semantic_summary_bounds.append(panel_bbox)
checklist = []
for index, (day, y) in enumerate(values):
    row_i, col_i = divmod(index, max_cols)
    xx = panel_x + col_i * cell_w
    yy = panel_y + row_i * cell_h
    chosen = day == selected
    color = (255, 0, 255) if chosen else (0, 140, 185)
    draw.rectangle((xx, yy, xx + cell_w - 3, yy + cell_h - 2), fill=(255, 232, 255) if chosen else (238, 249, 252),outline=color, width=3 if chosen else 1)
    draw.text((xx + 3, yy + 5), 'D%d:y%d' % (day, y), fill=color, font=semantic_font)
    checklist.append({'day_index': day, 'y': y, 'selected': chosen})
draw_semantic_label(('HIGHEST' if highest else 'LOWEST') + ' D%d y=%d / %d CANDIDATES' % (selected, chosen_y,len(values)), int(chart_bbox[0]) + 8, int(chart_bbox[1]) + 48)
semantic_annotations.append({'type': 'extrema_candidate_geometry', 'selected_day': selected, 'selected_y': chosen_y,'candidate_count': len(values), 'kind': 'highest' if highest else 'lowest', 'checklist': checklist, 'panel_bbox':panel_bbox, 'all_candidate_endpoints_visible': True, 'candidate_index_checklist_visible': True})
mods=set(contract.get("modalities") or [])
if mods=={"volume"}:
    y_top,y_bottom=int(volume_bbox[1]),int(volume_bbox[3])
elif "volume" in mods and ("candle" in mods or "ma" in mods):
    y_top,y_bottom=int(chart_bbox[1]),int(volume_bbox[3])
else:
    y_top,y_bottom=int(chart_bbox[1]),int(chart_bbox[3])
visible_scope=sorted({int(d) for key in ("scope_days","candidate_days","positive_days") for d in (contract.get(key) or []) if 1<=int(d)<=day_count})
if visible_scope:
    x_left=int(round(centers[min(visible_scope)-1]-pitch*.7));x_right=int(round(centers[max(visible_scope)-1]+pitch*.7))
else:
    x_left,x_right=int(chart_bbox[0]),int(chart_bbox[2])
pad=max(6,int(round(pitch*.35)))
if semantic_summary_bounds:
    x_left=min([x_left]+[int(box[0])-pad for box in semantic_summary_bounds]); x_right=max([x_right]+[int(box[2])+pad for box in semantic_summary_bounds])
    y_top=min([y_top]+[int(box[1])-pad for box in semantic_summary_bounds]); y_bottom=max([y_bottom]+[int(box[3])+pad for box in semantic_summary_bounds])
evidence_crop_bbox=[max(0,x_left-pad),max(0,y_top-pad),min(result_image.width,x_right+pad),min(result_image.height,y_bottom+pad)]
result_image=result_image.crop(tuple(evidence_crop_bbox))
payload = {'stage': 'render_semantic_answer_evidence_v2', 'task': TASK, 'answer': task_result['answer'],'evidence_contract': contract, 'evidence_regions': regions, 'semantic_annotations': semantic_annotations,'whole_panel_fallback': False}
payload['source_coordinate_frame']='reviewed_focus_crop'; payload['focus_origin_in_original']=focus_origin;payload['result_image_crop_bbox']=evidence_crop_bbox; payload['result_image_size']=list(result_image.size);payload['semantic_summary_bounds']=semantic_summary_bounds
print('RESULT_JSON_START')
print(json.dumps(payload, ensure_ascii=False, separators=(',', ':')))
print('RESULT_JSON_END')
\end{TraceCode}

\traceoutputheading

\Needspace{4\baselineskip}
\begin{TraceOutput}
{"stage": "render_semantic_answer_evidence_v2", "task": "interval_lowest_day", "answer": "20", "evidence_contract":
{"version": "semantic_evidence_v2", "kind": "interval_predicate", "modalities": ["candle"], "scope_days":
[9, 10, 11, 12, 13, 14, 15, 16, 17, 18, 19, 20, 21], "candidate_days": [9, 10, 11, 12, 13, 14, 15, 16,
17, 18, 19, 20, 21], "positive_days": [20], "checked_pairs": [], "negative_case": false, "negative_support":
{"checked_days": [9, 10, 11, 12, 13, 14, 15, 16, 17, 18, 19, 20, 21], "checked_pairs": [], "all_predicates_false":
false}, "field_support": {}, "field_modalities": {}}, "evidence_regions": [{"artifact": "checked_scope",
"day_indices": [9, 10, 11, 12, 13, 14, 15, 16, 17, 18, 19, 20, 21], "relation": "bounded_check_range",
"pixel_regions": [[221, 12, 514, 37]]}, {"artifact": "extrema", "day_indices": [20], "relation": "task_relevant_objects",
"pixel_regions": [[478, 253, 495, 362]]}], "semantic_annotations": [{"type": "task_summary", "task":
"interval_lowest_day", "label": "LOWEST D20 y=359 / 13 CANDIDATES"}, {"type": "extrema_candidate_geometry",
"selected_day": 20, "selected_y": 359, "candidate_count": 13, "kind": "lowest", "checklist": [{"day_index":
9, "y": 302, "selected": false}, {"day_index": 10, "y": 311, "selected": false}, {"day_index": 11, "y":
320, "selected": false}, {"day_index": 12, "y": 275, "selected": false}, {"day_index": 13, "y": 252,
"selected": false}, {"day_index": 14, "y": 288, "selected": false}, {"day_index": 15, "y": 254, "selected":
false}, {"day_index": 16, "y": 255, "selected": false}, {"day_index": 17, "y": 249, "selected": false},
{"day_index": 18, "y": 293, "selected": false}, {"day_index": 19, "y": 296, "selected": false}, {"day_index":
20, "y": 359, "selected": true}, {"day_index": 21, "y": 353, "selected": false}], "panel_bbox": [4, 85,
933, 157], "all_candidate_endpoints_visible": true, "candidate_index_checklist_visible": true}], "whole_panel_fallback":
false, "source_coordinate_frame": "reviewed_focus_crop", "focus_origin_in_original": [0, 756], "result_image_crop_bbox":
[0, 4, 949, 789], "result_image_size": [949, 785], "semantic_summary_bounds": [[4, 85, 933, 157], [3,
58, 189, 76]]}
\end{TraceOutput}

\traceimage[0.5]{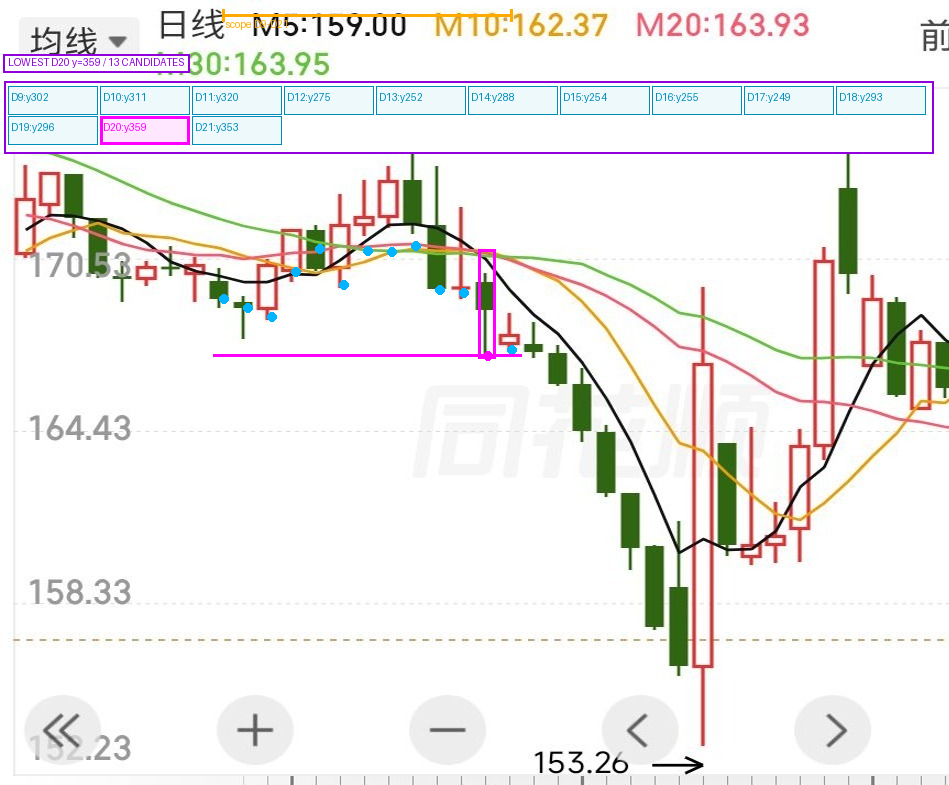}{Answer-level evidence for the lowest-price day.}

\subsubsection{Final Evidence Review and Answer}

The evidence image lists low\_y for every day from D9 through D21 and highlights D20:y359 with the label 'LOWEST D20 y=359 / 13 CANDIDATES.' This agrees with answer=20, so the final response can now be issued.

\begin{center}\Large\bfseries 20\end{center}

\subsection{Example 2: First Candlestick Date}

\traceimage[0.7]{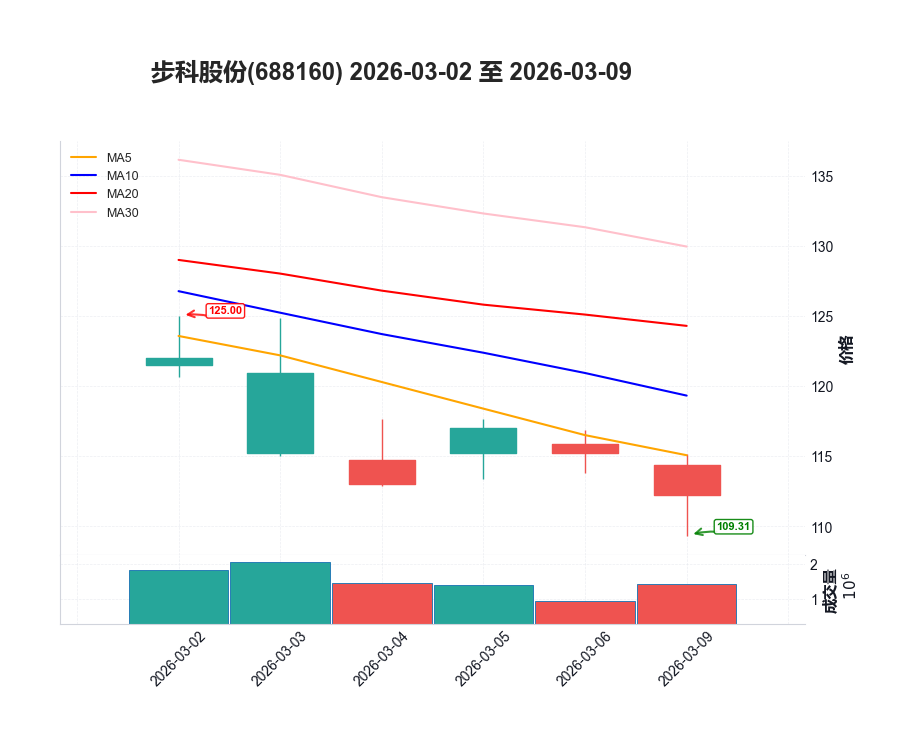}{Input chart for Example 2.}

\paragraph{Question.}

This chart shows the share-price history of an A-share-listed company. What date corresponds to the first candlestick in the chart? Use the format '1997-01-01'.

\paragraph{Replay Result.} \fbox{\textbf{2026-03-02}}

\subsubsection{Turn 1 of 2}

\traceplan{The title reads 'Kinco Automation (688160), 2026-03-02 to 2026-03-09,' above a short candlestick-and-volume chart with only a few trading days. Confirm the short-period layout and image dimensions before locating the first date.}

\Needspace{4\baselineskip}
\begin{TraceCode}
from PIL import Image
import numpy as np, json, re
image = np.asarray(Image.open(input_path).convert('RGB'))
h, w = image.shape[:2]
if h / max(w, 1) >= 1.45:
    raise ValueError('expected a visible short K-line layout')
rgb = image.astype(np.int16)
r, g, b = (rgb[:, :, 0], rgb[:, :, 1], rgb[:, :, 2])
red = (r > 190) & (g > 25) & (g < 145) & (b < 145) & (r - g > 70)
green = (r < 130) & (g > 100) & (b > 90) & (g - r > 35) & (b - r > 35)
candle_mask = red | green
chart_bbox = [0, int(h * 0.18), w, int(h * 0.75)]
volume_bbox = [0, int(h * 0.68), w, int(h * 0.88)]
print('RESULT_JSON_START')
print(json.dumps({'stage': 'inspect_layout', 'layout_kind': 'short_kline', 'image_size': [w, h], 'chart_bbox':chart_bbox, 'volume_bbox': volume_bbox}, ensure_ascii=False))
print('RESULT_JSON_END')
\end{TraceCode}

\traceoutputheading

\Needspace{4\baselineskip}
\begin{TraceOutput}
{"stage": "inspect_layout", "layout_kind": "short_kline", "image_size": [920, 749], "chart_bbox": [0,
134, 920, 561], "volume_bbox": [0, 509, 920, 659]}
\end{TraceOutput}

\subsubsection{Turn 2 of 2}

\traceplan{The disclosed result confirms a 920 x 749 short-period chart whose main panel spans approximately y=134-561. Crop and enlarge the title region above it so the visible date range can be read directly.}

\Needspace{4\baselineskip}
\begin{TraceCode}
import re
from PIL import Image
import numpy as np, json
base = Image.open(input_path).convert('RGB')
arr = np.asarray(base)
h, w = arr.shape[:2]
top = arr[:max(1, int(h * 0.18))]
dark = top.max(2) < 115
ys, xs = np.where(dark)
if xs.size < 50:
    raise ValueError('visible title text not found')
bbox = [max(0, int(xs.min()) - 18), max(0, int(ys.min()) - 14), min(w, int(xs.max()) + 19), min(h, int(ys.max()) +15)]
crop = base.crop(tuple(bbox))
result_image = crop.resize((crop.width * 3, crop.height * 3))
payload = {'stage': 'crop_visible_title_for_reading', 'crop_bbox': bbox, 'output_size': list(result_image.size),'evidence_regions': [{'artifact': 'visible_title_text', 'pixel_regions': [bbox]}], 'visual_read_required': True}
print('RESULT_JSON_START')
print(json.dumps(payload, ensure_ascii=False))
print('RESULT_JSON_END')
\end{TraceCode}

\traceoutputheading

\Needspace{4\baselineskip}
\begin{TraceOutput}
{"stage": "crop_visible_title_for_reading", "crop_bbox": [134, 46, 649, 99], "output_size": [1545, 159],
"evidence_regions": [{"artifact": "visible_title_text", "pixel_regions": [[134, 46, 649, 99]]}], "visual_read_required":
true}
\end{TraceOutput}

\traceimage[0.7]{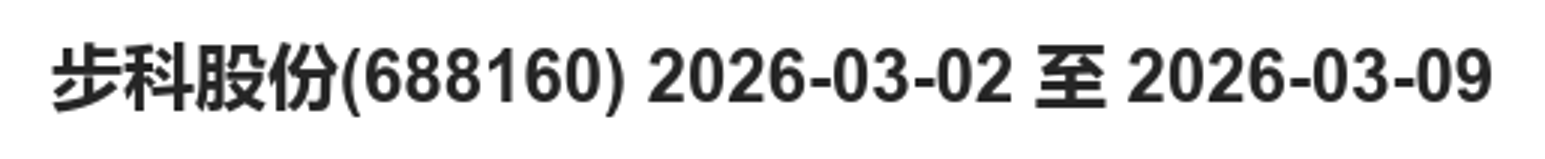}{Enlarged title evidence used to read the first date.}

\subsubsection{Final Evidence Review and Answer}

The enlarged title clearly gives the range 2026-03-02 to 2026-03-09. Because the question asks for the date of the first candlestick, the answer is the interval's first day, 2026-03-02.

\begin{center}\Large\bfseries 2026-03-02\end{center}

\subsection{Example 3: Maximum-Volume Candlestick}

\traceimage[0.78]{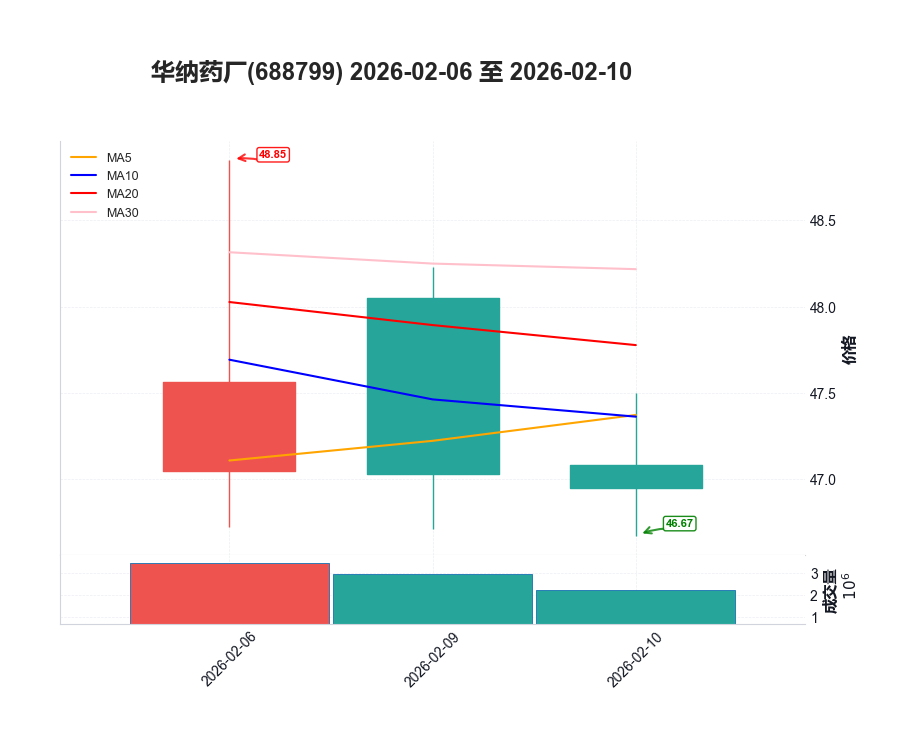}{Input chart for Example 3.}

\paragraph{Question.}

Counting candlesticks from left to right, which candlestick has the highest trading volume? Answer with the number only.

\paragraph{Replay Result.} \fbox{\textbf{1}}

\subsubsection{Turn 1 of 6}

\traceplan{Three volume bars are visible in the lower panel and align with three candlesticks from left to right; the first bar appears tallest. Locate the chart and volume-panel boundaries before measuring the bars.}

\Needspace{4\baselineskip}
\begin{TraceCode}
from PIL import Image
import numpy as np, json, re
image = np.asarray(Image.open(input_path).convert('RGB'))
h, w = image.shape[:2]
if h / max(w, 1) >= 1.45:
    raise ValueError('expected a visible short K-line layout')
rgb = image.astype(np.int16)
r, g, b = (rgb[:, :, 0], rgb[:, :, 1], rgb[:, :, 2])
red = (r > 190) & (g > 25) & (g < 145) & (b < 145) & (r - g > 70)
green = (r < 130) & (g > 100) & (b > 90) & (g - r > 35) & (b - r > 35)
candle_mask = red | green
chart_bbox = [0, int(h * 0.18), w, int(h * 0.75)]
volume_bbox = [0, int(h * 0.68), w, int(h * 0.88)]
print('RESULT_JSON_START')
print(json.dumps({'stage': 'inspect_layout', 'layout_kind': 'short_kline', 'image_size': [w, h], 'chart_bbox':chart_bbox, 'volume_bbox': volume_bbox}, ensure_ascii=False))
print('RESULT_JSON_END')
\end{TraceCode}

\traceoutputheading

\Needspace{4\baselineskip}
\begin{TraceOutput}
{"stage": "inspect_layout", "layout_kind": "short_kline", "image_size": [920, 749], "chart_bbox": [0,
134, 920, 561], "volume_bbox": [0, 509, 920, 659]}
\end{TraceOutput}

\subsubsection{Turn 2 of 6}

\traceplan{The short\_kline layout has a main-chart box of [0,134,920,561] and a volume region of [0,509,920,659]. Recover the three candlestick centers and verify their left-to-right order before mapping the tallest bar to its candlestick index.}

\Needspace{4\baselineskip}
\begin{TraceCode}
from PIL import Image, ImageDraw
import re
import json
import numpy as np
REQUIRED_PUBLIC_STATE = ('image', 'h', 'w', 'candle_mask')
missing_public_state = [name for name in REQUIRED_PUBLIC_STATE if name not in globals()]
if missing_public_state:
    raise RuntimeError('missing prior public state: ' + ','.join(missing_public_state))
GRID_CONTRACT = {'source': 'task_contract', 'expected_visible_days': [3, 12]}
def runs(values, max_gap=0):
    values = [int(v) for v in values]
    if not values:
        return []
    out = []
    start = previous = values[0]
    for value in values[1:]:
        if value <= previous + max_gap + 1:
            previous = value
        else:
            out.append((start, previous))
            start = previous = value
    out.append((start, previous))
    return out
chart_top, chart_bottom = (int(h * 0.18), int(h * 0.75))
columns = np.where(candle_mask[chart_top:chart_bottom].sum(0) >= 3)[0]
widths = sorted((right - left + 1 for left, right in runs(columns) if max(20, int(w * 0.04)) <= right - left + 1 <=int(w * 0.22)))
if not widths:
    raise ValueError('no measurable candle body width')
body_width = float(np.median(widths[:max(1, (len(widths) + 1) // 2)]))
volume_columns = np.where(candle_mask[int(h * 0.68):int(h * 0.88)].sum(0) >= 1)[0]
if not len(volume_columns):
    raise ValueError('no visible volume span')
x_left, x_right = (int(volume_columns.min()), int(volume_columns.max()))
span = x_right - x_left + 1
day_count = int(round(2.0 / 3.0 * span / max(body_width, 1.0)))
if not 3 <= day_count <= 12:
    raise ValueError('short grid estimate outside visible bounds')
pitch = span / day_count
if abs(2.0 / 3.0 * span / day_count - body_width) > max(3.0, body_width * 0.12):
    raise ValueError('grid/body geometry is inconsistent')
centers = [x_left + (index + 0.5) * pitch for index in range(day_count)]
chart_bbox = [x_left, chart_top, x_right + 1, chart_bottom]
volume_bbox = [x_left, int(h * 0.68), x_right + 1, int(h * 0.88)]
print('RESULT_JSON_START')
print(json.dumps({'stage': 'recover_grid', 'day_count': day_count, 'pitch': round(float(pitch), 3), 'centers':[round(float(x), 2) for x in centers], 'chart_bbox': chart_bbox, 'volume_bbox': volume_bbox, 'contract_prior':GRID_CONTRACT}, ensure_ascii=False))
print('RESULT_JSON_END')
\end{TraceCode}

\traceoutputheading

\Needspace{4\baselineskip}
\begin{TraceOutput}
{"stage": "recover_grid", "day_count": 3, "pitch": 202.0, "centers": [231.0, 433.0, 635.0], "chart_bbox":
[130, 134, 736, 561], "volume_bbox": [130, 509, 736, 659], "contract_prior": {"source": "task_contract",
"expected_visible_days": [3, 12]}}
\end{TraceOutput}

\subsubsection{Turn 3 of 6}

\traceplan{The refined volume panel is [130,509,736,659], with candlestick centers at x=231, 433, and 635. Create a focused crop so the height difference among the three aligned volume bars can be checked directly.}

\Needspace{4\baselineskip}
\begin{TraceCode}
import numpy as np
import json
import re
from PIL import Image, ImageDraw
FOCUS_MODALITIES=['volume']
source_image=Image.fromarray(image.astype("uint8")).convert("RGB")
if FOCUS_MODALITIES==["volume"]:
    raw_box=list(volume_bbox); purpose="volume_panel"
elif "volume" in FOCUS_MODALITIES and ("candle" in FOCUS_MODALITIES or "ma" in FOCUS_MODALITIES):
    legend_margin=max(70,int((chart_bbox[3]-chart_bbox[1])*.13)) if "ma" in FOCUS_MODALITIES else 0
    raw_box=[chart_bbox[0],max(0,chart_bbox[1]-legend_margin),chart_bbox[2],volume_bbox[3]];purpose="price_volume_alignment"
else:
    legend_margin=max(70,int((chart_bbox[3]-chart_bbox[1])*.13)) if "ma" in FOCUS_MODALITIES else 0
    raw_box=[chart_bbox[0],max(0,chart_bbox[1]-legend_margin),chart_bbox[2],chart_bbox[3]];purpose="price_panel_with_ma_legend" if "ma" in FOCUS_MODALITIES else "price_panel"
pad=max(6,int(round(pitch*.5)))
focus_crop_bbox=[max(0,int(raw_box[0])-pad),max(0,int(raw_box[1])-pad),min(source_image.width,int(raw_box[2])+pad),min(source_image.height,int(raw_box[3])+pad)]
result_image=source_image.crop(tuple(focus_crop_bbox))
border=ImageDraw.Draw(result_image);border.rectangle((1,1,result_image.width-2,result_image.height-2),outline=(255,170,0),width=3)
# Promote the reviewed crop to the persistent visual workspace.  All later
# perception cells use these local pixels and local coordinates; the original
# screenshot is not silently re-read.
focus_origin=[focus_crop_bbox[0],focus_crop_bbox[1]]
crop_x0,crop_y0,crop_x1,crop_y1=focus_crop_bbox
image=np.asarray(source_image.crop(tuple(focus_crop_bbox))).copy(); h,w=image.shape[:2]
chart_bbox=[int(chart_bbox[0])-crop_x0,int(chart_bbox[1])-crop_y0,int(chart_bbox[2])-crop_x0,int(chart_bbox[3])-crop_y0]
volume_bbox=[int(volume_bbox[0])-crop_x0,int(volume_bbox[1])-crop_y0,int(volume_bbox[2])-crop_x0,int(volume_bbox[3])-crop_y0]
centers=[float(value)-crop_x0 for value in centers]
if "candle_mask" in globals(): candle_mask=candle_mask[crop_y0:crop_y1,crop_x0:crop_x1].copy()
for name in ("rgb","r","g","b","red","green","red_all","green_all","mask_all"):
    if name in globals(): globals()[name]=globals()[name][crop_y0:crop_y1,crop_x0:crop_x1].copy()
# Layout/grid cells also publish these scalar aliases.  Rebind them to the
# promoted crop coordinate frame; leaving the original-image y1/y2 values in
# the persistent sandbox makes later extractors scan the wrong rows.
x1,y1,x2,y2=[int(value) for value in chart_bbox]
payload={"stage":"render_focus_crop","task":'volume_extreme',"purpose":purpose,"modalities":FOCUS_MODALITIES,"source_crop_bbox":focus_crop_bbox,"result_image_size":list(result_image.size),"must_review_before_next_extraction":True,"workspace_promoted":True,"workspace_coordinate_frame":"focus_crop_local"}
print("RESULT_JSON_START"); print(json.dumps(payload,ensure_ascii=False,separators=(",",":")));print("RESULT_JSON_END")
\end{TraceCode}

\traceoutputheading

\Needspace{4\baselineskip}
\begin{TraceOutput}
{"stage": "render_focus_crop", "task": "volume_extreme", "purpose": "volume_panel", "modalities": ["volume"],
"source_crop_bbox": [29, 408, 837, 749], "result_image_size": [808, 341], "must_review_before_next_extraction":
true, "workspace_promoted": true, "workspace_coordinate_frame": "focus_crop_local"}
\end{TraceOutput}

\traceimage[0.7]{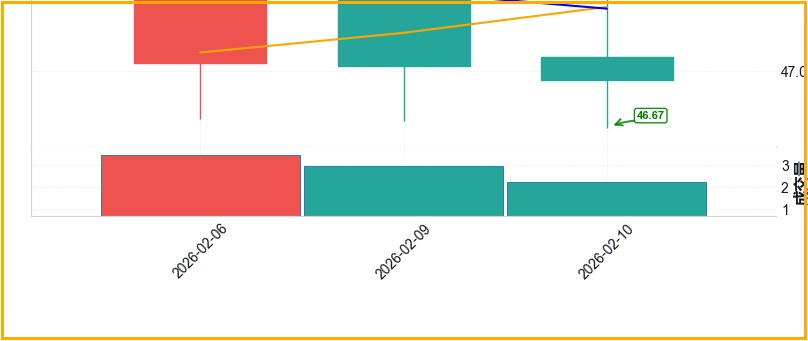}{Focused volume-panel evidence for the three candidates.}

\subsubsection{Turn 4 of 6}

\traceplan{The focused image contains only the volume panel and the three corresponding day positions; the leftmost bar is visibly tallest. Measure all three bar heights in their established horizontal order.}

\Needspace{4\baselineskip}
\begin{TraceCode}
from PIL import Image, ImageDraw
import re
import json
import numpy as np
REQUIRED_PUBLIC_STATE = ('image', 'volume_bbox', 'centers', 'pitch', 'red', 'green')
missing_public_state = [name for name in REQUIRED_PUBLIC_STATE if name not in globals()]
if missing_public_state:
    raise RuntimeError('missing prior public state: ' + ','.join(missing_public_state))
volume_bars = []
vleft, vtop, vright, vbottom = [int(value) for value in volume_bbox]
for day, center in enumerate(centers, 1):
    half = max(6, int(pitch * 0.38))
    left, right = (max(0, int(center) - half), min(w, int(center) + half + 1))
    candidates = []
    for color, mask in (('red', red), ('green', green)):
        local = mask[vtop:vbottom, left:right]
        rows = np.where(local.sum(1) >= max(2, int(pitch * 0.25)))[0]
        if len(rows):
            candidates.append((int(local.sum()), color, rows))
    if not candidates:
        raise ValueError('missing short volume bar at day %d' % day)
    _, color, rows = max(candidates, key=lambda item: item[0])
    volume_bars.append({'day_index': day, 'x': round(float(center), 2), 'height_px': int(rows.max() - rows.min() + 1), 'top_y': int(rows.min() + vtop), 'bottom_y': int(rows.max() + vtop), 'color': color})
print('RESULT_JSON_START')
print(json.dumps({'stage': 'extract_volume', 'coverage': len(volume_bars), 'bars': volume_bars}, ensure_ascii=False))
print('RESULT_JSON_END')
\end{TraceCode}

\traceoutputheading

\Needspace{4\baselineskip}
\begin{TraceOutput}
{"stage": "extract_volume", "coverage": 3, "bars": [{"day_index": 1, "x": 202.0, "height_px": 60, "top_y":
156, "bottom_y": 215, "color": "red"}, {"day_index": 2, "x": 404.0, "height_px": 50, "top_y": 166, "bottom_y":
215, "color": "green"}, {"day_index": 3, "x": 606.0, "height_px": 34, "top_y": 182, "bottom_y": 215,
"color": "green"}]}
\end{TraceOutput}

\subsubsection{Turn 5 of 6}

\traceplan{The measured heights are 60, 50, and 34 pixels for Candlesticks 1, 2, and 3, respectively. A direct comparison therefore identifies Candlestick 1 as the maximum-volume day.}

\Needspace{4\baselineskip}
\begin{TraceCode}
from PIL import Image, ImageDraw
import numpy as np
import re
import json
REQUIRED_PUBLIC_STATE = ('day_count', 'volume_bars')
missing_public_state = [name for name in REQUIRED_PUBLIC_STATE if name not in globals()]
if missing_public_state:
    raise RuntimeError('missing prior public state: ' + ','.join(missing_public_state))
TASK = 'volume_extreme'
PARAMS = {}
c_by_day = {}
v_by_day = {int(row['day_index']): row for row in volume_bars}
ma_by_day = {}
public_rows = [{'day_index': d, 'height_px': v_by_day[d]['height_px']} for d in sorted(v_by_day)]
selected = max(public_rows, key=lambda row: (row['height_px'], -row['day_index']))
answer = str(selected['day_index'])
evidence_days = [selected['day_index']]
flat_rows = []
stack = [public_rows]
while stack:
    item = stack.pop()
    if isinstance(item, dict):
        if 'day_index' in item or 'start_day' in item or 'end_day' in item:
            flat_rows.append(item)
        else:
            stack.extend(reversed(list(item.values())))
    elif isinstance(item, list):
        stack.extend(reversed(item))
candidate_days = sorted({int(row[key]) for row in flat_rows for key in ('day_index', 'start_day', 'end_day') if row.get(key) is not None})
checked_pairs = []
for row in flat_rows:
    if row.get('previous_day') is not None and row.get('day_index') is not None:
        checked_pairs.append([int(row['previous_day']), int(row['day_index'])])
    elif False and row.get('day_index') is not None and (int(row['day_index']) > 1):
        checked_pairs.append([int(row['day_index']) - 1, int(row['day_index'])])
checked_pairs = sorted({tuple(pair) for pair in checked_pairs})
scope_days = [] or candidate_days or list(range(1, day_count + 1))
positive_days = sorted(set((int(d) for d in evidence_days if d is not None)))
evidence_contract = {'version': 'semantic_evidence_v2', 'kind': 'volume_objects', 'modalities': ['volume'],'scope_days': scope_days, 'candidate_days': candidate_days, 'positive_days': positive_days, 'checked_pairs':[list(pair) for pair in checked_pairs], 'negative_case': str(answer) == '0' and (not positive_days),'negative_support': {'checked_days': candidate_days, 'checked_pairs': [list(pair) for pair in checked_pairs],'all_predicates_false': str(answer) == '0' and (not positive_days)}, 'field_support': {}, 'field_modalities': {}}
task_result = {'stage': 'compute_task', 'task': TASK, 'parameters': PARAMS, 'public_rows': public_rows,'evidence_days': positive_days, 'evidence_contract': evidence_contract, 'answer': str(answer)}
print('RESULT_JSON_START')
print(json.dumps(task_result, ensure_ascii=False, separators=(',', ':')))
print('RESULT_JSON_END')
\end{TraceCode}

\traceoutputheading

\Needspace{4\baselineskip}
\begin{TraceOutput}
{"stage": "compute_task", "task": "volume_extreme", "parameters": {}, "public_rows": [{"day_index": 1,
"height_px": 60}, {"day_index": 2, "height_px": 50}, {"day_index": 3, "height_px": 34}], "evidence_days":
[1], "evidence_contract": {"version": "semantic_evidence_v2", "kind": "volume_objects", "modalities":
["volume"], "scope_days": [1, 2, 3], "candidate_days": [1, 2, 3], "positive_days": [1], "checked_pairs":
[], "negative_case": false, "negative_support": {"checked_days": [1, 2, 3], "checked_pairs": [], "all_predicates_false":
false}, "field_support": {}, "field_modalities": {}}, "answer": "1"}
\end{TraceOutput}

\subsubsection{Turn 6 of 6}

\traceplan{The computed comparison confirms heights of 60, 50, and 34 pixels and selects only Day 1. Render answer-level evidence that preserves this maximum-volume relationship.}

\Needspace{4\baselineskip}
\begin{TraceCode}
import numpy as np
import json
import re
from PIL import Image, ImageDraw
result_image = Image.fromarray(image.astype("uint8")).convert("RGB")
draw = ImageDraw.Draw(result_image)
contract = task_result['evidence_contract']
kind = 'interval_predicate'
cmap = {'primary': (255, 0, 255), 'secondary': (0, 180, 255), 'negative': (235, 70, 50), 'context': (255, 170, 0),'muted': (70, 170, 255)}
regions = []
def add_region(name, boxes, days=None, relation=None):
    clean = [[int(v) for v in box] for box in boxes if len(box) == 4 and box[2] > box[0] and (box[3] > box[1])]
    if clean:
        regions.append({'artifact': name, 'day_indices': sorted(set((int(d) for d in days or [] if d is not None))),'relation': relation, 'pixel_regions': clean})
def candle_box(day, body_only=False):
    row = c_by_day.get(int(day))
    if not row:
        return None
    x = int(round(centers[int(day) - 1]))
    top, bottom = row['body_y'] if body_only else (row['high_y'], row['low_y'])
    return [int(x - pitch * 0.34), int(top) - 2, int(x + pitch * 0.34), int(bottom) + 3]
def volume_box(day):
    row = v_by_day.get(int(day))
    if not row or row.get('height_px') is None:
        return None
    x = int(round(centers[int(day) - 1]))
    return [int(x - pitch * 0.34), int(row['top_y']) - 2, int(x + pitch * 0.34), int(row['bottom_y']) + 2]
def ma_boxes(day, names=None):
    values = ma_by_day.get(int(day), {})
    x = int(round(centers[int(day) - 1]))
    boxes = []
    for name in names or sorted(values):
        if name not in values:
            continue
        y = int(round(values[name]))
        boxes.append([x - 5, y - 5, x + 6, y + 6])
    return boxes
flat = []
stack = [task_result['public_rows']]
semantic_annotations = []
negative_rendered = False
while stack:
    item = stack.pop()
    if isinstance(item, dict):
        if 'day_index' in item or 'start_day' in item or 'end_day' in item:
            flat.append(item)
        else:
            stack.extend(reversed(list(item.values())))
    elif isinstance(item, list):
        stack.extend(reversed(item))
by_day = {int(row['day_index']): row for row in flat if row.get('day_index') is not None}
scope = [int(d) for d in contract['scope_days'] if 1 <= int(d) <= day_count]
positive = [int(d) for d in contract['positive_days'] if 1 <= int(d) <= day_count]
candidate = [int(d) for d in contract['candidate_days'] if 1 <= int(d) <= day_count]
if scope:
    left, right = (min(scope), max(scope))
    x_left = int(round(centers[left - 1]))
    x_right = int(round(centers[right - 1]))
    scope_top = int(volume_bbox[1]) if set(contract.get('modalities') or []) == {'volume'} else int(chart_bbox[1])
    y = scope_top + 7
    draw.line((x_left, y, x_right, y), fill=cmap['context'], width=3)
    draw.line((x_left, y - 6, x_left, y + 6), fill=cmap['context'], width=3)
    draw.line((x_right, y - 6, x_right, y + 6), fill=cmap['context'], width=3)
    draw.text((x_left + 3, y + 3), 'scope D%d-D%d' % (left, right), fill=cmap['context'])
    add_region('checked_scope', [[min(x_left, x_right) - 2, y - 7, max(x_left, x_right) + 3, y + 18]], scope,'bounded_check_range')
boxes = []
if TASK in {'interval_bullish_ma_alignment_count', 'interval_bearish_ma_alignment_count'}:
    names = ('MA5', 'MA10', 'MA20', 'MA30')
    palette = [(30, 30, 30), (230, 155, 0), (240, 80, 120), (85, 180, 65)]
    rows = sorted(flat, key=lambda row: int(row['day_index']))
    checked = []
    for row in rows:
        day = int(row['day_index'])
        values = [float(value) for value in row['values']]
        x = int(round(centers[day - 1]))
        local = []
        for name, y, color in zip(names, values, palette):
            box = [x - 4, int(round(y)) - 4, x + 5, int(round(y)) + 5]
            draw.ellipse(tuple(box), outline=color, fill=color, width=1)
            local.append(box)
        draw.line([(x, int(round(y))) for y in values], fill=cmap['primary'] if row.get('match') else cmap['negative'], width=2)
        mark = 'T' if row.get('match') else 'F'
        draw.text((x - 4, max(0, min((int(round(y)) for y in values)) - 14)), mark, fill=cmap['primary'] if row.get('match') else cmap['negative'])
        boxes.extend(local)
        checked.append((day, bool(row.get('match'))))
    hits = sum((match for _, match in checked))
    alignment_name = 'bullish' if 'bullish' in TASK else 'bearish'
    label = '%s %d/%d; T/F GRID BELOW' % (alignment_name.upper(), hits, len(checked))
    label_x = max(8, int(round(centers[min(scope) - 1] - pitch * 0.35)) if scope else int(chart_bbox[0]) + 8)
    label_y = int(chart_bbox[1]) + 38
    draw.rectangle((label_x - 5, label_y - 4, label_x + max(520, len(label) * 6), label_y + 17), fill=(255, 255, 255), outline=cmap['negative'] if not positive else cmap['primary'], width=2)
    draw.text((label_x, label_y), label, fill=cmap['negative'] if not positive else cmap['primary'])
    semantic_annotations.append({'type': 'ma_alignment_summary', 'label': label, 'checked_days': [day for day, _ in checked], 'match_days': [day for day, match in checked if match]})
    negative_rendered = contract['negative_case']
    add_region('daily_four_ma_order', boxes, [day for day, _ in checked],'four_named_ma_points_and_strict_order_predicate')
elif TASK in {'volume_extreme', 'volume_max', 'volume_min'}:
    for day in candidate:
        box = volume_box(day)
        if not box:
            continue
        row = by_day.get(day, {})
        selected = day in positive
        color = cmap['primary'] if selected else cmap['secondary']
        draw.rectangle(tuple(box), outline=color, width=4 if selected else 2)
        if selected:
            draw.text((box[0], max(0, box[1] - 14)), 'SELECTED D%d' % day, fill=color)
        boxes.append(box)
    add_region('compared_volume_bars', boxes, candidate, 'argmax_or_argmin_over_all_volume_heights')
else:
    days = [] if contract['negative_case'] else positive or candidate or scope
    for day in days:
        local = []
        if 'candle' in contract['modalities']:
            box = candle_box(day, True)
            local += [box] if box else []
        if 'volume' in contract['modalities']:
            box = volume_box(day)
            local += [box] if box else []
        if 'ma' in contract['modalities']:
            local += ma_boxes(day)
        color = cmap['primary'] if day in positive else cmap['negative'] if contract['negative_case'] else cmap['secondary']
        for box in local:
            draw.rectangle(tuple(box), outline=color, width=3 if day in positive else 2)
        boxes.extend(local)
    add_region('interval_predicate_objects', boxes, days, 'complete_interval_predicate_support')
if contract['negative_case'] and (not negative_rendered):
    checked = candidate or scope
    marks = []
    y = int(chart_bbox[1]) + 28
    for day in checked:
        x = int(round(centers[day - 1]))
        box = [x - 4, y - 4, x + 5, y + 5]
        draw.line((box[0], box[1], box[2], box[3]), fill=cmap['negative'], width=2)
        draw.line((box[0], box[3], box[2], box[1]), fill=cmap['negative'], width=2)
        marks.append(box)
    if marks:
        draw.text((marks[0][0], y + 8), '0 matches / %d checked' % len(checked), fill=cmap['negative'])
    add_region('negative_predicate_checks', marks, checked, 'all_checked_predicates_false')
if not regions:
    raise ValueError('semantic evidence renderer produced no answer support')
from PIL import ImageFont
try:
    semantic_font = ImageFont.truetype('DejaVuSans.ttf', 14)
except Exception:
    semantic_font = ImageFont.load_default()
semantic_summary_bounds = []
def draw_semantic_label(text, x, y, color=(145, 0, 220)):
    x, y = (int(x), int(y))
    bbox = draw.textbbox((x, y), text, font=semantic_font)
    box = [bbox[0] - 5, bbox[1] - 4, bbox[2] + 5, bbox[3] + 4]
    draw.rectangle(tuple(box), fill=(255, 255, 255), outline=color, width=2)
    draw.text((x, y), text, fill=color, font=semantic_font)
    semantic_summary_bounds.append(box)
    semantic_annotations.append({'type': 'task_summary', 'task': TASK, 'label': text})
rows = sorted(list(task_result.get('public_rows') or []), key=lambda row: int(row['day_index']))
if not rows:
    raise ValueError('volume extrema evidence lacks public rows')
selected_days = sorted((int(day) for day in contract.get('positive_days') or []))
if len(selected_days) != 1:
    raise ValueError('volume extrema evidence requires one selected day')
selected_day = selected_days[0]
selected_row = next((row for row in rows if int(row['day_index']) == selected_day), None)
if selected_row is None:
    raise ValueError('selected volume day is absent from public rows')
cols = min(12, len(rows))
cell_w = max(70, int(result_image.width / cols))
cell_h = 30
grid_rows = (len(rows) + cols - 1) // cols
summary_h = 30
panel_h = summary_h + grid_rows * cell_h + 8
old_image = result_image
extended = Image.new('RGB', (old_image.width, old_image.height + panel_h), (255, 255, 255))
extended.paste(old_image, (0, 0))
result_image = extended
draw = ImageDraw.Draw(result_image)
panel_top = old_image.height
mode = 'MAX' if TASK in {'volume_extreme', 'volume_max'} else 'MIN'
summary = '%s VOLUME: D%d = %spx; COMPARED %d DAYS' % (mode, selected_day, selected_row.get('height_px', '?'),len(rows))
draw.rectangle((1, panel_top + 1, result_image.width - 2, result_image.height - 2), fill=(255, 255, 255),outline=(145, 0, 220), width=3)
draw.text((8, panel_top + 7), summary, fill=(145, 0, 220), font=semantic_font)
for index, row in enumerate(rows):
    rr, cc = divmod(index, cols)
    xx = cc * cell_w
    yy = panel_top + summary_h + rr * cell_h
    day = int(row['day_index'])
    chosen = day == selected_day
    color = (255, 0, 255) if chosen else (0, 150, 205)
    fill = (255, 230, 255) if chosen else (242, 250, 255)
    draw.rectangle((xx + 2, yy + 1, min(result_image.width - 2, xx + cell_w - 2), yy + cell_h - 2), fill=fill,outline=color, width=3 if chosen else 1)
    draw.text((xx + 7, yy + 7), 'D%d  %spx' % (day, row.get('height_px', '?')), fill=color, font=semantic_font)
panel_bbox = [1, panel_top + 1, result_image.width - 2, result_image.height - 2]
semantic_summary_bounds.append(panel_bbox)
semantic_annotations.append({'type': 'volume_extrema_audit_grid', 'mode': mode, 'selected_day': selected_day,'selected_height_px': selected_row.get('height_px'), 'candidate_count': len(rows), 'all_day_height_pairs_visible':True, 'panel_bbox': panel_bbox})
mods=set(contract.get("modalities") or [])
if mods=={"volume"}:
    y_top,y_bottom=int(volume_bbox[1]),int(volume_bbox[3])
elif "volume" in mods and ("candle" in mods or "ma" in mods):
    y_top,y_bottom=int(chart_bbox[1]),int(volume_bbox[3])
else:
    y_top,y_bottom=int(chart_bbox[1]),int(chart_bbox[3])
visible_scope=sorted({int(d) for key in ("scope_days","candidate_days","positive_days") for d in (contract.get(key) or []) if 1<=int(d)<=day_count})
if visible_scope:
    x_left=int(round(centers[min(visible_scope)-1]-pitch*.7));x_right=int(round(centers[max(visible_scope)-1]+pitch*.7))
else:
    x_left,x_right=int(chart_bbox[0]),int(chart_bbox[2])
pad=max(6,int(round(pitch*.35)))
if semantic_summary_bounds:
    x_left=min([x_left]+[int(box[0])-pad for box in semantic_summary_bounds]); x_right=max([x_right]+[int(box[2])+pad for box in semantic_summary_bounds])
    y_top=min([y_top]+[int(box[1])-pad for box in semantic_summary_bounds]); y_bottom=max([y_bottom]+[int(box[3])+pad for box in semantic_summary_bounds])
evidence_crop_bbox=[max(0,x_left-pad),max(0,y_top-pad),min(result_image.width,x_right+pad),min(result_image.height,y_bottom+pad)]
result_image=result_image.crop(tuple(evidence_crop_bbox))
payload = {'stage': 'render_semantic_answer_evidence_v2', 'task': TASK, 'answer': task_result['answer'],'evidence_contract': contract, 'evidence_regions': regions, 'semantic_annotations': semantic_annotations,'whole_panel_fallback': False}
payload['source_coordinate_frame']='reviewed_focus_crop'; payload['focus_origin_in_original']=focus_origin;payload['result_image_crop_bbox']=evidence_crop_bbox; payload['result_image_size']=list(result_image.size);payload['semantic_summary_bounds']=semantic_summary_bounds
print('RESULT_JSON_START')
print(json.dumps(payload, ensure_ascii=False, separators=(',', ':')))
print('RESULT_JSON_END')
\end{TraceCode}

\traceoutputheading

\Needspace{4\baselineskip}
\begin{TraceOutput}
{"stage": "render_semantic_answer_evidence_v2", "task": "volume_extreme", "answer": "1", "evidence_contract":
{"version": "semantic_evidence_v2", "kind": "volume_objects", "modalities": ["volume"], "scope_days":
[1, 2, 3], "candidate_days": [1, 2, 3], "positive_days": [1], "checked_pairs": [], "negative_case": false,
"negative_support": {"checked_days": [1, 2, 3], "checked_pairs": [], "all_predicates_false": false},
"field_support": {}, "field_modalities": {}}, "evidence_regions": [{"artifact": "checked_scope", "day_indices":
[1, 2, 3], "relation": "bounded_check_range", "pixel_regions": [[200, 101, 609, 126]]}, {"artifact":
"compared_volume_bars", "day_indices": [1, 2, 3], "relation": "argmax_or_argmin_over_all_volume_heights",
"pixel_regions": [[133, 154, 270, 217], [335, 164, 472, 217], [537, 180, 674, 217]]}], "semantic_annotations":
[{"type": "volume_extrema_audit_grid", "mode": "MAX", "selected_day": 1, "selected_height_px": 60, "candidate_count":
3, "all_day_height_pairs_visible": true, "panel_bbox": [1, 342, 806, 407]}], "whole_panel_fallback":
false, "source_coordinate_frame": "reviewed_focus_crop", "focus_origin_in_original": [29, 408], "result_image_crop_bbox":
[0, 30, 808, 409], "result_image_size": [808, 379], "semantic_summary_bounds": [[1, 342, 806, 407]]}
\end{TraceOutput}

\traceimage[0.7]{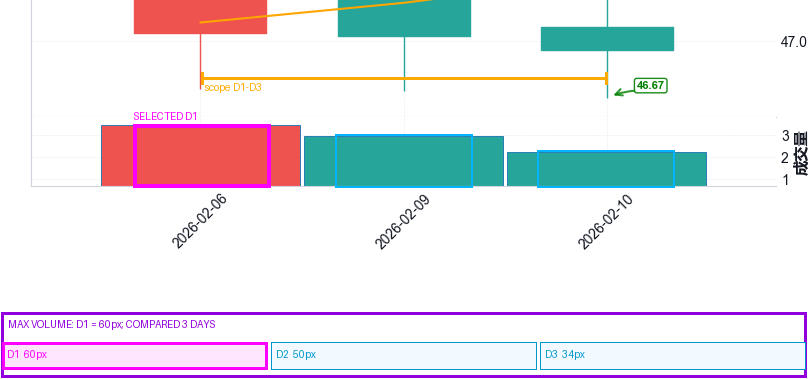}{Answer-level evidence for the maximum-volume candlestick.}

\subsubsection{Final Evidence Review and Answer}

The evidence strip states 'MAX VOLUME: D1 = 60px; COMPARED 3 DAYS' and lists D1=60px, D2=50px, and D3=34px, with D1 selected. The final answer is therefore the first candlestick.

\begin{center}\Large\bfseries 1\end{center}

\end{document}